\documentclass{article}
\usepackage{arxiv}
\usepackage[T1]{fontenc}
\usepackage[utf8]{inputenc}
\usepackage{amsmath,amssymb,booktabs,tabularx,array,longtable}
\usepackage{xcolor,graphicx,microtype}
\usepackage{placeins}
\usepackage[numbers,sort&compress]{natbib}
\usepackage[colorlinks=true]{hyperref}
\newcolumntype{Y}{>{\raggedright\arraybackslash}X}
\definecolor{reportblue}{HTML}{245A81}
\definecolor{reportgreen}{HTML}{26745A}
\definecolor{reportorange}{HTML}{B56D28}
\hypersetup{linkcolor=reportblue,citecolor=reportblue,urlcolor=reportblue,
  pdftitle={How Good Are Coding Agents at Automatic Algorithm Design?},
  pdfsubject={Technical report}}
\renewcommand{\arraystretch}{1.12}
\usepackage{capt-of}
\newsavebox{\AppendixTableBox}
\newcommand{\AppendixFitTable}[1]{%
  \sbox{\AppendixTableBox}{#1}%
  \ifdim\wd\AppendixTableBox>\linewidth
    \resizebox{\linewidth}{!}{\usebox{\AppendixTableBox}}%
  \else
    \usebox{\AppendixTableBox}%
  \fi}

\usepackage{fvextra}

\usepackage{multirow}

\renewcommand{\headeright}{}
\renewcommand{\undertitle}{}
\renewcommand{\shorttitle}{Coding Agents for Automatic Algorithm Design}
\title{How Good Are Coding Agents at\\ Automatic Algorithm Design?}
\author{%
Haoze Lv\textsuperscript{1,2} \qquad Shaofeng Zhang\textsuperscript{1,2} \qquad Shengcai Liu\textsuperscript{1,2}\thanks{Corresponding authors.} \qquad Ke Tang\textsuperscript{1,2}\footnotemark[1]\\[0.5em]
\normalfont\small\textsuperscript{1}Guangdong Provincial Key Laboratory of Brain-Inspired Intelligent Computation,\\
\normalfont\small Department of Computer Science and Engineering,\\
\normalfont\small Southern University of Science and Technology, Shenzhen 518055, China\\
\normalfont\small\textsuperscript{2}Zhongguancun Academy, Beijing, China\\
\normalfont\small\texttt{liusc3@sustech.edu.cn}, \texttt{tangk3@sustech.edu.cn}%
}
\date{}

\newif\ifshowappendices
\showappendicestrue
\ifshowappendices\else
  \input{frontmatter/hidden_appendix_references}
\fi
\begin{document}
\maketitle

\begin{abstract}
High-performance solvers for combinatorial optimization (CO) problems typically rely on human expert design. In recent years, large language models (LLMs) have advanced automatic algorithm design (AAD), with most existing methods focusing on its subfield of automatic heuristic design (AHD)---using predefined evolutionary or tree-search workflows to organize heuristic generation, evaluation, and improvement. As models become more capable of planning and tool use, general-purpose coding agents can autonomously locate and modify code and adapt their development process based on execution feedback, offering new opportunities to move beyond heuristic-component generation toward broader solver-level AAD, reducing reliance on fixed workflows and human-specified modification targets. However, a systematic evaluation of their performance in AAD for discrete optimization remains lacking. To bridge this gap, we conduct an empirical study across discrete optimization problems spanning different levels of difficulty---from classical problems of increasing complexity (permutation flow-shop scheduling and the traveling salesman problem) to complex real-world industrial challenges (the 2018, 2020, and 2022 ROADEF/EURO Challenges)---under two practical scenarios: improving existing solvers and developing solvers from scratch. Within a unified evaluation framework, we benchmark four coding agents paired with frontier LLMs---Codex (GPT-6-Astra), Kimi-code (KIMI-K3), Zcode (GLM-5.3), and DeepSeek Harness (DeepSeek-V4-Flash)---against fixed-workflow baselines (ReEvo and OpenEvolve) and strong human references, evaluating solution quality, solving success rates, held-out generalization, and code modifications. Results show that coding agents can match or outperform the evaluated fixed-workflow methods and further improve existing high-performance solvers, with gains transferring to unseen instances across multiple domains. However, their capabilities remain bounded by problem complexity and the development starting point: when developing solvers from scratch, coding agents can outperform human references on relatively simple problems but still fall far behind mature expert solvers on complex ones; moreover, their improvements to highly complex solvers do not necessarily surpass simple parameter adjustments.
\end{abstract}
\begin{center}
\textbf{Code and Benchmarks:} \url{https://github.com/Antoniano1963/Agentic-AAD}
\end{center}

\section{Introduction}

Combinatorial optimization (CO) concerns finding high-quality decisions in discrete, constrained search spaces and underpins applications in transportation, production scheduling, and resource allocation \cite{vidal2022cvrp,tang2009carp,dauzere2024fjsp,delorme2016bpp,katoh1998ra}. Many important CO problems are NP-hard, and obtaining exact optima for large or heavily constrained instances can be impractical within available computing budgets. Heuristics and metaheuristics therefore play a central role in obtaining useful solutions within reasonable time. Their effectiveness depends on choices about solution representations, construction procedures, neighborhoods, acceptance criteria, and search control, as well as the efficiency of their implementation. Making these choices requires substantial domain knowledge and algorithmic expertise. Even for extensively studied problems, competitive solvers emerge through repeated analysis, implementation, and experimentation, and adapting them to new instance distributions or execution limits remains a demanding engineering task \cite{nco}. This dependence on expertise and costly trial and error motivates the automation of algorithm design.

Automatic algorithm design (AAD) aims to reduce reliance on human expertise by automating the design and refinement of complete algorithms. Traditional AAD approaches, exemplified by genetic programming, evolve decision rules under manually specified representations and building blocks \cite{gphh}. Because such work typically targets an individual decision rule rather than an entire algorithm, we refer to it in this study as automatic heuristic design (AHD)---a subset of AAD focused on optimizing a specific operator or heuristic component within a fixed, human-designed algorithmic framework. More recently, large language models (LLMs) have greatly accelerated this direction through their ability to understand problem descriptions, express algorithmic ideas, and generate code. Recent studies embed LLMs within population-based evolutionary search, tree search, and related workflows to iteratively refine candidate heuristics through generation--evaluation--feedback loops, achieving competitive performance on diverse CO tasks \cite{funsearch,eoh,reevo,mcts}.

Although these LLM-based methods have achieved notable progress, they still largely remain at the AHD level and exhibit three clear limitations compared with realizing true AAD. First, many methods rely on predefined evolutionary or tree-search procedures, with external rules governing the model's generation, evaluation, and refinement of heuristics, restricting its scope for autonomously planning the development process \cite{eoh,reevo,mcts}. Second, many methods require human experts to first identify one or a few heuristic components that most strongly affect the target algorithm's performance; their autonomy primarily concerns searching for implementations within designated interfaces rather than diagnosing and improving a complete solver. Finally, existing evaluations often focus on tasks with limited instance sizes and short solver runtime budgets, and the heuristic components being optimized are relatively simple, providing insufficient evidence of complex algorithm-development capabilities; comparisons with advanced human-designed algorithms in the relevant domains under the same computational conditions also remain insufficient. Although a few studies have attempted to improve high-performing algorithms in specific domains, their modification scope often still requires human specification, or they rely on workflows specially designed for particular tasks \cite{ailsahd,pyvrpplus,automodsat}. As LLMs advance in planning and tool use, general-purpose coding agents can now understand codebases, autonomously locate and modify code, and use execution feedback to carry out sustained development tasks, offering new possibilities for overcoming these limitations \cite{sweagent}. These capabilities allow models to decide what to modify, how to validate changes, and when to adjust their development direction, potentially extending automation from component-level heuristic design to complete solver-level AAD. Although prior work has begun to explore agent-driven algorithm design and evaluation \cite{ahdagent,simpleevol,alebench,frontieror}, the ability of general-purpose coding agents to autonomously design and improve algorithms for large-scale, complex discrete optimization problems---particularly to compete with established, high-performance human-designed methods under longer solver runtime budgets---remains to be systematically tested. Our central research question is therefore:
\begin{quote}\centering\emph{How good are coding agents at automatic algorithm design with little human intervention?}\end{quote}

To better understand the capabilities and limitations of general-purpose coding agents in AAD for CO, we conduct a comparative empirical study. We select two classic CO problems, the traveling salesman problem (TSP) \cite{tspbook} and the permutation flow-shop scheduling problem (PFSP) \cite{pfspreview}, and use large-scale, challenging datasets \cite{tsplib,nationaltsp,vlsitsp,vfr} to examine agents' ability to develop high-performance algorithms on demanding tasks. We also include problems from the 2018, 2020, and 2022 editions of the ROADEF Challenge \cite{roadef,roadef2020,r22subject} to cover CO problems with complex constraints and engineering requirements that more closely reflect practical applications. Across these problems, we consider two development scenarios commonly encountered in practice: development from scratch without a complete solver supplied, and improvement of existing algorithms to meet application requirements. To assess current agentic capabilities, we evaluate representative coding agents paired with frontier LLMs---including Codex (GPT-6-Astra), Kimi-code (KIMI-K3), Zcode (GLM-5.3), and DeepSeek Harness (DeepSeek-V4-Flash)---and compare them against strong human-designed reference solvers, representative AHD methods such as ReEvo \cite{reevo}, and program-evolution frameworks for broader code-optimization tasks such as OpenEvolve \cite{openevolve}. Evaluation focuses on effectiveness, efficiency, and generalization, examining the solution quality, computational cost, and performance on unseen instances of the developed algorithms. We further analyze development trajectories, code changes, stopping behavior, and computational expenditure, linking final solver performance to the development process to better understand the benefits, limitations, and applicability of different development methods.

Our study yields three main findings:
\begin{itemize}
\item \textbf{Competitiveness against fixed workflows.} General-purpose coding agents can autonomously choose modification targets and development steps while achieving performance comparable to or better than fixed-workflow methods. They outperform the ReEvo adaptation, which requires a human-specified optimization target, on PFSP, and outperform OpenEvolve using the same underlying model on the more complex TSP/EAX improvement task, whereas OpenEvolve achieves the lower same-model PFSP gap. On ROADEF 2018, OpenEvolve retains the starting solver, and on ROADEF 2020, it improves the human baseline on commonly solved test instances.
\item \textbf{Effectiveness in improving existing algorithms.} Coding agents show strong capabilities in improving existing high-performance algorithms. They reduce training gaps on classical CO problems and improve solution quality and solving success rates on industrial problems; these gains persist on unseen instances in multiple domains, indicating that agents can effectively use and further enhance existing expert designs.
\item \textbf{Limits imposed by algorithmic complexity and starting point.} These capabilities remain limited by algorithmic complexity and the development starting point. Although development from scratch outperforms the human algorithms compared here on PFSP, it remains substantially behind mature expert solvers on TSP, ROADEF 2018, and ROADEF 2022. For a complex algorithm such as EAX, gains over the original implementation also do not exceed those obtainable through a simple parameter adjustment.
\end{itemize}
These results show that coding agents can already serve as effective algorithm-development tools, but constructing advanced algorithms without an existing implementation and improving the core mechanisms of complex solvers remain important challenges.

The remainder of this report is organized as follows. Section~\ref{sec:review} reviews LLM-based algorithm design and its evaluation. Section~\ref{sec:design} describes the comparative study. Section~\ref{sec:results} presents the experimental results and analysis. Section~\ref{sec:discussion} discusses the implications, limitations, and conclusions.

\section{LLM-Based Algorithm Design: Methods and Evaluation}

\label{sec:review}

\subsection{Preliminaries}

Automatic algorithm design (AAD) aims to automatically create, configure, or improve algorithms that perform well on a target class of problems, encompassing complete solvers, search architectures, hybrid exact--heuristic pipelines, data structures, and runtime resource management. Automatic heuristic design (AHD) is a key subset of AAD that focuses specifically on heuristics and their constituent decision rules or operators (such as priority functions or destroy--repair rules), typically operating within a human-specified solver scaffold and closely connected to the heuristic-generation branch of hyper-heuristics. This research direction predates LLMs: genetic programming, for example, evolves executable expressions or programs from a specified representation and primitive set \cite{gphh}. LLM-based AHD continues this objective while changing how candidates are represented and generated: natural-language problem descriptions, existing code, and execution feedback can jointly condition the generation of new heuristic implementations \cite{eoh,reevo,mcts}. Across both AHD and broader AAD, the object being designed is reusable algorithmic code that is subsequently executed on problem instances, rather than asking an LLM to output the solution to a single instance. In this report, we adopt the broader AAD perspective, covering both heuristic-component generation and codebase-level development or improvement of complete discrete optimization solvers.

Coding agents arise from the extension of code-generation models into interactive software-development systems. Code completion and conversational assistance primarily return code for a developer to integrate and execute. An agent instead connects the model to a working environment and lets it choose a sequence of actions: inspect a repository, locate relevant functions, edit files, invoke a compiler or test suite, and revise its implementation after observing the results. SWE-agent illustrates this transition in research by studying how an agent--computer interface enables an LLM to navigate repositories, modify code, and address software issues \cite{sweagent}. The central change is therefore from generating a code response to carrying out a development task through repeated interaction with tools and feedback.

This approach is reflected in a broad product ecosystem. Cursor integrates agentic code search, editing, and terminal execution into an editor; Anthropic's Claude Code provides repository-level development through terminal and other interfaces; OpenAI's Codex exposes a coding-agent execution system through its application, command-line interface, and IDE extension; and Google's Gemini CLI provides an open-source terminal agent \cite{cursoragent,claudecode,codexharness,geminicli}. These examples illustrate the spread of agentic development across software vendors. A coding agent is a system built around a model, instructions, context management, and tools, rather than simply another name for the underlying LLM. For algorithm development, this distinction matters because the system can investigate and modify a codebase beyond a predefined heuristic function. It provides a practical basis for asking whether general software-development agents can also improve discrete optimization algorithms.

\subsection{LLM-Based Automatic Algorithm Design}

LLMs' capabilities in natural-language understanding, code generation, and reasoning from feedback have made them important tools for automatic algorithm design, evolving from localized heuristic generation toward complete solver development. Research has developed along two main directions: general-purpose frameworks that organize candidate generation, evaluation, and improvement across problems, and methods that exploit specific problem or solver structures to design and optimize heuristic rules, search operators, and hybrid solver pipelines.

\subsubsection{General-purpose AAD frameworks}

General-purpose AAD frameworks use LLMs to generate candidate code and execution feedback to organize iterative search, progressing from component-level AHD to full-program evolution. Early representative methods typically adopt evolutionary frameworks: FunSearch combines program generation, automated evaluation, and population evolution to discover improved mathematical constructions and online bin-packing heuristics \cite{funsearch}; EoH jointly evolves natural-language algorithmic ideas and code implementations \cite{eoh}; and ReEvo uses short- and long-term reflection to translate performance differences between candidates into guidance for subsequent generation, with applications to constructive heuristics, guided local search, ant colony optimization, and genetic algorithms \cite{reevo}.

Subsequent work further improves search organization, diversity control, and the scope of design objectives. MCTS-AHD organizes candidate heuristics and their modification histories into a search tree, using Monte Carlo tree search to allocate exploration resources so that temporarily weaker but potentially promising branches can undergo further development \cite{mcts}. HSEvo introduces quantitative diversity metrics and integrates harmony search with genetic algorithms to better balance exploration and exploitation in heuristic search \cite{hsevo}. Moving beyond single-objective and single-heuristic design, MEoH formulates heuristic evolution as a multi-objective problem, using a dominance--dissimilarity mechanism to evolve non-dominated heuristics that trade off solution quality against efficiency \cite{meoh}; EoH-S extends single-heuristic generation to automated heuristic set design, evolving compact sets of complementary heuristics that generalize across diverse instance distributions \cite{eohs}; MuEvo extends the design target to ensembles of interdependent heuristics, coordinating the evolution of component populations through dynamic component management and ensemble-level feedback, with evaluations on selection hyper-heuristics and componentized ant colony optimization \cite{muevo}; and ATLAS addresses full-algorithm generation without a prescribed internal scaffold, using program embeddings to guide quality-diversity search and preserve different algorithmic structures alongside high-quality candidates \cite{atlas}.

A further line of work moves beyond treating the LLM as a static generator and expands search to complete optimization algorithms and general programs. EvoTune integrates evolutionary program search with reinforcement learning, updating the LLM policy using execution feedback from discovered heuristics \cite{evotune}, while CALM combines prompt-level guidance with reinforcement-learning fine-tuning to co-evolve candidate algorithms and the underlying language model \cite{calm}. Meanwhile, LLaMEA combines code generation, performance evaluation, and evolutionary selection to automatically construct and improve complete black-box metaheuristics, evaluated on continuous optimization benchmarks \cite{llamea}; AlphaEvolve applies LLM-based code modification, automated evaluation, and evolutionary selection to mathematical, computational, and algorithmic discovery \cite{alphaevolve}; and OpenEvolve provides a configurable open-source implementation of program evolution, allowing different models to iteratively optimize a supplied program using an evaluator \cite{openevolve}. Beyond these frameworks, a growing body of work further advances LLM-based AAD along complementary directions, including niching-based search partitioning, neighborhood search, performance-prediction-assisted prompting, instance-specific heuristic generation and selection, parallel algorithm portfolio evolution, and systematic surveys of the field \cite{partevo,lhns,hercules,instspechh,papllm,mego,papbin,llm4adsurvey}.

\subsubsection{Problem- and algorithm-specific design}

For specific optimization problems, researchers integrate LLM-generated heuristics with existing solver structures to improve specialized algorithms. In vehicle routing, AILS-AHD evolves ruin heuristics within adaptive iterated local search and introduces an LLM-based acceleration mechanism to improve performance on moderate- and large-scale CVRP instances \cite{ailsahd}; PyVRP$^{+}$ uses a reason--act--reflect cycle and presupplied domain knowledge to diagnose failures and evolve core components of hybrid genetic search \cite{pyvrpplus}; and VRPAgent embeds generated problem-specific operators in large neighborhood search, refining them through genetic search for capacitated, time-window, and prize-collecting vehicle routing variants \cite{vrpagent}. In production scheduling, LSH combines LLM generation, benchmark evaluation, and evolutionary search to automatically generate effective heuristics for flow-shop, job-shop, and open-shop scheduling \cite{lsh}, while IG-DOE cooperatively evolves a destruction-operator ensemble within iterated greedy search for permutation flow-shop scheduling \cite{igdoe}. In Boolean satisfiability solving, AutoSAT uses LLMs to optimize heuristic functions within conflict-driven clause-learning solvers \cite{autosat}; AutoModSAT further combines solver modularization, automatic prompt optimization, presearch, and evolutionary search to improve heuristic discovery in complex solvers \cite{automodsat}; and AESAT adopts a memetic framework combining individual improvement with population exploration, using self-optimized prompts to continually evolve solver heuristics \cite{aesat}. To address dependencies among algorithmic components, G-LNS jointly evolves destroy and repair operators for large neighborhood search, using cooperative evaluation to identify effective operator pairs for TSP and CVRP \cite{glns}; E2OC targets multi-objective evolutionary algorithms, exploring combinations of interdependent operators through Monte Carlo tree search and operator rotation \cite{e2oc}. HeurEvo additionally extends automated design to hybrid algorithms that combine heuristics with mathematical programming solvers. It jointly evolves high-level algorithmic plans, code implementations, and reusable components, optimizing component combinations and runtime allocation across stages, thereby extending the design scope to solver invocation, the organization of algorithmic stages, and computational resource allocation \cite{heurevo}.

\subsection{Related benchmarks and study positioning}

\paragraph{CO-Bench.}
CO-Bench evaluates algorithm development across 36 combinatorial optimization problems, including routing, scheduling, packing, and cutting \cite{cobench}. Candidates are generated through a predefined \texttt{solve} interface and iteratively refined through code generation and evaluation feedback, with a ten-second, single-CPU solving limit per test instance. Although this interface can accommodate a complete algorithm, the setting does not directly examine how an agent understands a mature solver implementation, autonomously identifies modification targets, and coordinates changes across components. Our study further investigates this codebase-level capability for autonomous algorithm development.

\paragraph{HeuriGym.}
HeuriGym extends heuristic generation to nine scientific and engineering problems in electronic design automation, compilers, computational biology, and logistics \cite{heurigym}. Models construct complete programs from problem specifications and improve them using execution and verification feedback. Its Quality-Yield Index jointly measures feasible-solution yield and solution quality relative to expert implementations. The per-instance solving limits range from 10 to 300 seconds, with most tasks limited to ten seconds. Our experiments further examine large-scale discrete optimization under substantially longer solving budgets. In this setting, algorithmic competitiveness depends on both initial solution quality and the ability to make further progress through sustained search.

\paragraph{ALE-Bench.}
ALE-Bench contains 40 AtCoder Heuristic Contest tasks and a ten-task lite subset, supporting complete algorithm development, test-run feedback, and visualization \cite{alebench}. Its evaluations include sequential refinement, the general coding agent OpenHands, and the specialized ALE-Agent, with system performance contextualized against human contestants. The reported iterative experiments generally allow up to four hours of development, while individual solver executions are typically limited to two to ten seconds. It therefore provides a close precedent for agent-driven algorithm engineering. Our study emphasizes large-scale instances from classical combinatorial optimization and an industrial optimization competition, with longer solver-execution budgets and direct comparisons against established high-performing specialized algorithms.

\paragraph{FrontierOR.}
FrontierOR evaluates algorithm construction on 180 literature-grounded optimization tasks, including large-scale routing, scheduling, and packing problems \cite{frontieror}. It standardizes the optimization-solver environment around Gurobi and evaluates feasibility, solution quality, and runtime against expert-verified Gurobi reference implementations, while also permitting custom decomposition, local-search, and hybrid algorithms. Its one-hour solving horizon and feedback-driven program evolution overlap with our interests. The principal differences concern the starting point of algorithm development and the performance reference: FrontierOR constructs programs from problem descriptions and evolves model-generated seeds, whereas we use established specialized algorithms and high-performing optimization-competition implementations as references, considering both development from scratch and improvement of existing code.

Building on these studies, this report focuses on large-scale, structurally complex, and practically motivated discrete optimization, evaluating agents' algorithm-design capabilities through direct comparison with strong human-designed algorithms. We select PFSP and TSP, two classical CO problems with a long research history, whose extensive body of algorithmic knowledge and high-performance specialized solvers provide demanding references for assessing algorithmic competitiveness \cite{ig,eax}. We also include industrial problems from the ROADEF/EURO Challenge. Jointly organized by ROADEF and EURO in collaboration with industrial partners, the competition presents complex optimization problems arising from practical production and operational needs. The 2018 problem, proposed by Saint-Gobain, concerns two-dimensional cutting of glass sheets with defects, requiring guillotine cuts, production-order and stack constraints, and minimization of material loss \cite{roadef}. The 2020 problem, proposed by the French electricity transmission system operator RTE, concerns grid maintenance planning, requiring resource and intervention-exclusion constraints to be satisfied while controlling grid-operation risk across uncertain scenarios \cite{roadef2020}. The 2022 Renault challenge combines shipment timing, stacking, and truck loading, with a joint transport and inventory objective~\cite{r22subject}. Our held-out PFSP instances contain 300--800 jobs, and the TSP instances contain 21,215--47,608 cities; TSP and ROADEF 2018 allow one hour per held-out solve, ROADEF 2020 allows 15 minutes, and ROADEF 2022 allows 15 or 25 minutes. These settings require algorithms to handle complex constraints and make sustained, effective use of computational resources in large search spaces.

The challenge of this study also stems from its performance references. We compare agent-developed algorithms directly with human expert methods such as GA-EAX and a ROADEF 2018 champion-derived implementation \cite{eax,r18solver}, and measure the remaining gap on problems with best-known objective values. Producing executable code or finding a feasible solution is therefore only the beginning: an algorithm must approach or advance established high-quality results within the given solving budget. Under this objective, development from scratch examines whether an agent can autonomously discover and implement competitive algorithmic structures, while improvement of existing code examines whether it can understand a mature solver and identify and implement valuable improvements. Together, these settings investigate whether agents can translate code-generation and tool-use capabilities into practical algorithmic competitiveness on challenging discrete optimization problems.

\section{Comparative Study Design}

\label{sec:design}

This chapter describes the task settings and experimental methodology of the comparative study, providing a common basis for the subsequent analysis. We first define two task scenarios grounded in practical algorithm-development needs: improvement of existing algorithms and development from scratch. We then introduce the unified experimental framework, specifying the materials supplied to agents, their workspace and information-access boundaries, and the procedures for autonomous development, feedback-based evaluation, and independent testing. Building on this framework, we present formal definitions of the five benchmark problems, their instance splits, and the human reference algorithms and workflow baselines. We further specify the computational budgets for development and solver execution, repeated-run protocols, and execution settings. Finally, we define each problem's primary solution-quality metric and additional measures such as solving success rate, instance coverage, and reference-attainment counts, enabling comparisons across development scenarios, methods, and models under clearly specified evaluation conditions.

\subsection{Task scenarios}
\label{sec:development_tasks}

\paragraph{Improvement of an existing algorithm.}
In practice, an organization often already has a solver, or can start from a published implementation. Further development may be needed to obtain better solutions, reduce computational cost, handle larger instances, or adapt the implementation to a particular operating environment. This task supplies the problem materials and development data, together with an identified initial implementation and its build instructions. The agent investigates the code, chooses modification targets, and evaluates changes. Depending on the task, useful changes may involve search operators, initialization, diversification, scheduling of algorithmic stages, data structures, or implementation efficiency. The deliverable includes the modified program and the instructions needed to build, run, and validate it.

\paragraph{Development from scratch.}
A practitioner facing a new optimization requirement may have a problem specification and representative historical instances, but no solver that can be directly deployed. For example, a new cutting application may require a program that reads the production data, represents cutting patterns, respects manufacturing constraints, and outputs an executable plan. Our scratch-development task models this situation. The agent receives the problem definition, input/output specifications, development instances, and validation tools. It must choose data structures and algorithmic strategies, implement the solver, and improve it using development feedback. The final deliverable is a self-contained, reproducible program that handles instances in the specified format. No complete reference solver is supplied; the algorithms and knowledge available through model pretraining remain part of the agent's capability.

Both tasks reflect practical algorithm engineering. They share the objective, data-access rules, and final evaluation procedure; their principal distinction is the supplied algorithmic starting point. This makes it possible to investigate both the construction of competitive algorithms and the use of existing expert knowledge. Task-specific requirements, such as rebuilding candidates from a common frozen baseline, are stated in the instruction file and retained in the experiment record. Both scenarios are evaluated using the development and held-out protocols described below.

\subsection{Unified experimental framework}
\label{sec:unified_framework}

\begin{figure}[!htbp]
\centering
\includegraphics[width=\linewidth]{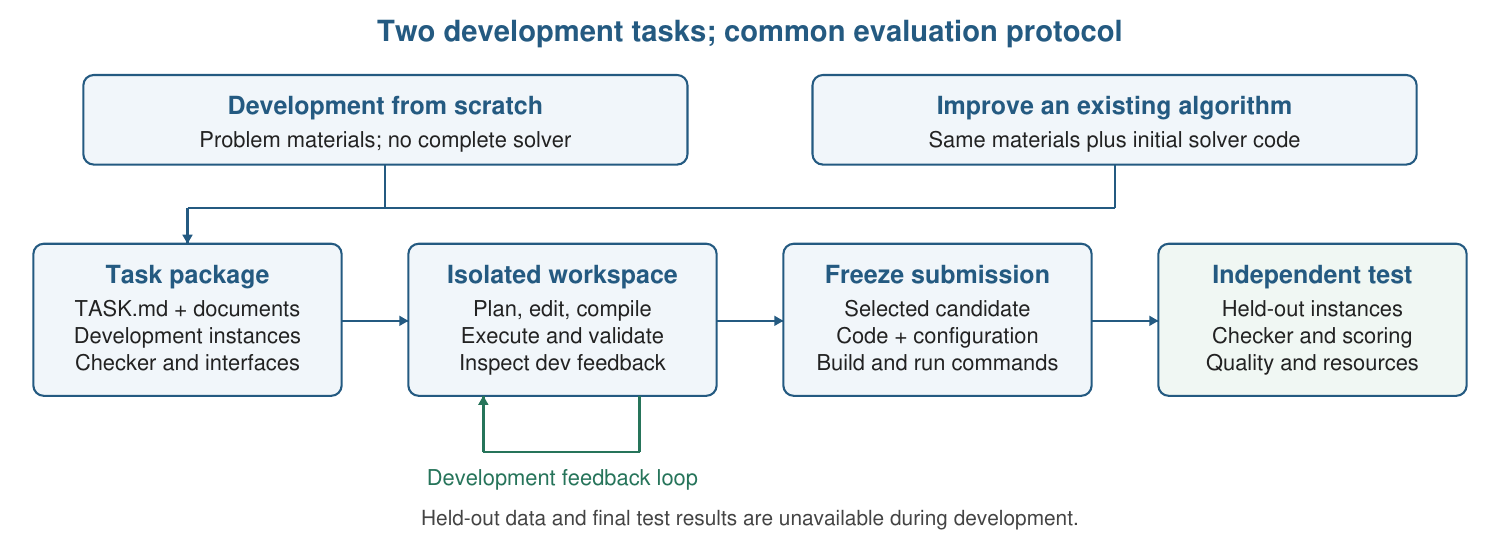}
\caption{Common experimental framework. The two tasks differ in the initial code supplied. Agent development uses authorized materials and development feedback; the frozen program is assessed independently on held-out instances.}
\label{fig:evaluation_framework}
\end{figure}

The framework separates agent development from independent final testing. A task package specifies the materials an agent may access, the work it must complete, and how its output will be evaluated. The common organization is illustrated in Figure~\ref{fig:evaluation_framework}. The framework standardizes task delivery, access boundaries, validation, and final testing; numerical budgets and implementation requirements are specified separately for each problem.

\paragraph{Task package.}
The instruction file (\texttt{TASK.md}) describes the objective, permitted actions, resource allowances, and submission requirements. Supporting materials contain the problem statement, input/output documentation, training/development instances, and the applicable checker and evaluation entry points. The improvement task additionally includes the initial solver and build dependencies. Reference documents and original data remain unchanged, while candidate implementations, dependencies, temporary files, and logs are kept in the experiment's working directory. Here, the training set denotes the instances used during algorithm development; it does not imply updating the LLM's parameters.

\paragraph{Workspace and information isolation.}
Account-based runs use a dedicated operating-system account and a task-specific directory, with filesystem permissions restricting access to the authorized materials. The agent may use the available editor, shell, compiler, interpreter, and approved libraries to carry out the task. Held-out instances, final test results, and other agents' work directories are excluded from development access. Scratch packages also withhold existing solver implementations. An example task document and the corresponding access requirements are provided in Appendix~\ref{app:task-example}. Filesystem isolation and resource monitoring are separate mechanisms.

\paragraph{Autonomous development and feedback.}
Within the permitted workspace, the agent reads the task and documentation, implements or modifies a candidate, and performs compilation, execution, and validation. Development evaluations return the permitted feedback, including run status, checker diagnostics, objective values or aggregate scores, and available resource measurements. A complete evaluation runs one fixed candidate across all development instances and prescribed repeats. Partial tests and debugging runs are recorded separately. The checker verifies feasibility and independently recomputes the objective where applicable; its result and the optimization score serve different purposes. Evaluation tools and reference inputs are protected from modification.

\paragraph{Submission and independent testing.}
The final candidate is selected using development information, then frozen together with its configuration, build/run commands, and source identity. A separate evaluator runs that candidate on held-out instances and checks the resulting solutions. Test scores are used to assess the frozen solver, rather than to continue selecting or tuning it. This common sequence supports both development tasks while preserving a clear boundary between algorithm construction and assessment.

\subsection{Problems, evaluated systems, and baselines}
\label{sec:benchmark_instances}

We study two classical combinatorial optimization problems, PFSP and TSP, and three industrial problems from the 2018, 2020, and 2022 ROADEF/EURO Challenges. Below, we first define their feasible solutions and original objectives, then introduce the evaluated coding agents, human reference solvers, and workflow baselines. Section~\ref{sec:performance_metrics} specifies how these objectives are converted into the reported experimental metrics.

\subsubsection{Problem definitions and instances}
\label{sec:problem_definitions}

\paragraph{Permutation flow-shop scheduling.}
An instance contains $n$ jobs and $m$ machines, with processing time $p_{j\ell}$ for job $j$ on machine $\ell$~\cite{pfspreview}. Every job visits the machines in the same order, every machine handles one job at a time, and processing is non-preemptive. A solution is a permutation $\pi=(\pi_1,\ldots,\pi_n)$ used on all machines. The completion time of the job in position $k$ on machine $\ell$ is
\begin{equation}
C_{k\ell}=\max\{C_{k-1,\ell},C_{k,\ell-1}\}+p_{\pi_k,\ell},
\qquad C_{0\ell}=C_{k0}=0.
\label{eq:pfsp_completion}
\end{equation}
The objective is to minimize the makespan $f_{\mathrm{PFSP}}(\pi)=C_{nm}$, the time at which all jobs finish. The checker verifies the permutation and independently recomputes this value. We use instances from the VFR benchmark~\cite{vfr}: development uses 100- and 200-job instances, and testing uses larger job counts with the same machine-count choices.

\paragraph{Traveling salesman problem.}
An instance of the symmetric Euclidean TSP consists of $n$ cities and pairwise distances $d_{uv}=d_{vu}$ obtained from their coordinates under the instance's distance convention~\cite{tspbook}. A feasible solution is a cyclic permutation visiting every city exactly once. For $\pi_{n+1}=\pi_1$, the objective is the total tour length
\begin{equation}
f_{\mathrm{TSP}}(\pi)=\sum_{k=1}^{n}d_{\pi_k,\pi_{k+1}}.
\label{eq:tsp_length}
\end{equation}
Independent validation checks the complete permutation and recomputes the length using the specified distance and rounding rules. Drawn from large-scale TSPLIB, National, and VLSI collections~\cite{tsplib,nationaltsp,vlsitsp}, development instances contain 5,087--19,402 cities, whereas the held-out instances contain 21,215--47,608 cities. These sizes require both effective search and efficient distance and tour operations.

\paragraph{ROADEF/EURO 2018 glass cutting.}
Given an ordered sequence of glass plates of width $W$ and height $H$, with plate-specific defects, the task is to cut a prescribed item set $I$ with dimensions $(w_i,h_i)$. A feasible cutting plan $P$ extracts every item exactly once, avoids item overlap and defects, and satisfies plate order, within-stack item order, staged guillotine-cutting rules, and the associated manufacturing constraints~\cite{roadef}. Let $K(P)$ be the number of used plates and $\rho(P)$ the width of the reusable residual to the right of the final first-stage cut on the last plate. With total item area $A=\sum_{i\in I}w_i h_i$, the original objective is to minimize the loss area
\begin{equation}
L(P)=K(P)WH-H\rho(P)-A.
\label{eq:r18_loss}
\end{equation}
Only the specified last-plate residual is deducted as reusable material. Thus, consumed area is $A+L(P)$; Section~\ref{sec:performance_metrics} normalizes this loss to a waste percentage. Development uses B1--B15, and testing uses X1--X15. Feasibility and cutting-plan quantities are checked using the official tools.

\paragraph{ROADEF/EURO 2020 grid maintenance planning.}
The RTE problem schedules a set of maintenance interventions over periods $t=1,\ldots,T$~\cite{roadef2020}. A solution assigns each intervention $i$ a permitted start time $s_i$; its duration, resource use, and scenario risks depend on that choice. All interventions must finish within the horizon without interruption. At each period, resource use must lie between its prescribed lower and upper bounds, and intervention pairs subject to seasonal exclusions must not overlap during the specified periods. These conditions define the feasible schedule set $\mathcal F_{20}$.

Let $\Omega_t$ be the scenarios in period $t$ and $r_{t\omega}(\mathbf s)=\sum_i r_{it\omega}(s_i)$ the total scenario risk of schedule $\mathbf s$. Define its mean risk $\mu_t(\mathbf s)=|\Omega_t|^{-1}\sum_{\omega\in\Omega_t}r_{t\omega}(\mathbf s)$ and empirical $\tau$-quantile $q_{t,\tau}(\mathbf s)$. The latter is the $\lceil\tau|\Omega_t|\rceil$-th smallest scenario risk. The official objective, minimized over $\mathbf s\in\mathcal F_{20}$, is
\begin{equation}
f_{\mathrm{R20}}(\mathbf s)
=\frac{1}{T}\sum_{t=1}^{T}\left[\alpha\mu_t(\mathbf s)
+(1-\alpha)\max\{0,q_{t,\tau}(\mathbf s)-\mu_t(\mathbf s)\}\right].
\label{eq:r20_objective}
\end{equation}
The instance supplies $\alpha$ and $\tau$. The objective balances average risk with the positive excess of the quantile above the period mean. Development uses C01--C15, and testing uses X01--X15. The official checker verifies feasibility and recomputes the objective from all supplied scenarios.

\paragraph{ROADEF/EURO 2022 truck loading.}
Renault's inbound logistics problem couples the assignment of automotive parts to trucks with their three-dimensional physical loading~\cite{r22subject}. A feasible plan $S\in\mathcal F_{22}$ specifies how items are grouped into stacks, which planned or additional trucks carry each stack, and where each stack is placed on the truck floor, satisfying item coverage, delivery time windows, stack compatibility, multi-stop loading order, geometric support, and dynamic axle-load limits. Consolidating freight can reduce the number of trucks required while delivering parts early and increasing plant inventory costs. Let $C_{\mathrm{trans}}(S)$ denote the instance-weighted transportation cost (including planned and extra trucks) and $C_{\mathrm{inv}}(S)$ the cumulative inventory cost for early deliveries. The official objective minimizes the total cost over $S\in\mathcal F_{22}$:
\begin{equation}
f_{\mathrm{R22}}(S) = C_{\mathrm{trans}}(S) + C_{\mathrm{inv}}(S).
\label{eq:r22_objective}
\end{equation}
Development uses the 50 final-phase C instances, and independent testing uses the 105 X instances; identically named instances across splits correspond to distinct data files and reference costs. Feasibility and objective values are verified by the official Java checker.

\begin{table}[htbp]
\centering\small
\caption{Development and held-out instances. Instance counts refer to distinct problems before repeated solver execution.}
\label{tab:instances}
\begin{tabularx}{\linewidth}{@{}p{0.12\linewidth}YY@{}}
\toprule
Problem & Development set & Held-out set \\
\midrule
PFSP & 60 instances; $n\in\{100,200\}$ jobs, $m\in\{20,40,60\}$ machines & 180 instances; $n\in\{300,400,500,600,700,800\}$, same machine counts; ten instances per size combination \\
TSP & 33 instances; 5,087--19,402 cities & 21 distinct instances; 21,215--47,608 cities \\
ROADEF 2018 & B1--B15; 15 instances & X1--X15; 15 instances \\
ROADEF 2020 & C01--C15; 15 instances & X01--X15; 15 instances \\
ROADEF 2022 & C, 50 instances (25 short, 25 long) & X, 105 instances (49 short, 56 long) \\

\bottomrule
\end{tabularx}
\end{table}

The development set supplies the feedback used for algorithm design and candidate selection. The held-out set is reserved for evaluating the frozen final solver. To preserve strict train--test separation and evaluation validity, the reported TSP comparison excludes instances overlapping with the development set as well as invalid historical runs (see Appendix~\ref{app:instances}). Detailed instance identifiers and record handling are retained in the appendices.

\subsubsection{Evaluated systems and baselines}
\label{sec:baseline_methods}

\paragraph{Evaluated coding agents.}
We evaluate four general-purpose coding agents paired with frontier LLMs: \textbf{Codex} (\texttt{GPT-6-Astra}, abbreviated \textbf{GPT}), \textbf{Kimi-code} (\texttt{KIMI-K3}, abbreviated \textbf{KIMI}), \textbf{Zcode} (\texttt{GLM-5.3}, abbreviated \textbf{GLM}), and \textbf{DeepSeek Harness} (\texttt{DeepSeek-V4-Flash}, abbreviated \textbf{DeepSeek}). Each agent interacts autonomously with its isolated workspace via shell, editor, compiler, and evaluation tools without external evolutionary orchestration. Kimi-code, Zcode, and DeepSeek Harness are evaluated across all five benchmark problems in both development scenarios (existing-algorithm improvement and development from scratch), whereas Codex is evaluated on the PFSP and TSP improvement tasks. Exact agent releases, full model identifiers, and reasoning-effort settings are documented in Appendix~\ref{app:environment}.

\paragraph{Human reference solvers.}
Each domain has a human-designed reference, used both as the starting point for existing-code improvement and as a quality reference for scratch development. No complete human solver is supplied to the scratch agents. The references are:
\begin{itemize}
\item \textbf{PFSP: Original IG and Basic ALNS.} The supplied IG follows the iterated-greedy algorithmic family of Ruiz and St\"utzle~\cite{ig}, combining construction, destruction and reinsertion, local search, and acceptance. A basic adaptive large neighborhood search implementation provides a second human-designed control. Both references are evaluated under the corresponding training and test budgets and shared across the two development scenarios. Basic ALNS denotes the tested implementation and configuration.
\item \textbf{TSP: GA-EAX Large.} The supplied solver uses Nagata and Kobayashi's edge assembly crossover within a genetic search~\cite{eax}. We report the original population-300 configuration and a population-100 control, both with 30 offspring. The population control uses the same underlying executable and isolates the effect of this configuration change. Population-100 EAX also serves as the configured human reference in the scratch comparison.
\item \textbf{ROADEF 2018: the champion-derived solver.} This implementation is based on the tree-search approach associated with Libralesso and Fontan~\cite{r18solver}. It supplies the expert cutting-search structure for the improvement task; its original version is the human reference in both development scenarios.
\item \textbf{ROADEF 2020: J49 ALNS.} The unmodified J49 adaptive large neighborhood search implementation~\cite{r20j49} is the human reference. Its search constructs and improves maintenance schedules through destroy-and-repair operations. It is supplied in the improvement task, where agents can modify the algorithm and its implementation autonomously.
\item \textbf{ROADEF 2022: S46--HiGHS V6.} We adopt the public sixth-place S46 matheuristic solver~\cite{r22s46,r22final}, ported to the open-source HiGHS mixed-integer solver with output-completeness fixes and frozen as V6. It serves both as the starting implementation supplied to all improvement agents and as the measured human reference across both development scenarios (noting that its costs can differ slightly from team S46's original competition submission).
\end{itemize}
Reference and candidate solvers follow the corresponding instance sets, repeated-run protocols, and external budgets in Section~\ref{sec:execution_settings}. Recorded execution differences are retained in the appendices.

\paragraph{Workflow methods.}
We include two representative LLM-driven workflow baselines in the existing-code improvement setting, both powered by the same DeepSeek model to enable direct same-model comparisons against DeepSeek Harness:
\begin{itemize}
\item \textbf{ReEvo.} ReEvo combines evolutionary search with reflective language feedback to generate and improve heuristics~\cite{reevo}. Evaluated on PFSP, our adaptation evolves a designated destruction operator while keeping the surrounding IG implementation fixed, representing a component-level AHD workflow.
\item \textbf{OpenEvolve.} OpenEvolve (v0.4.0) is an open-source program-evolution framework that allows changes across the supplied solver program through an external evaluator~\cite{openevolve}. Evaluated on PFSP, TSP, ROADEF 2018, and ROADEF 2020 with a 50-evaluation allowance, it selects the best fully valid training candidate on classical tasks and uses a feasibility-first evaluator (freezing the earliest top-scoring candidate, with compilation failures logged separately) on the industrial tasks.
\end{itemize}
Method and model are reported separately in all result tables, and editable scopes and actual development expenditure are documented with the corresponding records.

\subsection{Computational budgets and execution settings}
\label{sec:execution_settings}

\paragraph{Development allowance.}
The ROADEF campaigns allow 50 full evaluations per agent in each development scenario. Earlier PFSP improvement campaigns have different recorded stopping points; Appendix~\ref{app:all-trajectories} shows the first 50 evaluations for comparison. One complete evaluation uses a fixed candidate across every development instance and its prescribed repeats.

In the ROADEF 2020 improvement task, the initial evaluation of the unmodified ALNS baseline also counts toward this budget. Debugging, partial screening, failed evaluations, and early termination are recorded separately so that evaluation counts can be interpreted alongside actual expenditure. Equal nominal evaluation allowances do not imply equal development wall time, model calls, or total solver computation.

\paragraph{Per-instance solving budget.}
This budget limits execution of the generated algorithm on one instance. Different problems use allowances suited to their scale and computational characteristics; within a given problem and evaluation phase, candidate and reference solvers are compared under the same budget. PFSP and TSP use single-thread solver processes, with concurrency across instances increasing evaluation throughput. All three ROADEF problems permit multiple logical CPUs within one solve. Inter-instance concurrency and within-solve parallelism are recorded separately. Table~\ref{tab:budgets} summarizes the settings.

\begin{table}[htbp]
\centering\small
\caption{Development and held-out solver budgets. A full evaluation runs one fixed candidate over every instance and prescribed repeat.}
\label{tab:budgets}
\begin{tabularx}{\linewidth}{@{}p{0.12\linewidth}YYY@{}}
\toprule
Problem & Development evaluation & Held-out evaluation & Execution setting \\
\midrule
PFSP & $60\times10=600$ solves; $0.03nm$ seconds per solve & $180\times10=1{,}800$ solves; $0.06nm$ seconds per solve & Single-thread solver processes; up to 400 concurrent solves \\
TSP & $33\times10=330$ solves; 900 seconds for $n\leq10{,}000$, 1,200 otherwise & $21\times10=210$ solves; 3,600 seconds per solve & Shared clock includes preprocessing and initialization; at most 5 development or 10 test restarts \\
ROADEF 2018 & $15\times3=45$ solves; nominal 900 seconds per solve & $15\times3=45$ solves; 3,570 internal seconds within a 3,600-second external budget & Up to 8 logical CPUs and 48 GB per solve; 15 parallel instances, three sequential repeat batches \\
ROADEF 2020 & $15\times3=45$ solves; 900 seconds total wall time per solve & $15\times3=45$ solves; 900 seconds total wall time per solve & Up to 4 logical CPUs (2 physical cores and their SMT threads) and 16 GB per solve; 15 parallel instances, three sequential repeat batches \\
ROADEF 2022 & $50\times1$ during development; $50\times3$ in frozen confirmation & $105\times3$; 900 s (short) or 1,500 s (long) on both splits & 8 logical CPUs and 30 GB per solve; 25 concurrent during development \\

\bottomrule
\end{tabularx}
\end{table}

\paragraph{Repeated execution and timing.}
PFSP and TSP use ten independent runs per instance (TSP seeds 1111--1120). ROADEF 2018 and 2020 use three runs per instance executed in three consecutive batches, with each batch processing 15 instances concurrently before the next batch begins. For ROADEF 2022, development evaluations use one run per C instance across the 50-round allowance, while frozen training confirmations and held-out X evaluations run three independent seeds (1111, 2222, and 3333) on disjoint CPU groups with independent solution checking and process-group memory sampling under fixed per-solve resource limits. Across all problems, candidate source code, executables, and parameters remain fixed throughout an evaluation, and each solve starts independently without reusing prior search state; repeated solver runs measure execution variability of the frozen program rather than independent repetitions of the agent-development campaign.

TSP restarts share one total clock, including preprocessing, initialization, and all searches. ROADEF 2020's 900-second budget includes input reading, algorithm computation, output, and exit, so programs must reserve time for orderly termination; compilation and independent checker costs are recorded separately. CPU and memory allowances apply to a solve and all its subprocesses. For ROADEF 2020, CPU allocation follows the physical-core and SMT topology rather than thread count alone.

Resource monitoring and solution-feasibility checking are recorded separately. Account isolation does not constitute CPU or memory hard limits; in passive-monitoring experiments, the stated allowances are protocol requirements. Retained execution measurements and computing-environment information are provided in Appendices~\ref{app:execution} and~\ref{app:environment}.

\subsection{Main experimental metrics}
\label{sec:performance_metrics}

The main experiments report objective quality together with solution validity and reference-attainment statistics. The original objectives are defined in Section~\ref{sec:problem_definitions}; the following metrics specify how each checked solution and its repeated runs enter the result tables.

\subsubsection{Primary solution-quality metrics}

\paragraph{PFSP and TSP: gap to the fixed best-known reference.}
Let $f_{ir}$ denote the checked objective of instance $i$ on repeat $r$, namely the PFSP makespan or TSP tour length, and let $b_i$ denote its fixed benchmark best-known reference. For PFSP, $b_i$ is the supplied benchmark upper bound (UB); for TSP, it is the archived optimal or best-known tour length (BKS). We compute
\begin{equation}
g_{ir}=100\,\frac{f_{ir}-b_i}{b_i}.
\label{eq:gap}
\end{equation}
Gap is a percentage and lower is better: zero matches the reference, a positive value is worse, and a negative value improves on it. References remain fixed throughout the comparison, including after a run beats them. An upper bound or BKS need not be a proven optimum.

\paragraph{ROADEF 2018: waste percentage.}
For a valid cutting plan $P_{ir}$, let $L(P_{ir})$ be the loss area from Equation~\ref{eq:r18_loss} and $A_i$ the instance's total item area. We report
\begin{equation}
w_{ir}=100\,\frac{L(P_{ir})}{A_i+L(P_{ir})}.
\label{eq:waste}
\end{equation}
The denominator is consumed plate area after deducting the reusable residual on the last plate. Waste percentage measures the proportion of consumed material not recovered as required items; lower is better. It is not a gap to the champion or BKS. Changes in percentage points and relative percentage reductions are distinguished.

\paragraph{ROADEF 2020: official objective value.}
We report the official risk objective $f_{\mathrm{R20}}(\mathbf s_{ir})$ from Equation~\ref{eq:r20_objective}, recomputed by the checker, as the primary quality measure. For paired workflow comparisons, we also compute per-instance percentage gaps to the best published contest objective across the 15- and 90-minute settings~\cite{r20final} and average them with equal weights. Lower values indicate better schedules under the instance's risk model. Raw objective magnitudes are compared between methods on the same instance set; training and test sets contain different instances, so their numerical levels do not directly measure generalization loss.

\paragraph{ROADEF 2022: gap to the final-phase reference.}
For split $s$ and instance $i$, let $b_{si}$ be the lowest cost among positively scored entries for that instance in the organizers' final-phase results workbook~\cite{r22final}. A valid solve of cost $f_{ir}$ is evaluated by the same percentage gap $g_{ir}=100\,(f_{ir}-b_{si})/b_{si}$ as Equation~\ref{eq:gap}, where lower values are better and a negative value improves on the recorded contest reference. Repeated-run aggregation follows Equation~\ref{eq:quality_aggregation}, and both full-set and common-subset scores are reported alongside valid-solve and complete-instance counts.

\paragraph{Mean and best over repeated runs.}
Let $q_{ir}$ denote the reported per-run quality: $g_{ir}$ for PFSP, TSP and ROADEF 2022, $w_{ir}$ for ROADEF 2018, and $f_{\mathrm{R20}}(\mathbf s_{ir})$ for ROADEF 2020. For $N$ instances with $R$ prescribed valid repeats each, we report
\begin{equation}
Q_{\mathrm{mean}}=\frac{1}{N}\sum_{i=1}^{N}\frac{1}{R}\sum_{r=1}^{R}q_{ir},
\qquad
Q_{\mathrm{best}}=\frac{1}{N}\sum_{i=1}^{N}\min_{1\le r\le R}q_{ir}.
\label{eq:quality_aggregation}
\end{equation}
These appear as Mean/Best gap, Mean/Best waste, or Mean/Best objective. Each instance has equal weight. Mean summarizes average repeated-run quality; Best selects the best prescribed repeat for each instance and then averages those values. It therefore describes the quality reached by choosing among the repeated runs, not the single best score over the entire dataset. PFSP and TSP use ten repeats per instance, and the frozen evaluations for all three ROADEF problems use three.

\subsubsection{Additional information in the result tables}

\paragraph{Method, model, and reference groups.}
Method identifies the development system or human solver, while Model identifies the underlying LLM. Human reference, Workflow methods, and Coding agents distinguish the sources of the compared algorithms. A dash in Model indicates that the reference solver uses no LLM. Keeping these columns separate enables comparison of development methods using the same model and of the evaluated agent--model combinations.

\paragraph{Valid solves and solving success rate.}
For $N$ instances and $R$ prescribed repeats, let $v_{ir}=1$ if the run produces a checker-valid solution and $0$ otherwise. The Valid solves entry is $V/(NR)$, where $V=\sum_i\sum_r v_{ir}$; its percentage form, $100V/(NR)$, is the solving success rate. This measures repeated-run reliability. PFSP requires a complete job permutation and verified makespan, TSP a complete tour and verified length, and ROADEF a solution accepted by the official checker. Compilation failures, missing outputs, and invalid solutions do not count as successful solves. Resource compliance is recorded separately: checker validity alone does not establish compliance with the time or memory allowance.

\paragraph{Valid and complete instances.}
When repeated runs can fail, instance-level reliability is reported alongside Valid solves. For ROADEF 2020, \emph{Valid instances} reports $C_{\mathrm{valid}}/N$, where $C_{\mathrm{valid}}=\sum_i\mathbf 1\{\sum_r v_{ir}\geq1\}$ counts instances with at least one checker-valid schedule (measuring instance coverage). For ROADEF 2022, \emph{Complete instances} reports $C_{\mathrm{comp}}/N$, where $C_{\mathrm{comp}}=\sum_i\mathbf 1\{\sum_r v_{ir}=R\}$ counts instances on which all $R$ repeats produce valid solutions. Both are distinct from Valid solves: for instance, 15/15 valid instances does not imply 45/45 valid solves.

\paragraph{Below UB and BKS hits.}
PFSP's Below UB counts individual valid runs with $f_{ir}<b_i$, equivalently $g_{ir}<0$. It measures how often a program improves on the fixed supplied bound. TSP's BKS hits counts individual valid runs with $f_{ir}=b_i$, equivalently zero gap, measuring how often the reference tour length is reached. These are run counts, not counts of distinct instances. Their frequencies can differ from mean-gap rankings: a method can reach the reference more often yet have larger gaps on the remaining runs. An unavailable count is shown as a dash, not zero.

\paragraph{Wins, ties, and losses.}
Where the text reports pairwise win/tie/loss counts, methods are compared by their mean quality on each instance over the prescribed repeats. A win means a lower mean than the reference, a tie an equal mean under the recorded comparison precision, and a loss a higher mean. These counts show how broadly an aggregate gain is distributed across instances, but not the magnitude of each gain or a statistical significance level.

\paragraph{Full-set, common-subset, and incomplete results (NC).}
In tables where some solvers fail on a subset of runs (ROADEF 2020 and ROADEF 2022), results are grouped under \emph{Full set} and \emph{Common subset} columns. A \emph{Full set} Mean or Best metric averages across all $N$ instances of the split via Equation~\ref{eq:quality_aggregation} and is reported only when all $NR$ prescribed runs are valid; otherwise the entry is marked NC (not complete), while Valid solves and instance-reliability counts are always retained. Failed runs are never assigned a default score or silently dropped. To enable a fair quality comparison across solvers despite partial failures, \emph{Common subset} ($N_{\mathrm{c}}$) restricts Equation~\ref{eq:quality_aggregation} to the fixed intersection of $N_{\mathrm{c}}$ instances on which every compared solver completes all $R$ repeats validly (the exact subset and included solvers are stated in each table caption). Common-subset Mean and Best gaps or objectives therefore measure quality conditional on shared feasibility and complement, rather than replace, full-set reliability and quality scores.

\section{Experimental Results and Analysis}
\label{sec:results}

This chapter reports experimental results for coding agents in two algorithm-development scenarios. Section~\ref{sec:existing_improvement} examines their improvement of existing high-performing algorithms, and Section~\ref{sec:scratch_results} examines their ability to construct solvers from scratch. Both scenarios report training and test performance separately for each problem domain. Comparisons with human reference algorithms and, where applicable, workflow methods assess solution quality and solving success rates across agents and models. All metrics and result aggregation follow the definitions in Section~\ref{sec:performance_metrics}.

Section~\ref{sec:algorithm_findings} brings together the results in Sections~\ref{sec:existing_improvement} and~\ref{sec:scratch_results}, alongside the corresponding code changes, to discuss the main findings on coding agents' algorithm-design capabilities.

\subsection{Improving existing high-performing algorithms}
\label{sec:existing_improvement}

We first examine whether coding agents improve the supplied human-designed solvers, how they compare with workflow methods using the same model, and whether their gains persist on unseen instances. Workflow comparisons cover PFSP, TSP, ROADEF 2018 and ROADEF 2020. ROADEF 2022 compares the three coding agents with S46--HiGHS V6. Model rankings refer to the evaluated agent--model pairs. Editable scope and development expenditure follow the recorded campaigns, so framework comparisons describe their observed outcomes.

\subsubsection{Permutation flow-shop scheduling}
\label{sec:pfsp_improve}

\paragraph{Training performance.}
Table~\ref{tab:pfsptraining} reports results on 60 training instances with 100 or 200 jobs, ten repeats per instance, and a solving allowance of $0.03nm$ seconds. All four coding agents reduce the original IG's mean gap of 3.003299\% to 0.982869--1.655192\%, corresponding to reductions of 44.9--67.3\% in mean gap. They also outperform Basic ALNS, whose mean gap is 3.902482\%. Thus, the improvement is observed against both supplied human references.

The same-model comparison gives a more differentiated result. With DeepSeek, OpenEvolve achieves the lowest training mean of 1.368726\%, followed by DeepSeek Harness at 1.655192\% and ReEvo at 2.807877\%. Agent-directed development therefore improves on the ReEvo adaptation but does not match OpenEvolve in this PFSP campaign. Across the four coding-agent variants, GLM performs best, followed by KIMI, GPT, and DeepSeek; the first three also have lower means than both workflow methods, although those comparisons use different models. Each agent produces 600/600 valid solves, but none of its selected-candidate training runs improves on the fixed supplied UB.

\begin{table}[!htbp]
\centering\small
\caption{PFSP improvement: training results on 60 instances (10 repeats, $T=0.03nm$).}
\label{tab:pfsptraining}
\AppendixFitTable{\begin{tabular}{@{}llrrrr@{}}
\toprule
Method & Model & Mean gap (\%) & Best gap (\%) & Valid solves & Below UB \\
\midrule
\multicolumn{6}{@{}l}{\textit{Human references}} \\
Original IG & \textemdash & 3.003299 & 2.493410 & 600/600 & 0 \\
Basic ALNS & \textemdash & 3.902482 & 3.491333 & 600/600 & 0 \\
\midrule
\multicolumn{6}{@{}l}{\textit{Workflow methods}} \\
OpenEvolve & DeepSeek & 1.368726 & 0.961820 & 600/600 & \textemdash \\
ReEvo & DeepSeek & 2.807877 & 2.343383 & 600/600 & 0 \\
\midrule
\multicolumn{6}{@{}l}{\textit{Coding agents}} \\
Zcode & GLM & 0.982869 & 0.638629 & 600/600 & 0 \\
Kimi-code & KIMI & 1.130911 & 0.773076 & 600/600 & 0 \\
Codex & GPT & 1.296465 & 0.884800 & 600/600 & 0 \\
DeepSeek Harness & DeepSeek & 1.655192 & 1.174233 & 600/600 & 0 \\
\bottomrule
\end{tabular}}
\end{table}

\FloatBarrier
\paragraph{Test performance.}
On the 180 held-out instances with 300--800 jobs, every coding agent remains better than both human references (Table~\ref{tab:igresults}). Their mean gaps range from 0.450912\% to 1.076633\%, compared with 2.053454\% for the equal-budget original IG and 2.346119\% for Basic ALNS. The reductions relative to the original IG's mean gap range from 47.6\% to 78.0\%, showing that the developed changes remain effective at the larger test sizes and the $0.06nm$ solving allowance.

All four agents also outperform ReEvo, which obtains 1.935213\%. With the same DeepSeek model, however, OpenEvolve remains better than DeepSeek Harness: 0.910495\% versus 1.076633\%. GLM has the lowest coding-agent mean and best gap, followed by GPT, KIMI, and DeepSeek. GLM leads in 12 of the 18 size groups and wins 152/180 instance means against GPT, 147/180 against KIMI, and 180/180 against DeepSeek. KIMI leads in four large-job, 20-machine groups, while GPT leads in the 700-by-40 and 800-by-40 groups. The instance-level scores underlying these comparisons appear in Appendix~\ref{app:test-pfsp}.

The UB counts provide an additional distinction. GLM, GPT, and KIMI improve on the fixed UB in five, two, and one runs, respectively, whereas DeepSeek does so in none. GPT also moves ahead of KIMI in testing despite its worse training mean. The results therefore support transfer of the improvements while showing that development-set ranking alone does not determine the best solver for the target instances. Basic ALNS is the tested configuration, rather than an exhaustive assessment of the ALNS algorithm family.

\begin{table}[!htbp]
\centering\small
\caption{PFSP improvement: held-out results on 180 instances (10 repeats, $T=0.06nm$).}
\label{tab:igresults}
\AppendixFitTable{\begin{tabular}{@{}llrrrr@{}}
\toprule
Method & Model & Mean gap (\%) & Best gap (\%) & Valid solves & Below UB \\
\midrule
\multicolumn{6}{@{}l}{\textit{Human references}} \\
Original IG & \textemdash & 2.053454 & 1.827103 & 1800/1800 & 0 \\
Basic ALNS & \textemdash & 2.346119 & 1.926043 & 1800/1800 & 0 \\
\midrule
\multicolumn{6}{@{}l}{\textit{Workflow methods}} \\
OpenEvolve & DeepSeek & 0.910495 & 0.676979 & 1800/1800 & \textemdash \\
ReEvo & DeepSeek & 1.935213 & 1.715714 & 1800/1800 & 0 \\
\midrule
\multicolumn{6}{@{}l}{\textit{Coding agents}} \\
Zcode & GLM & 0.450912 & 0.245851 & 1800/1800 & 5 \\
Codex & GPT & 0.529947 & 0.301607 & 1800/1800 & 2 \\
Kimi-code & KIMI & 0.694289 & 0.487671 & 1800/1800 & 1 \\
DeepSeek Harness & DeepSeek & 1.076633 & 0.837990 & 1800/1800 & 0 \\
\bottomrule
\end{tabular}}
\end{table}

\FloatBarrier
\subsubsection{Traveling salesman problem}

\paragraph{Training performance.}
Table~\ref{tab:tsptraining} compares the developed EAX variants on 33 training instances, with ten repeats and a size-dependent allowance of 900 or 1,200 seconds. All four coding agents improve on the original population-300 EAX mean of 0.370844\%. They also improve on the stronger population-100 control at 0.021718\%, obtaining means of 0.003481--0.008386\%. On these development instances, the selected programs therefore offer gains beyond those obtained by changing the reference population size alone.

Using the same DeepSeek model, DeepSeek Harness achieves a mean gap of 0.004819\%, lower than OpenEvolve's 0.009293\%; its best gap is also lower, at 0.001467\% versus 0.004344\%. This ordering reverses the PFSP comparison. Across coding agents, GPT has the lowest training mean, followed by DeepSeek, KIMI, and GLM. The different ordering across PFSP and TSP indicates that the strongest tested agent--model pair depends on the underlying algorithm-development task.

BKS hits do not follow the mean-gap ranking. KIMI records the most training hits among the coding agents, 151/330, whereas GPT records 113/330 despite its lower mean gap. The population-300 reference itself records 140 hits but has a much larger mean gap. Reaching BKS on some instances and maintaining a small average gap over the full set are therefore distinct aspects of performance. All retained training methods have 330/330 valid solves; OpenEvolve's training hit count is unavailable in the retained summaries.

\begin{table}[!htbp]
\centering\small
\caption{TSP improvement: training results on 33 instances (10 repeats, 900/1,200~s).}
\label{tab:tsptraining}
\AppendixFitTable{\begin{tabular}{@{}llrrrr@{}}
\toprule
Method & Model & Mean gap (\%) & Best gap (\%) & Valid solves & BKS hits \\
\midrule
\multicolumn{6}{@{}l}{\textit{Human references}} \\
EAX (population 300) & \textemdash & 0.370844 & 0.129431 & 330/330 & 140 \\
EAX (population 100) & \textemdash & 0.021718 & 0.001332 & 330/330 & 82 \\
\midrule
\multicolumn{6}{@{}l}{\textit{Workflow method}} \\
OpenEvolve & DeepSeek & 0.009293 & 0.004344 & 330/330 & \textemdash \\
\midrule
\multicolumn{6}{@{}l}{\textit{Coding agents}} \\
Codex & GPT & 0.003481 & 0.000719 & 330/330 & 113 \\
Kimi-code & KIMI & 0.006006 & 0.002053 & 330/330 & 151 \\
DeepSeek Harness & DeepSeek & 0.004819 & 0.001467 & 330/330 & 95 \\
Zcode & GLM & 0.008386 & 0.001529 & 330/330 & 135 \\
\bottomrule
\end{tabular}}
\end{table}

\FloatBarrier
\paragraph{Test performance.}
The held-out results on 21 instances with 21,215--47,608 cities show a different comparison with the human reference (Table~\ref{tab:eaxresults}). All four coding agents substantially improve on the original population-300 mean gap of 2.240530\%, but none obtains a lower mean than the population-100 control at 0.014315\%. KIMI comes closest at 0.014367\%, followed by DeepSeek at 0.020342\%, GLM at 0.051087\%, and GPT at 0.073410\%. Thus, much of the apparent advantage over the original default is also obtainable through a population-only adjustment, and the additional changes do not establish a consistent held-out advantage over that stronger reference.

The same-model workflow comparison remains favorable to DeepSeek Harness: its mean gap of 0.020342\% is lower than OpenEvolve's 0.041151\%. OpenEvolve is nevertheless better than the GLM and GPT variants. Among coding agents, KIMI has the lowest instance mean on 13 instances, including one tie; it wins 18 instances against DeepSeek and records 17 wins, one tie, and three losses against GPT. Among the coding agents, GPT moves from the lowest training mean to the highest test mean, demonstrating a substantial change in ranking at the intended scale and solving horizon.

The BKS results again reveal a different ordering. GLM records eight hits, compared with two for KIMI, one for DeepSeek, and none for GPT or OpenEvolve. The population-300 and population-100 references record seven and one, respectively. GLM's larger hit count therefore does not translate into the lowest full-set mean. GPT's results are particularly affected by the two largest instances, \texttt{rbz43748} and \texttt{fht47608}: their mean gaps are 0.363388\% and 0.861124\%, accounting for 79.43\% of its summed gaps. Both remain in the aggregate.

All reported variants produce 210/210 valid tours. OpenEvolve was evaluated on CPU4 with its restart cap aligned to the final test protocol (without using test feedback for candidate selection); instance-level results and full execution details appear in Appendices~\ref{app:test-tsp} and~\ref{app:environment}.

\begin{table}[!htbp]
\centering\small
\caption{TSP improvement: held-out results on 21 common instances (10 repeats, 3,600~s).}
\label{tab:eaxresults}
\AppendixFitTable{\begin{tabular}{@{}llrrrr@{}}
\toprule
Method & Model & Mean gap (\%) & Best gap (\%) & Valid solves & BKS hits \\
\midrule
\multicolumn{6}{@{}l}{\textit{Human references}} \\
EAX (population 300) & \textemdash & 2.240530 & 1.624973 & 210/210 & 7 \\
EAX (population 100) & \textemdash & 0.014315 & 0.008909 & 210/210 & 1 \\
\midrule
\multicolumn{6}{@{}l}{\textit{Workflow method}} \\
OpenEvolve & DeepSeek & 0.041151 & 0.029491 & 210/210 & 0 \\
\midrule
\multicolumn{6}{@{}l}{\textit{Coding agents}} \\
Kimi-code & KIMI & 0.014367 & 0.009930 & 210/210 & 2 \\
DeepSeek Harness & DeepSeek & 0.020342 & 0.013077 & 210/210 & 1 \\
Zcode & GLM & 0.051087 & 0.014253 & 210/210 & 8 \\
Codex & GPT & 0.073410 & 0.032602 & 210/210 & 0 \\
\bottomrule
\end{tabular}}
\end{table}

\FloatBarrier
\subsubsection{ROADEF 2018 glass cutting}

\paragraph{Training performance.}
The agents improve a solver derived from the ROADEF 2018 champion, using B1--B15 with three 900-second runs per instance. Table~\ref{tab:r18training} shows that all three selected variants reduce the human reference's mean waste of 5.957267\%. GLM achieves 5.855420\%, KIMI 5.917638\%, and DeepSeek 5.920859\%, corresponding to reductions of 0.101847, 0.039629, and 0.036408 percentage points. The waste reductions are modest in absolute percentage points and arise from an already competitive contest implementation.

GLM has the lowest training waste, while KIMI and DeepSeek are close in mean quality. KIMI has a lower best waste than DeepSeek, 5.903115\% versus 5.920859\%. All methods produce 45/45 valid solutions.

\begin{table}[!htbp]
\centering\small
\caption{ROADEF 2018 improvement: training results on B1--B15 (3 repeats, 900~s).}
\label{tab:r18training}
\AppendixFitTable{\begin{tabular}{@{}llrrr@{}}
\toprule
Method & Model & Mean waste (\%) & Best waste (\%) & Valid solves \\
\midrule
\multicolumn{5}{@{}l}{\textit{Human reference}} \\
Original champion & \textemdash & 5.957267 & 5.957267 & 45/45 \\
\midrule
\multicolumn{5}{@{}l}{\textit{Workflow method}} \\
OpenEvolve & DeepSeek & 5.957267 & 5.957267 & 45/45 \\
\midrule
\multicolumn{5}{@{}l}{\textit{Coding agents}} \\
Kimi-code & KIMI & 5.917638 & 5.903115 & 45/45 \\
Zcode & GLM & 5.855420 & 5.855420 & 45/45 \\
DeepSeek Harness & DeepSeek & 5.920859 & 5.920859 & 45/45 \\
\bottomrule
\end{tabular}}
\end{table}

\FloatBarrier
\paragraph{Test performance.}
On X1--X15, KIMI provides the strongest improvement over the original champion (Table~\ref{tab:r18results}). Its mean waste decreases from 5.344709\% to 5.186161\%, a reduction of 0.158549 percentage points, or approximately 2.97\% relative. KIMI improves 13 instance means and ties the other two. DeepSeek achieves 5.299660\%, a reduction of 0.045049 percentage points, with five wins and ten ties. These results show that autonomous modification can yield gains beyond the development instances even when the starting point is a high-performing expert solver.

OpenEvolve selected its initial champion program after 50 counted evaluations. Its frozen results reproduce the human reference on B and tie it on all 15 X instances. With the same DeepSeek model, the Harness reduces mean held-out waste by 0.0450 percentage points.

The model ranking changes from GLM--KIMI--DeepSeek in training to KIMI--DeepSeek--GLM in testing. GLM obtains 5.317694\%, with six wins, seven ties, and two losses, but 15 of its 45 executions exceed the memory threshold. Its lower mean waste therefore does not establish improvement within the common memory allowance. KIMI and DeepSeek have no such violations. All outputs pass the solution checker, illustrating why solution validity and resource compliance must be assessed separately.

\begin{table}[!htbp]
\centering\small
\caption{ROADEF 2018 improvement: held-out results on X1--X15 (3 repeats, 3,600~s).}
\label{tab:r18results}
\AppendixFitTable{\begin{tabular}{@{}llrrr@{}}
\toprule
Method & Model & Mean waste (\%) & Best waste (\%) & Valid solves \\
\midrule
\multicolumn{5}{@{}l}{\textit{Human reference}} \\
Original champion & \textemdash & 5.344709 & 5.344709 & 45/45 \\
\midrule
\multicolumn{5}{@{}l}{\textit{Workflow method}} \\
OpenEvolve & DeepSeek & 5.344709 & 5.344709 & 45/45 \\
\midrule
\multicolumn{5}{@{}l}{\textit{Coding agents}} \\
Kimi-code & KIMI & 5.186161 & 5.186161 & 45/45 \\
DeepSeek Harness & DeepSeek & 5.299660 & 5.299660 & 45/45 \\
Zcode & GLM & 5.317694 & 5.317694 & 45/45 \\
\bottomrule
\end{tabular}}
\end{table}

\FloatBarrier
\subsubsection{ROADEF 2020 maintenance scheduling}

\paragraph{Training performance.}
Table~\ref{tab:r20-improve-train} reports the final-test-matched ALNS variants on C01--C15, with three runs per instance and a nominal 900-second allowance. Feasibility improves relative to the supplied human implementation: the original ALNS has 39/45 valid runs and returns a valid solution on 14/15 instances, whereas KIMI has 44/45 valid runs and GLM and DeepSeek each have 45/45. All three agents cover all 15 instances. Thus, improving the supplied solver affects both its ability to obtain valid schedules and its objective quality.

With the same DeepSeek model, OpenEvolve selects attempt 51 (counted evaluation 48) and completes all 45 training solves, achieving a full-set mean objective of 19511.28 and a 14-instance mean gap of 1.3671\% (compared with 19459.55 and 1.1379\% for DeepSeek Harness).

On the 12 training instances for which the human reference and all three agents have three valid runs, the original ALNS has mean objective 15130.78, compared with 15089.55 for KIMI, 15050.51 for GLM, and 15056.60 for DeepSeek. The corresponding reductions are 0.27\%, 0.53\%, and 0.49\%. These shared-subset results establish modest quality gains over the reference. On the full training set, DeepSeek has a slightly lower objective than GLM, 19459.55 versus 19467.36. KIMI cannot be placed in that full-set ranking because one of its 45 runs is invalid, despite returning valid solutions on every instance. The incomplete human reference similarly prevents a full-set objective comparison.

\begin{table}[!htbp]
\centering\small
\caption{ROADEF 2020 improvement: training results on C01--C15 (3 repeats, 900~s). Common subset covers the 14 instances (C01--C13 and C15) with three valid runs across all four evolved methods; Full set reports official risk objectives across all 15 instances.}
\label{tab:r20-improve-train}
\AppendixFitTable{\begin{tabular}{@{}llrrrrrr@{}}
\toprule
\multirow{2}{*}{Method} & \multirow{2}{*}{Model} & \multicolumn{2}{c}{Common subset (14)} & \multicolumn{2}{c}{Full set (15)} & \multirow{2}{*}{\shortstack[r]{Valid\\solves}} & \multirow{2}{*}{\shortstack[r]{Valid\\instances}} \\
\cmidrule(lr){3-4}\cmidrule(lr){5-6}
 & & Mean gap (\%) & Best gap (\%) & Mean obj & Best obj & & \\
\midrule
\multicolumn{8}{@{}l}{\textit{Human reference}} \\
Original ALNS & \textemdash & NC & NC & NC & NC & 39/45 & 14/15 \\
\midrule
\multicolumn{8}{@{}l}{\textit{Workflow method}} \\
OpenEvolve & DeepSeek & 1.3671 & 1.2603 & 19511.28 & 19495.88 & 45/45 & 15/15 \\
\midrule
\multicolumn{8}{@{}l}{\textit{Coding agents}} \\
Kimi-code & KIMI & 1.3245 & 1.1657 & NC & NC & 44/45 & 15/15 \\
Zcode & GLM & 1.1580 & 1.0446 & 19467.36 & 19443.78 & 45/45 & 15/15 \\
DeepSeek Harness & DeepSeek & 1.1379 & 0.9778 & 19459.55 & 19438.08 & 45/45 & 15/15 \\
\bottomrule
\end{tabular}}
\end{table}

\FloatBarrier
\paragraph{Test performance.}
Feasibility is the main limitation on the held-out X instances (Table~\ref{tab:r20-improve-valid}). The original ALNS returns valid solutions on 9/15 instances (24/45 valid solves), compared with 10/15 (27/45) for KIMI, 14/15 (38/45) for GLM, 12/15 (34/45) for DeepSeek Harness, and 13/15 (34/45, with all three repeats valid on 10 instances under the enforced 16 GB address-space limit) for OpenEvolve. GLM therefore achieves the broadest coverage and the most valid runs among the improved solvers. Nevertheless, none of the methods completes all 45 runs validly, so none has a full-set mean or best objective. KIMI's three failed X06 runs terminate with SIGABRT; other failures include checker-rejected schedules.

On the six instances with three valid runs for every compared method, all three improvements achieve lower objectives and smaller gaps to the published contest reference than the original ALNS (Table~\ref{tab:r20-improve-valid}). GLM is best at 16959.30 (2.3266\% gap), followed by KIMI at 16980.52 (2.4674\%) and DeepSeek at 16986.54 (2.4121\%), compared with 17166.21 (3.2006\%) for ALNS and 17045.31 (2.6635\%) for OpenEvolve. The objective reductions relative to ALNS range from 1.05\% to 1.21\%. These conditional gains coexist with the losses of feasibility on the other instances: good training objectives do not ensure reliable test execution. The same six-instance subset is used for the scratch comparison in Section~\ref{sec:scratch_results}; execution differences are discussed there.

\begin{table}[!htbp]
\centering\small
\caption{ROADEF 2020 improvement: held-out results on X01--X15 (3 repeats, 900~s). All full-set objectives are NC; Common subset covers the six-instance intersection (X01, X02, X07, X08, X09, and X13) with three valid runs across the reference and all agent variants in both scenarios.}
\label{tab:r20-improve-valid}
\AppendixFitTable{\begin{tabular}{@{}llrrrrrr@{}}
\toprule
\multirow{2}{*}{Method} & \multirow{2}{*}{Model} & \multicolumn{4}{c}{Common subset (6)} & \multicolumn{2}{c}{Full set (15)} \\
\cmidrule(lr){3-6}\cmidrule(lr){7-8}
 & & Mean obj & Best obj & Mean gap (\%) & Best gap (\%) & Valid solves & Valid instances \\
\midrule
\multicolumn{8}{@{}l}{\textit{Human reference}} \\
Original ALNS & \textemdash & 17166.21 & 17115.97 & 3.2006 & 2.9441 & 24/45 & 9/15 \\
\midrule
\multicolumn{8}{@{}l}{\textit{Workflow method}} \\
OpenEvolve & DeepSeek & 17045.31 & 17023.09 & 2.6635 & 2.5484 & 34/45 & 13/15 \\
\midrule
\multicolumn{8}{@{}l}{\textit{Coding agents}} \\
Kimi-code & KIMI & 16980.52 & 16955.08 & 2.4674 & 2.3078 & 27/45 & 10/15 \\
Zcode & GLM & 16959.30 & 16937.96 & 2.3266 & 2.1999 & 38/45 & 14/15 \\
DeepSeek Harness & DeepSeek & 16986.54 & 16949.29 & 2.4121 & 2.1768 & 34/45 & 12/15 \\
\bottomrule
\end{tabular}}
\end{table}
\FloatBarrier

\subsubsection{ROADEF 2022 truck loading}
\label{sec:r22-improve}
\paragraph{Training performance.}
Table~\ref{tab:r22-improvement-C} reports the independent three-repeat training confirmations on set C (50 instances). To enable a fair quality comparison across both development scenarios, the common subset retains the 47 instances on which all seven frozen solvers (S46--HiGHS V6, the three improved variants, and the three scratch solvers in Section~\ref{sec:r22-scratch}) complete all three repeats validly. On this 47-instance subset, all three coding agents reduce V6's mean gap of 1.0378\%: KIMI achieves the largest improvement to 0.0789\% (with a best gap of $-0.0102$\%), followed by DeepSeek at 0.8289\% and GLM at 0.9117\%. In terms of reliability, GLM is the only improved solver achieving 150/150 valid solves across all 50 training instances (full-set mean gap 0.8573\%), while KIMI (149/150) and DeepSeek (147/150) also improve upon V6's 146/150 valid solves.
\begin{table}[!htbp]
\centering\small
\caption{ROADEF 2022 improvement: training results on set C (50 instances, 3 repeats, 900/1,500~s). Common subset covers the 47 instances with three valid runs across all seven frozen solvers (V6, three improved, and three scratch); Complete instances counts instances with all three repeats valid.}
\label{tab:r22-improvement-C}
\AppendixFitTable{\begin{tabular}{@{}llrrrrrr@{}}
\toprule
\multirow{2}{*}{Method} & \multirow{2}{*}{Model} & \multicolumn{2}{c}{Common subset (47)} & \multicolumn{2}{c}{Full set (50)} & \multirow{2}{*}{\shortstack[r]{Valid\\solves}} & \multirow{2}{*}{\shortstack[r]{Complete\\instances}} \\
\cmidrule(lr){3-4}\cmidrule(lr){5-6}
 & & Mean gap (\%) & Best gap (\%) & Mean gap (\%) & Best gap (\%) & & \\
\midrule
\multicolumn{8}{@{}l}{\textit{Human reference}} \\
S46--HiGHS V6 & \textemdash & 1.0378 & 0.9910 & NC & NC & 146/150 & 48/50 \\
\midrule
\multicolumn{8}{@{}l}{\textit{Coding agents}} \\
Kimi-code & KIMI & 0.0789 & $-0.0102$ & NC & NC & 149/150 & 49/50 \\
Zcode & GLM & 0.9117 & 0.8396 & 0.8573 & 0.7864 & 150/150 & 50/50 \\
DeepSeek Harness & DeepSeek & 0.8289 & 0.7776 & NC & NC & 147/150 & 49/50 \\
\bottomrule
\end{tabular}}
\end{table}
\FloatBarrier
\paragraph{Test performance.}
On the 105 held-out X instances, all three improved solvers maintain lower mean gaps than V6 on the 83-instance common subset (Table~\ref{tab:r22-improvement-X}). KIMI delivers the strongest transfer in both quality and reliability: its common-subset mean gap drops from 0.4188\% for V6 to $-0.2928$\% (a reduction of 0.7115 percentage points, surpassing the aggregated final-phase contest reference), while increasing valid solves from 271/315 to 299/315 and complete instances from 90/105 to 93/105. DeepSeek and GLM also lower the common-subset mean gap to 0.2106\% and 0.2662\%, respectively, with GLM gaining four valid solves (275/315) and DeepSeek matching V6's 271/315. Because no improved variant completes all 315 runs validly, all full-set X gaps remain NC. Instance-level results appear in Appendix~\ref{app:test-r22}.
\begin{table}[!htbp]
\centering\small
\caption{ROADEF 2022 improvement: held-out results on set X (105 instances, 3 repeats, 900/1,500~s). Common subset covers the 83 instances with three valid runs across all seven frozen solvers.}
\label{tab:r22-improvement-X}
\AppendixFitTable{\begin{tabular}{@{}llrrrrrr@{}}
\toprule
\multirow{2}{*}{Method} & \multirow{2}{*}{Model} & \multicolumn{2}{c}{Common subset (83)} & \multicolumn{2}{c}{Full set (105)} & \multirow{2}{*}{\shortstack[r]{Valid\\solves}} & \multirow{2}{*}{\shortstack[r]{Complete\\instances}} \\
\cmidrule(lr){3-4}\cmidrule(lr){5-6}
 & & Mean gap (\%) & Best gap (\%) & Mean gap (\%) & Best gap (\%) & & \\
\midrule
\multicolumn{8}{@{}l}{\textit{Human reference}} \\
S46--HiGHS V6 & \textemdash & 0.4188 & 0.3574 & NC & NC & 271/315 & 90/105 \\
\midrule
\multicolumn{8}{@{}l}{\textit{Coding agents}} \\
Kimi-code & KIMI & $-0.2928$ & $-0.4083$ & NC & NC & 299/315 & 93/105 \\
Zcode & GLM & 0.2662 & 0.1884 & NC & NC & 275/315 & 91/105 \\
DeepSeek Harness & DeepSeek & 0.2106 & 0.1708 & NC & NC & 271/315 & 90/105 \\
\bottomrule
\end{tabular}}
\end{table}
\FloatBarrier

\subsection{Algorithm development from scratch}
\label{sec:scratch_results}

We next evaluate the three coding agents when no complete human solver is supplied. Every frozen program is independently reevaluated on both training and held-out instances. The human references and solving allowances are shared with the improvement comparison. Workflow baselines were evaluated only in the improvement setting; this section therefore compares the scratch programs with the human solvers and with one another. The results reveal a strong dependence on the problem domain: construction from scratch is competitive on PFSP, remains behind the expert references on TSP, ROADEF 2018 and ROADEF 2022, and has mixed quality and feasibility outcomes on ROADEF 2020.

\subsubsection{Permutation flow-shop scheduling}

\paragraph{Training performance.}
All three scratch solvers outperform both supplied human references on the 60 training instances (Table~\ref{tab:scratch-pfsp-train}). KIMI has the lowest mean gap at $-0.227942$\%, followed by GLM at $-0.204571$\% and DeepSeek at $-0.152874$\%, compared with 3.003299\% for original IG and 3.902482\% for Basic ALNS. KIMI also has the lowest best gap, $-0.436801$\%. The scratch programs therefore construct effective search procedures without receiving the original implementation.

The negative-gap counts strengthen this comparison. KIMI improves on the fixed UB in 478/600 runs, GLM in 473/600, and DeepSeek in 415/600; neither human reference has such a run. All 1,800 agent runs are valid. These results establish improvement over the supplied fixed upper bounds and reference programs; they do not identify those bounds as proven optima.

\begin{table}[!htbp]
\centering\small
\caption{PFSP scratch development: training results on 60 instances (10 repeats, $T=0.03nm$).}
\label{tab:scratch-pfsp-train}
\AppendixFitTable{\begin{tabular}{@{}llrrrr@{}}
\toprule
Method & Model & Mean gap (\%) & Best gap (\%) & Valid solves & Below UB \\
\midrule
\multicolumn{6}{@{}l}{\textit{Human references}} \\
Original IG & \textemdash & 3.003299 & 2.493410 & 600/600 & 0 \\
Basic ALNS & \textemdash & 3.902482 & 3.491333 & 600/600 & 0 \\
\midrule
\multicolumn{6}{@{}l}{\textit{Coding agents}} \\
Kimi-code & KIMI & $-0.227942$ & $-0.436801$ & 600/600 & 478 \\
Zcode & GLM & $-0.204571$ & $-0.418319$ & 600/600 & 473 \\
DeepSeek Harness & DeepSeek & $-0.152874$ & $-0.395464$ & 600/600 & 415 \\
\bottomrule
\end{tabular}}
\end{table}

\FloatBarrier
\paragraph{Test performance.}
All three programs retain their advantage on the 180 larger held-out instances (Table~\ref{tab:final-PFSP}). Their mean gaps are $-0.222135$\% for KIMI, $-0.449989$\% for GLM, and $-0.456489$\% for DeepSeek, compared with 2.053454\% for original IG and 2.346119\% for Basic ALNS. DeepSeek has the lowest mean and best gap, with GLM close behind; KIMI, which led training, ranks third in testing. The supplied IG reference is therefore surpassed in both development scenarios, while the strongest scratch programs also improve substantially on the fixed test UBs.

The frequency of beating UB has a different ordering from mean quality. GLM records 1,527 such runs, KIMI 1,333, and DeepSeek 1,303, out of 1,800 each. DeepSeek's lower average gap thus coexists with fewer below-UB runs than GLM. All 5,400 scratch test runs are valid. KIMI, GLM, and DeepSeek exceed the nominal budget by more than one second in 279, two, and 26 runs, with maximum excesses of 5.750, 1.428, and 2.600 seconds. The quality scores describe completed runs under the recorded watchdog, and do not certify strict deadline truncation. Competitiveness here is established against the tested references, rather than every available PFSP solver.

\begin{table}[!htbp]
\centering\small
\caption{PFSP scratch development: held-out results on 180 instances (10 repeats, $T=0.06nm$).}
\label{tab:final-PFSP}
\AppendixFitTable{\begin{tabular}{@{}llrrrr@{}}
\toprule
Method & Model & Mean gap (\%) & Best gap (\%) & Valid solves & Below UB \\
\midrule
\multicolumn{6}{@{}l}{\textit{Human references}} \\
Original IG & \textemdash & 2.053454 & 1.827103 & 1800/1800 & 0 \\
Basic ALNS & \textemdash & 2.346119 & 1.926043 & 1800/1800 & 0 \\
\midrule
\multicolumn{6}{@{}l}{\textit{Coding agents}} \\
Kimi-code & KIMI & $-0.222135$ & $-0.404220$ & 1800/1800 & 1333 \\
Zcode & GLM & $-0.449989$ & $-0.563074$ & 1800/1800 & 1527 \\
DeepSeek Harness & DeepSeek & $-0.456489$ & $-0.587405$ & 1800/1800 & 1303 \\
\bottomrule
\end{tabular}}
\end{table}

\FloatBarrier
\subsubsection{Traveling salesman problem}

\paragraph{Training performance.}
TSP shows a different outcome from PFSP (Table~\ref{tab:scratch-tsp-train}). All three scratch solvers produce valid tours on the 33 training instances, but their mean and best gaps remain above the configured human EAX reference. DeepSeek is best at 0.3875\% mean gap, followed by KIMI at 0.5793\% and GLM at 1.6294\%, whereas population-100 EAX obtains 0.021718\%. DeepSeek's best gap, 0.2842\%, is also well above EAX's 0.001332\%. The remaining difference is therefore evident even on the instances used for development.

None of the 990 agent training runs attains BKS, while the EAX reference does so in 82/330 runs. Producing valid tours is consequently much easier for the tested agents than reproducing the solution quality of the mature genetic solver. Across the three models, DeepSeek gives the strongest training result, in contrast to KIMI's leading scratch PFSP training score.

\begin{table}[!htbp]
\centering\small
\caption{TSP scratch development: training results on 33 instances (10 repeats, 900/1,200~s).}
\label{tab:scratch-tsp-train}
\AppendixFitTable{\begin{tabular}{@{}llrrrr@{}}
\toprule
Method & Model & Mean gap (\%) & Best gap (\%) & Valid solves & BKS hits \\
\midrule
\multicolumn{6}{@{}l}{\textit{Human reference}} \\
EAX (population 100) & \textemdash & 0.021718 & 0.001332 & 330/330 & 82 \\
\midrule
\multicolumn{6}{@{}l}{\textit{Coding agents}} \\
Kimi-code & KIMI & 0.5793 & 0.4227 & 330/330 & 0 \\
Zcode & GLM & 1.6294 & 1.2377 & 330/330 & 0 \\
DeepSeek Harness & DeepSeek & 0.3875 & 0.2842 & 330/330 & 0 \\
\bottomrule
\end{tabular}}
\end{table}

\FloatBarrier
\paragraph{Test performance.}
The same ordering persists on the 21 larger held-out instances (Table~\ref{tab:final-TSP}). DeepSeek achieves mean gap 0.3995\%, KIMI 0.7289\%, and GLM 2.1973\%, compared with 0.0143\% for population-100 EAX. Their best gaps also remain higher than the human reference. Although the scratch programs improve on the original population-300 default reported in Table~\ref{tab:eaxresults}, none approaches the quality of the stronger population-100 configuration. This difference highlights the importance of comparing generated algorithms with a competitive reference configuration.

All 630 agent test tours are valid, but none attains BKS; the human reference records one hit. The test deficit therefore concerns optimization quality rather than failure to construct tours. DeepSeek remains the strongest scratch solver across training and testing, while all three remain behind the expert implementation. Test overruns above one second occur in three KIMI, one GLM, and 26 DeepSeek runs, with maximum excesses of 7.518, 3.353, and 7.301 seconds; DeepSeek also has nine such training runs. These are completed-run quality measurements under the recorded timing tolerance.

\begin{table}[!htbp]
\centering\small
\caption{TSP scratch development: held-out results on 21 instances (10 repeats, 3,600~s).}
\label{tab:final-TSP}
\AppendixFitTable{\begin{tabular}{@{}llrrrr@{}}
\toprule
Method & Model & Mean gap (\%) & Best gap (\%) & Valid solves & BKS hits \\
\midrule
\multicolumn{6}{@{}l}{\textit{Human reference}} \\
EAX (population 100) & \textemdash & 0.0143 & 0.0089 & 210/210 & 1 \\
\midrule
\multicolumn{6}{@{}l}{\textit{Coding agents}} \\
Kimi-code & KIMI & 0.7289 & 0.6345 & 210/210 & 0 \\
Zcode & GLM & 2.1973 & 1.9101 & 210/210 & 0 \\
DeepSeek Harness & DeepSeek & 0.3995 & 0.3291 & 210/210 & 0 \\
\bottomrule
\end{tabular}}
\end{table}

\FloatBarrier
\subsubsection{ROADEF 2018 glass cutting}

\paragraph{Training performance.}
The scratch solvers all return valid cutting plans, but none matches the champion reference on B1--B15 (Table~\ref{tab:scratch-r18-train}). GLM has the lowest agent mean waste, 18.7575\%, followed by DeepSeek at 22.8825\% and KIMI at 27.1226\%; the champion obtains 5.9573\%. The best-waste values show the same ordering and remain far above the human result. With all 45 runs valid for each method, the large quality deficit is already present during development.

This result contrasts with improving the supplied champion, where every tested agent reduces its training waste. For this constrained industrial problem, the available scratch campaigns construct feasible programs but do not reproduce the effectiveness of the expert algorithmic structure. GLM is the strongest of the three scratch systems, even though its improvement variant does not retain first place in the held-out comparison.

\begin{table}[!htbp]
\centering\small
\caption{ROADEF 2018 scratch development: training results on B1--B15 (3 repeats, 900~s).}
\label{tab:scratch-r18-train}
\AppendixFitTable{\begin{tabular}{@{}llrrr@{}}
\toprule
Method & Model & Mean waste (\%) & Best waste (\%) & Valid solves \\
\midrule
\multicolumn{5}{@{}l}{\textit{Human reference}} \\
Original champion & \textemdash & 5.9573 & 5.9573 & 45/45 \\
\midrule
\multicolumn{5}{@{}l}{\textit{Coding agents}} \\
Kimi-code & KIMI & 27.1226 & 26.9064 & 45/45 \\
Zcode & GLM & 18.7575 & 18.5739 & 45/45 \\
DeepSeek Harness & DeepSeek & 22.8825 & 22.5871 & 45/45 \\
\bottomrule
\end{tabular}}
\end{table}

\FloatBarrier
\paragraph{Test performance.}
The quality difference remains on X1--X15 under the longer 3,600-second allowance (Table~\ref{tab:final-R18}). GLM again performs best among the scratch solvers, with mean waste 18.7134\%, compared with 23.4481\% for DeepSeek and 27.3646\% for KIMI. The champion obtains 5.3447\%, leaving a difference of 13.3687 percentage points even for the strongest scratch program. The best-waste values do not close this difference.

All 45 test runs for every method are valid, and the GLM--DeepSeek--KIMI ranking is preserved from training. Thus, the main difficulty is not feasibility transfer but attaining competitive cutting quality. The contrast with the successful champion improvements indicates the practical value of starting from an established solver in this domain. KIMI frequently terminates before its allowance is exhausted; a larger permitted budget alone does not guarantee that the generated algorithm uses the additional search time effectively.

\begin{table}[!htbp]
\centering\small
\caption{ROADEF 2018 scratch development: held-out results on X1--X15 (3 repeats, 3,600~s).}
\label{tab:final-R18}
\AppendixFitTable{\begin{tabular}{@{}llrrr@{}}
\toprule
Method & Model & Mean waste (\%) & Best waste (\%) & Valid solves \\
\midrule
\multicolumn{5}{@{}l}{\textit{Human reference}} \\
Original champion & \textemdash & 5.3447 & 5.3447 & 45/45 \\
\midrule
\multicolumn{5}{@{}l}{\textit{Coding agents}} \\
Kimi-code & KIMI & 27.3646 & 27.2006 & 45/45 \\
Zcode & GLM & 18.7134 & 18.4862 & 45/45 \\
DeepSeek Harness & DeepSeek & 23.4481 & 23.1829 & 45/45 \\
\bottomrule
\end{tabular}}
\end{table}

\FloatBarrier
\subsubsection{ROADEF 2020 maintenance scheduling}
\label{sec:r20-scratch}

\paragraph{Training performance.}
All three scratch solvers achieve 45/45 valid runs and cover all 15 training instances, improving on the human ALNS reference's 39/45 valid runs and 14/15-instance coverage (Table~\ref{tab:scratch-r20-train}). KIMI has the lowest full-set mean objective, 19304.7048, followed by DeepSeek at 19345.4636 and GLM at 20611.9668. KIMI also has the lowest best objective. A full-set human mean is unavailable because its training runs are not all valid.

On the 12 training instances with three valid runs for the human reference and all agents, KIMI and DeepSeek obtain mean objectives 14946.02 and 14968.73, below ALNS's 15130.78, whereas GLM obtains 15799.42. The corresponding changes are improvements of 1.22\% and 1.07\% for KIMI and DeepSeek, and a deterioration of 4.42\% for GLM. Hence, the common-valid comparison supports a quality advantage for two scratch solvers, while feasibility coverage improves for all three.

\begin{table}[!htbp]
\centering\small
\caption{ROADEF 2020 scratch development: training official risk objectives on C01--C15 (3 repeats, 900~s).}
\label{tab:scratch-r20-train}
\AppendixFitTable{\begin{tabular}{@{}llrrrr@{}}
\toprule
Method & Model & Mean obj & Best obj & Valid solves & Valid instances \\
\midrule
\multicolumn{6}{@{}l}{\textit{Human reference}} \\
Original ALNS & \textemdash & NC & NC & 39/45 & 14/15 \\
\midrule
\multicolumn{6}{@{}l}{\textit{Coding agents}} \\
Kimi-code & KIMI & 19304.7048 & 19295.7457 & 45/45 & 15/15 \\
Zcode & GLM & 20611.9668 & 20573.9951 & 45/45 & 15/15 \\
DeepSeek Harness & DeepSeek & 19345.4636 & 19332.8633 & 45/45 & 15/15 \\
\bottomrule
\end{tabular}}
\end{table}

\FloatBarrier
\paragraph{Test performance.}
Each scratch solver returns at least one valid schedule on every X instance (Table~\ref{tab:final-R20}), compared with nine covered instances for the original ALNS. GLM and DeepSeek retain 45/45 valid runs, while KIMI has 44/45 because one X12 run is invalid. Covering all 15 instances therefore does not imply that all repeats are reliable. Only GLM and DeepSeek receive full-set quality scores; DeepSeek's mean objective of 28266.3098 is 6.15\% lower than GLM's 30119.1656, and its best objective is lower as well. KIMI cannot be ranked against these full-set means.

The six-instance common-valid comparison gives a more detailed view of quality (Table~\ref{tab:final-R20}). KIMI is best among the scratch solvers at 16696.18, followed by DeepSeek at 16740.43, compared with 17166.21 for original ALNS. These are improvements of 2.74\% and 2.48\%. GLM obtains 17887.77, approximately 4.20\% worse than ALNS. Accordingly, broader feasibility coverage is not sufficient to ensure superior objective quality, and the model with the best conditional score is not necessarily the one with fully reliable repeats.

\begin{table}[!htbp]
\centering\small
\caption{ROADEF 2020 scratch development: held-out official risk objectives on X01--X15 (3 repeats, 900~s). Common subset covers the same six-instance intersection as Table~\ref{tab:r20-improve-valid}.}
\label{tab:final-R20}
\AppendixFitTable{\begin{tabular}{@{}llrrrrrr@{}}
\toprule
\multirow{2}{*}{Method} & \multirow{2}{*}{Model} & \multicolumn{2}{c}{Common subset (6)} & \multicolumn{2}{c}{Full set (15)} & \multirow{2}{*}{\shortstack[r]{Valid\\solves}} & \multirow{2}{*}{\shortstack[r]{Valid\\instances}} \\
\cmidrule(lr){3-4}\cmidrule(lr){5-6}
 & & Mean obj & Best obj & Mean obj & Best obj & & \\
\midrule
\multicolumn{8}{@{}l}{\textit{Human reference}} \\
Original ALNS & \textemdash & 17166.21 & 17115.97 & NC & NC & 24/45 & 9/15 \\
\midrule
\multicolumn{8}{@{}l}{\textit{Coding agents}} \\
Kimi-code & KIMI & 16696.18 & 16673.56 & NC & NC & 44/45 & 15/15 \\
Zcode & GLM & 17887.77 & 17857.35 & 30119.1656 & 30046.9917 & 45/45 & 15/15 \\
DeepSeek Harness & DeepSeek & 16740.43 & 16725.96 & 28266.3098 & 28244.9791 & 45/45 & 15/15 \\
\bottomrule
\end{tabular}}
\end{table}

Training and test objectives concern different instance sets, so their raw magnitudes do not measure generalization loss. The shared quality subsets are selected by common validity and do not replace full-set scores. Execution conditions also differ across campaigns: improvement candidates use CPU3 with an internal 850-second cutoff and four OpenMP threads, whereas the CPU4 reference uses an internal 870-second cutoff and a one-thread OpenMP environment. The common 900-second external allowance therefore supports a descriptive comparison of recorded outcomes rather than a strictly controlled causal comparison of development strategies.

\FloatBarrier

\subsubsection{ROADEF 2022 truck loading}
\label{sec:r22-scratch}
\paragraph{Training performance.}
All three scratch solvers complete 150/150 valid solves across all 50 training instances on set C, surpassing V6's 146/150 reliability (Table~\ref{tab:r22-scratch-C}). However, their optimization quality remains far behind the matheuristic reference: full-set mean gaps range from 23.8173\% for KIMI and 34.2910\% for DeepSeek to 68.0991\% for GLM. Restricting the comparison to the same 47-instance common subset used in Section~\ref{sec:r22-improve} preserves this ordering and confirms a gap of more than 23 percentage points even between the strongest scratch solver (KIMI, 24.0884\%) and V6 (1.0378\%).
\begin{table}[!htbp]
\centering\small
\caption{ROADEF 2022 scratch development: training results on set C (50 instances, 3 repeats, 900/1,500~s), using the same 47-instance common subset as Table~\ref{tab:r22-improvement-C}.}
\label{tab:r22-scratch-C}
\AppendixFitTable{\begin{tabular}{@{}llrrrrrr@{}}
\toprule
\multirow{2}{*}{Method} & \multirow{2}{*}{Model} & \multicolumn{2}{c}{Common subset (47)} & \multicolumn{2}{c}{Full set (50)} & \multirow{2}{*}{\shortstack[r]{Valid\\solves}} & \multirow{2}{*}{\shortstack[r]{Complete\\instances}} \\
\cmidrule(lr){3-4}\cmidrule(lr){5-6}
 & & Mean gap (\%) & Best gap (\%) & Mean gap (\%) & Best gap (\%) & & \\
\midrule
\multicolumn{8}{@{}l}{\textit{Human reference}} \\
S46--HiGHS V6 & \textemdash & 1.0378 & 0.9910 & NC & NC & 146/150 & 48/50 \\
\midrule
\multicolumn{8}{@{}l}{\textit{Coding agents}} \\
Kimi-code & KIMI & 24.0884 & 23.8850 & 23.8173 & 23.6229 & 150/150 & 50/50 \\
Zcode & GLM & 68.7203 & 68.5193 & 68.0991 & 67.9064 & 150/150 & 50/50 \\
DeepSeek Harness & DeepSeek & 33.8891 & 33.8891 & 34.2910 & 34.2910 & 150/150 & 50/50 \\
\bottomrule
\end{tabular}}
\end{table}
\FloatBarrier
\paragraph{Test performance.}
This contrast between feasibility and cost optimization persists on the 105 held-out X instances (Table~\ref{tab:r22-scratch-X}). Every scratch program achieves 315/315 valid solves across all 105 instances, outperforming V6's 271/315 coverage, yet their full-set mean gaps remain high at 22.4221\% (KIMI), 34.7338\% (DeepSeek), and 64.1247\% (GLM). Restricting the comparison to the 83-instance common subset preserves the exact ordering: even the strongest scratch solver, KIMI (21.7191\%), trails V6 (0.4188\%) by 21.3004 percentage points, while GLM trails by 62.9986 percentage points. Thus, coding agents can reliably satisfy the intricate 3D stacking, loading-order, and axle-load constraints from scratch, but leave substantial transport and inventory cost savings unrealized compared with an expert MIP-based matheuristic.
\begin{table}[!htbp]
\centering\small
\caption{ROADEF 2022 scratch development: held-out results on set X (105 instances, 3 repeats, 900/1,500~s), using the same 83-instance common subset as Table~\ref{tab:r22-improvement-X}.}
\label{tab:r22-scratch-X}
\AppendixFitTable{\begin{tabular}{@{}llrrrrrr@{}}
\toprule
\multirow{2}{*}{Method} & \multirow{2}{*}{Model} & \multicolumn{2}{c}{Common subset (83)} & \multicolumn{2}{c}{Full set (105)} & \multirow{2}{*}{\shortstack[r]{Valid\\solves}} & \multirow{2}{*}{\shortstack[r]{Complete\\instances}} \\
\cmidrule(lr){3-4}\cmidrule(lr){5-6}
 & & Mean gap (\%) & Best gap (\%) & Mean gap (\%) & Best gap (\%) & & \\
\midrule
\multicolumn{8}{@{}l}{\textit{Human reference}} \\
S46--HiGHS V6 & \textemdash & 0.4188 & 0.3574 & NC & NC & 271/315 & 90/105 \\
\midrule
\multicolumn{8}{@{}l}{\textit{Coding agents}} \\
Kimi-code & KIMI & 21.7191 & 21.4283 & 22.4221 & 22.1823 & 315/315 & 105/105 \\
Zcode & GLM & 63.4174 & 62.7365 & 64.1247 & 63.5765 & 315/315 & 105/105 \\
DeepSeek Harness & DeepSeek & 34.0963 & 34.0963 & 34.7338 & 34.7338 & 315/315 & 105/105 \\
\bottomrule
\end{tabular}}
\end{table}
\FloatBarrier

\subsection{Main findings}
\label{sec:algorithm_findings}

The results address five questions about autonomous algorithm development: how coding agents compare with specialized AHD frameworks, whether they improve supplied solvers on training instances, whether those improvements transfer, what they can build from scratch, and how far they can change the mechanisms of an established algorithm.

\subsubsection{Comparison with fixed workflows}

General-purpose coding agents can achieve results comparable to or better than fixed workflows in algorithm-development tasks. In this study, their advantage is more pronounced when improving a complex solver such as EAX, showing the potential of autonomous exploration and iterative optimization for developing mature algorithms.

On PFSP, the four coding agents achieve test mean gaps of 0.450912--1.076633\%, all substantially below ReEvo's 1.935213\% (Table~\ref{tab:igresults}). DeepSeek Harness also performs better when using the same DeepSeek model. In our ReEvo adaptation, the destruction operator to be optimized is specified by a human, and the model searches implementations within that interface rather than autonomously choosing which part of the solver to modify. Coding agents, in contrast, can locate modification targets and coordinate changes across different parts of the solver. They therefore obtain better final solution quality while reducing reliance on a human-specified optimization target.

The comparison with OpenEvolve further reveals differences between tasks. With DeepSeek, OpenEvolve performs better on PFSP, obtaining a test mean gap of 0.910495\%, compared with 1.076633\% for DeepSeek Harness, a lead of approximately 0.166 percentage points. On the more complex TSP/EAX task, however, the ordering reverses: OpenEvolve obtains 0.041151\%, whereas DeepSeek Harness achieves 0.020342\%, reducing the gap by approximately 50.6\% (Table~\ref{tab:eaxresults}). Thus, the advantage of the fixed program-evolution workflow on PFSP does not persist on the EAX improvement task; an agent-directed development process produces the better result for this more complex solver.

On the industrial tasks, the same-model advantage of DeepSeek Harness over OpenEvolve persists. On ROADEF 2018, OpenEvolve reverts to the unmodified champion solver, whereas DeepSeek Harness reduces held-out mean waste by 0.0450 percentage points (five wins and ten ties across X1--X15; Table~\ref{tab:r18results}). On ROADEF 2020, both methods improve on the original ALNS with identical overall valid-run counts (34/45), while DeepSeek Harness achieves a lower mean gap on the six commonly valid X instances (2.4121\% versus 2.6635\%, a 0.2513-point advantage on that subset; Table~\ref{tab:r20-improve-valid}).

Overall, fixed workflows show no consistent advantage in these experiments. Coding agents outperform the component-level ReEvo adaptation without requiring a predefined modification target, and outperform OpenEvolve on TSP/EAX even though it permits changes to the full program. These results show that autonomously selecting modification targets, organizing development actions, and using execution feedback can constitute an effective approach to algorithm development, with competitive results on complex solver improvement.

\subsubsection{Can coding agents improve advanced algorithms?}

The experiments show that LLM-driven coding agents can further improve the performance of existing advanced algorithms on training data. These gains appear as lower objective values on classical combinatorial optimization problems, and as improvements in solution quality and solving success rates on industrial problems.

On PFSP, the four coding agents achieve training mean gaps of 0.982869--1.655192\%, all below the original IG's 3.003299\% (Table~\ref{tab:pfsptraining}). On TSP, their training mean gaps are 0.003481--0.008386\%, all below both the original population-300 EAX's 0.370844\% and the population-100 EAX control's 0.021718\% (Table~\ref{tab:tsptraining}). Thus, the agents improve on the original implementation and also outperform this configuration control on the training set.

The industrial problems show the same improvement trend. On ROADEF 2018, the three coding agents achieve training mean waste of 5.855420--5.920859\%, all below the original champion-derived solver's 5.957267\% (Table~\ref{tab:r18training}), demonstrating that agents can further reduce material loss even when starting from a high-performance contest algorithm. On ROADEF 2020, the agents improve both solving reliability and objective quality: valid runs increase from the original ALNS's 39/45 to 44/45 for KIMI and 45/45 for both GLM and DeepSeek (Table~\ref{tab:r20-improve-train}), while on the 12 commonly valid training instances their mean objectives drop to 15050.51--15089.55 compared with 15130.78 for ALNS (reductions of 0.27--0.53\%). Similarly, on ROADEF 2022, all three S46 improvements reduce V6's training mean gap of 1.0378\% on the 47 common C instances---down to 0.0789\% for KIMI, 0.8289\% for DeepSeek, and 0.9117\% for GLM---while also increasing valid solves, with GLM achieving 150/150 complete training reliability (Table~\ref{tab:r22-improvement-C}). Autonomous modification of existing algorithms can therefore improve both the reliability of obtaining valid solutions and optimization quality across diverse industrial domains.

\subsubsection{Do coding agents' algorithm improvements generalize to unseen instances?}

The experiments show that the algorithm improvements obtained by coding agents during training can generalize to unseen instances, continuing to outperform the original human-designed algorithms on larger or structurally different test tasks. This generalization appears both in improved solution quality on classical combinatorial optimization problems and in greater valid-instance coverage and lower objective values on industrial problems.

On classical combinatorial optimization problems, the four coding agents achieve mean gaps of 0.450912--1.076633\% on the 180 larger PFSP test instances, all below the original IG's 2.053454\% (Table~\ref{tab:igresults}). On the 21 larger TSP test instances, their mean gaps are 0.014367--0.073410\%, also all below the original EAX's 2.240530\% (Table~\ref{tab:eaxresults}). Thus, modifications developed on the training instances transfer to larger problems while retaining a performance advantage over the original algorithms.

The industrial problems further demonstrate this generalization capability. On ROADEF 2018, KIMI, DeepSeek, and GLM achieve test mean waste of 5.186161\%, 5.299660\%, and 5.317694\%, respectively, all below the original champion-derived solver's 5.344709\% (Table~\ref{tab:r18results}), with KIMI improving on 13 instances and tying on two. On ROADEF 2020, the original ALNS obtains valid solutions on nine of the 15 test instances, whereas KIMI, GLM, and DeepSeek cover ten, 14, and 12 instances, respectively; valid runs increase from 24/45 to 27/45, 38/45, and 34/45, and on the six common instances all three agents reduce the mean objective from 17166.21 to 16959.30--16986.54, corresponding to reductions of 1.05--1.21\% (Table~\ref{tab:r20-improve-valid}). On ROADEF 2022, all three S46 improvements likewise retain lower mean gaps than V6 (0.4188\%) on the 83 common X instances: KIMI achieves the largest gain at $-0.2928$\% while adding 28 valid solves (299/315 vs.\ 271/315), GLM reaches 0.2662\% with 275/315 valid solves, and DeepSeek lowers the common-instance gap to 0.2106\% while matching V6's 271/315 coverage (Table~\ref{tab:r22-improvement-X}). These results show that training-stage improvements can continue to enhance both solving success rates and solution quality on unseen industrial instances.

However, generalization does not preserve every training ranking or execution property. GPT moves from third in training to second in testing among the agents on PFSP, but from first to last on TSP. On ROADEF 2018, GLM has the lowest training waste, whereas KIMI has the lowest test waste, and GLM incurs memory violations during testing. On ROADEF 2022, GLM achieves the only complete training reliability (150/150), whereas KIMI leads both quality and reliability on the held-out set. The strongest agent on the training set therefore need not remain the strongest on unseen instances. Coding agents can develop algorithm improvements that generalize, but their final competitiveness must be assessed jointly through test solution quality, solving success rates, and resource compliance.

\subsubsection{Can coding agents design advanced algorithms from scratch?}

The experiments show that coding agents can build efficient, competitive algorithms from scratch on problems with relatively simple algorithmic structures, even outperforming the human-designed algorithms compared here and improving existing upper bounds. As algorithmic structures and constraint interactions become more complex, however, development from scratch becomes substantially harder, and agents find it more difficult to consistently match the performance of mature human-designed algorithms.

PFSP provides the clearest success. On the training set, the three agents achieve mean gaps ranging from $-0.227942$\% to $-0.152874$\%, all substantially below the original IG's 3.003299\% and Basic ALNS's 3.902482\% (Table~\ref{tab:scratch-pfsp-train}). On the test set, KIMI, GLM, and DeepSeek achieve mean gaps of $-0.222135$\%, $-0.449989$\%, and $-0.456489$\%, respectively, again substantially below the original IG's 2.053454\% and Basic ALNS's 2.346119\% (Table~\ref{tab:final-PFSP}). The frequency of beating the supplied UB further demonstrates their quality: the three agents obtain negative gaps in 478, 473, and 415 of their respective 600 training runs, and in 1333, 1527, and 1303 of their respective 1800 test runs, whereas neither human reference beats the UB. Thus, agents can independently construct complete solvers and design PFSP search methods that consistently outperform the human algorithms compared here while repeatedly improving existing upper bounds.

TSP reveals a gap relative to mature human-designed algorithms. On the training set, KIMI, GLM, and DeepSeek achieve mean gaps of 0.5793\%, 1.6294\%, and 0.3875\%, respectively, compared with 0.021718\% for the human EAX reference (Table~\ref{tab:scratch-tsp-train}). On the test set, their mean gaps are 0.7289\%, 2.1973\%, and 0.3995\%, all above EAX's 0.0143\% (Table~\ref{tab:final-TSP}). Although every run produces a valid tour, none attains BKS. Agents can therefore design TSP algorithms with useful optimization capability from scratch, but do not yet match the solution quality of the mature expert algorithm.

The difference is more pronounced on ROADEF 2018. On the training set, KIMI, GLM, and DeepSeek achieve mean waste of 27.1226\%, 18.7575\%, and 22.8825\%, respectively, whereas the champion-derived solver achieves only 5.9573\% (Table~\ref{tab:scratch-r18-train}). On the test set, the agents obtain mean waste of 27.3646\%, 18.7134\%, and 23.4481\%, again far above the human reference's 5.3447\% (Table~\ref{tab:final-R18}). Even the strongest agent, GLM, has test waste 13.3687 percentage points above the human algorithm. Every run produces a valid solution, indicating that the main difficulty lies in coordinating complex cutting constraints and using material effectively, rather than merely satisfying feasibility requirements.

ROADEF 2020 presents a more mixed outcome. On the training set, all three agents complete 45/45 valid runs, compared with 39/45 for the original ALNS. On the 12 commonly valid training instances, KIMI and DeepSeek obtain mean objectives of 14946.02 and 14968.73, below the original ALNS's 15130.78, whereas GLM obtains 15799.42 (Section~\ref{sec:r20-scratch}). On the test set, all three agents cover all 15 instances, compared with nine for the original ALNS; KIMI completes 44/45 valid runs, while GLM and DeepSeek each complete 45/45 (Table~\ref{tab:final-R20}). On the six commonly valid test instances, KIMI and DeepSeek obtain mean objectives of 16696.18 and 16740.43, outperforming the original ALNS's 17166.21, whereas GLM obtains 17887.77 (Table~\ref{tab:final-R20}). Some agents can therefore build algorithms from scratch that outperform the supplied human reference on this industrial problem, but this advantage is not consistent across models.

ROADEF 2022 mirrors the ROADEF 2018 pattern of strong feasibility paired with a large cost deficit. All three scratch solvers achieve 150/150 valid training runs and 315/315 valid held-out runs across every instance, surpassing V6's coverage, yet their common-subset test mean gaps remain between 21.7191\% (KIMI) and 63.4174\% (GLM)---at least 21.30 percentage points above V6's 0.4188\% (Table~\ref{tab:r22-scratch-X}). Satisfying the complex 3D loading and axle-weight rules is thus well within reach of scratch-built solvers, whereas matching a mature MIP-based matheuristic on joint transport and inventory cost remains far more challenging.

Overall, the ability to develop algorithms from scratch depends strongly on the problem domain: agents design strong algorithms on PFSP, remain substantially behind mature human methods on TSP, ROADEF 2018 and ROADEF 2022, and show model-dependent results on ROADEF 2020. These results indicate that constructing executable, feasible solvers is increasingly achievable, while independently developing advanced algorithmic structures for complex problems remains an important challenge for coding agents.

\subsubsection{Do coding agents deliver substantive algorithmic improvements?}

The performance gains delivered by coding agents do not always reflect the same kind of algorithmic change. On PFSP, where the algorithmic structure is relatively simple, agents reorganize search, modify key operators, and construct solvers from scratch that substantially outperform the human-designed algorithms compared here, demonstrating algorithm-design capabilities beyond parameter adjustment alone. For a mature, complex algorithm such as EAX, however, gains over the original implementation may also be obtained through simple manual parameter changes, without evidence that the agents improve its core algorithmic mechanism.

PFSP demonstrates substantive changes to algorithmic mechanisms. On the test set, the original IG and Basic ALNS achieve mean gaps of 2.053454\% and 2.346119\%, respectively, whereas the four agents improving existing algorithms obtain 0.450912--1.076633\%, and the three agents developing from scratch further obtain $-0.456489$\% to $-0.222135$\% (Tables~\ref{tab:igresults} and~\ref{tab:final-PFSP}). These gains accompany changes to search mechanisms rather than parameter adjustment alone: GPT combines adaptive destruction with late acceptance and final-budget intensification; GLM jointly modifies neighborhood evaluation, intensification, perturbation, and acceptance strategies; KIMI organizes perturbed NEH initialization and destruction around critical-path structure; and DeepSeek modifies multi-start search and reconstruction. Agents can therefore combine and adapt search mechanisms around problem structure to develop competitive algorithms.

The EAX results show that a substantial performance gain need not represent a breakthrough in the algorithmic core. All four agents retain the core EAX crossover mechanism, with changes concentrated on population configuration, phase scheduling, initialization, crossover selection, and local intensification. Although these variants reduce the test mean gap from the original algorithm's 2.240530\% to 0.014367--0.073410\%, simply changing the original algorithm's population size from 300 to 100 achieves 0.014315\%, slightly below the strongest agent (Table~\ref{tab:eaxresults}). In these experiments, the agents' EAX modifications therefore do not demonstrate additional gains beyond a simple configuration adjustment, nor do they further improve the performance of its mature core mechanism.

Evaluating coding agents' algorithm-design capabilities thus requires examining both performance changes and the content of their modifications. PFSP demonstrates their ability to reorganize search mechanisms and construct effective algorithms, whereas EAX shows that large gains over a default implementation may primarily reflect discovering a stronger configuration rather than improving the underlying algorithmic core.

\section{Conclusions}

\label{sec:discussion}

LLMs have advanced automatic algorithm design (AAD)---especially within its subfield of automatic heuristic design (AHD)---while general-purpose coding agents capable of autonomous planning, code modification, and tool use offer a new paradigm for solver-level AAD in combinatorial optimization. This study establishes a unified development and testing framework to assess their practical algorithmic competitiveness on large-scale, complex discrete optimization problems under two common scenarios: improving existing algorithms and developing solvers from scratch. The experiments cover two classical CO problems, PFSP and TSP, together with the industrial glass-cutting, grid-maintenance, and truck-loading problems from the 2018, 2020, and 2022 ROADEF/EURO Challenges. By systematically benchmarking four coding agents paired with frontier LLMs---Codex (GPT-6-Astra), Kimi-code (KIMI-K3), Zcode (GLM-5.3), and DeepSeek Harness (DeepSeek-V4-Flash)---against fixed-workflow baselines (ReEvo and OpenEvolve) and strong human reference algorithms on training and held-out test instances, we address five research questions concerning competitiveness against fixed workflows, improvement of existing algorithms, generalization of those improvements, design from scratch, and the sources of performance gains.

The results show that general-purpose coding agents can autonomously select modification targets and development steps while matching or outperforming the fixed-workflow methods evaluated. They outperform the ReEvo adaptation, which requires a human-specified optimization target, on PFSP, and outperform OpenEvolve using the same underlying model on TSP/EAX as well as on the evaluated industrial tasks (ROADEF 2018 and ROADEF 2020), where OpenEvolve either reverts to the unmodified champion or trails on the commonly valid held-out instances. When improving existing algorithms, agents further reduce training gaps on classical CO problems and improve both solution quality and solving success rates on industrial problems. These gains persist on unseen instances across all five domains---including reducing held-out costs and expanding valid-solve coverage on ROADEF 2022---demonstrating that agents can leverage execution feedback to enhance mature expert designs. However, training advantages do not always transfer fully to testing: model rankings can change, and some solvers encounter feasibility or resource-compliance problems.

Development from scratch further reveals the boundaries of current capabilities. On PFSP, agents construct algorithms that substantially outperform the IG and basic ALNS references compared here, obtaining objective values below the supplied upper bounds in many runs. On TSP, ROADEF 2018, and ROADEF 2022, however, solvers developed from scratch remain substantially behind mature human-designed algorithms in solution quality, even when achieving complete feasibility across every held-out run. ROADEF 2020 presents a different outcome: some scratch solvers achieve broader valid-instance coverage and outperform the human reference on commonly valid instances. These results indicate that design from scratch is strongly problem-dependent, and success on some tasks---or reliable constraint satisfaction alone---does not establish a general ability to handle complex algorithmic structures and problem constraints. Moreover, performance gains do not necessarily imply advances in algorithmic mechanisms. PFSP improvements involve reorganizing the search process and key operators, whereas on EAX, simply adjusting the original algorithm's population size achieves performance comparable to the best agent-developed variant. Assessing algorithm-design capability therefore requires considering final performance, reasonably configured human controls, and the actual code changes together to determine the sources of gains.

These findings suggest three directions for further research. First, agents need stronger capabilities for understanding complex solver structures and diagnosing performance, enabling them to move beyond configuration changes and local implementation edits toward the design of core search mechanisms. Second, improving their ability to develop complex algorithms from scratch requires studying how domain knowledge, algorithmic literature, and execution feedback can be used effectively to construct competitive algorithmic structures. Finally, development should incorporate stronger validation of generalization and reliability so that algorithmic improvements remain effective on unseen instances with larger scales, different distributions, and complex constraints.

This study shows that general-purpose coding agents can already serve as effective tools for improving CO algorithms. Constructing complex high-performance algorithms from scratch and advancing the core mechanisms of mature solvers nevertheless remain important challenges. Future work should examine these findings across a broader range of problems and algorithmic systems to further establish the scope and limits of their capabilities.

\clearpage

\begingroup
\raggedright
\bibliographystyle{unsrtnat}
\bibliography{references}
\endgroup
\clearpage
\ifshowappendices
\appendix
\section{Instance-level held-out results}
\label{app:results}
\label{app:heldout-expanded}
The following tables consolidate the test results by problem. Each table presents human references, tested workflow methods where applicable, and coding agents under both development scenarios on the same instance rows. In the first four tables, each numeric cell shows the mean above the best value over the prescribed repeats. Bold indicates the lowest reported value across methods for each instance, separately for the mean and best rows; ties at the displayed precision are all highlighted. Blank cells indicate unavailable instance-level measurements; they do not denote zero. Incomplete evaluations are marked NC. DS Harness abbreviates DeepSeek Harness. Instance names span the mean and best rows. Each centered method heading includes its model label where applicable.

\subsection{Permutation flow-shop scheduling}
\label{app:test-pfsp}
\begingroup\fontsize{7}{9}\selectfont\setlength{\LTcapwidth}{\linewidth}\setlength{\tabcolsep}{1.6pt}\renewcommand{\arraystretch}{1.13}
\begin{longtable}{@{}cccccccccccc@{}}
\caption{Permutation flow-shop scheduling: held-out gap percentages; ten valid runs per populated cell. OpenEvolve uses the frozen round046 candidate from the main text; all 1,800 test solves are valid. Its aggregate mean / best gap is 0.9105\% / 0.6770\%.}\label{tab:appendix-pfsp}\\
\toprule
\multirow{3}{*}{Instance} & \multicolumn{2}{c}{Human reference} & \multicolumn{2}{c}{Workflow methods} & \multicolumn{4}{c}{Coding agents: improve} & \multicolumn{3}{c}{Coding agents: scratch} \\
\cmidrule(lr){2-3}\cmidrule(lr){4-5}\cmidrule(lr){6-9}\cmidrule(lr){10-12}
 & \multicolumn{1}{c}{\multirow{2}{*}{IG}} & \multicolumn{1}{c}{\multirow{2}{*}{Basic ALNS}} & \multicolumn{1}{c}{\multirow{2}{*}{\shortstack{ReEvo\\DeepSeek}}} & \multicolumn{1}{c}{\multirow{2}{*}{\shortstack{OpenEvolve\\DeepSeek}}} & \multicolumn{1}{c}{\multirow{2}{*}{\shortstack{Codex\\GPT}}} & \multicolumn{1}{c}{\multirow{2}{*}{\shortstack{Kimi-code\\KIMI}}} & \multicolumn{1}{c}{\multirow{2}{*}{\shortstack{Zcode\\GLM}}} & \multicolumn{1}{c}{\multirow{2}{*}{\shortstack{DS Harness\\DeepSeek}}} & \multicolumn{1}{c}{\multirow{2}{*}{\shortstack{Kimi-code\\KIMI}}} & \multicolumn{1}{c}{\multirow{2}{*}{\shortstack{Zcode\\GLM}}} & \multicolumn{1}{c}{\multirow{2}{*}{\shortstack{DS Harness\\DeepSeek}}} \\*
 &  &  &  &  &  &  &  &  &  &  &  \\ \midrule
\endfirsthead
\multicolumn{12}{c}{\tablename\ \thetable\ (continued)}\\
\toprule
\multirow{3}{*}{Instance} & \multicolumn{2}{c}{Human reference} & \multicolumn{2}{c}{Workflow methods} & \multicolumn{4}{c}{Coding agents: improve} & \multicolumn{3}{c}{Coding agents: scratch} \\
\cmidrule(lr){2-3}\cmidrule(lr){4-5}\cmidrule(lr){6-9}\cmidrule(lr){10-12}
 & \multicolumn{1}{c}{\multirow{2}{*}{IG}} & \multicolumn{1}{c}{\multirow{2}{*}{Basic ALNS}} & \multicolumn{1}{c}{\multirow{2}{*}{\shortstack{ReEvo\\DeepSeek}}} & \multicolumn{1}{c}{\multirow{2}{*}{\shortstack{OpenEvolve\\DeepSeek}}} & \multicolumn{1}{c}{\multirow{2}{*}{\shortstack{Codex\\GPT}}} & \multicolumn{1}{c}{\multirow{2}{*}{\shortstack{Kimi-code\\KIMI}}} & \multicolumn{1}{c}{\multirow{2}{*}{\shortstack{Zcode\\GLM}}} & \multicolumn{1}{c}{\multirow{2}{*}{\shortstack{DS Harness\\DeepSeek}}} & \multicolumn{1}{c}{\multirow{2}{*}{\shortstack{Kimi-code\\KIMI}}} & \multicolumn{1}{c}{\multirow{2}{*}{\shortstack{Zcode\\GLM}}} & \multicolumn{1}{c}{\multirow{2}{*}{\shortstack{DS Harness\\DeepSeek}}} \\*
 &  &  &  &  &  &  &  &  &  &  &  \\ \midrule
\endhead
\midrule\multicolumn{12}{c}{Continued on next page}\\\endfoot
\bottomrule\endlastfoot
\multirow{2}{*}{VFR300\_20\_1} & 2.0020 & 1.7239 & 2.0590 & 0.7852 & 0.5815 & 0.6459 & 0.4421 & 1.0106 & -0.0737 & \textbf{-0.1647} & -0.1461 \\*
 & 1.6595 & 1.2880 & 1.7277 & 0.4458 & 0.2105 & 0.4954 & 0.2043 & 0.8483 & \textbf{-0.3715} & -0.2972 & -0.2105 \\[2pt]
\multirow{2}{*}{VFR300\_20\_2} & 1.5201 & 1.4353 & 1.6982 & 0.7013 & 0.5420 & 0.4046 & 0.3500 & 0.9181 & -0.0430 & \textbf{-0.0751} & 0.0357 \\*
 & 1.2597 & 0.7389 & 1.4474 & 0.5390 & 0.3210 & 0.1635 & 0.1575 & 0.6117 & -0.1030 & \textbf{-0.1938} & -0.0303 \\[2pt]
\multirow{2}{*}{VFR300\_20\_3} & 1.9056 & 1.9458 & 2.1350 & 0.9738 & 0.6857 & 0.6665 & 0.4804 & 1.3980 & -0.0927 & \textbf{-0.1911} & -0.0476 \\*
 & 1.4098 & 1.5953 & 1.9168 & 0.6183 & 0.5874 & 0.3772 & 0.2782 & 0.9955 & -0.2968 & \textbf{-0.3092} & -0.1237 \\[2pt]
\multirow{2}{*}{VFR300\_20\_4} & 2.1501 & 1.8262 & 2.0883 & 0.8331 & 0.5587 & 0.6860 & 0.5167 & 1.1977 & 0.0470 & \textbf{-0.1038} & 0.1063 \\*
 & 1.9282 & 1.5141 & 1.7181 & 0.6674 & 0.2966 & 0.5253 & 0.3399 & 0.9456 & -0.1607 & \textbf{-0.2348} & -0.1978 \\
\end{longtable}\endgroup

\clearpage\subsection{Traveling salesman problem}
\label{app:test-tsp}
\begingroup\fontsize{8}{10}\selectfont\setlength{\LTcapwidth}{\linewidth}\setlength{\tabcolsep}{2.5pt}\renewcommand{\arraystretch}{1.13}\begin{longtable}{@{}ccccccccccc@{}}
\caption{Traveling salesman problem: held-out gap percentages; ten valid runs per populated cell. OpenEvolve uses round038, the same candidate as the main text.}\label{tab:appendix-tsp}\\
\toprule
\multirow{3}{*}{Instance} & \multicolumn{2}{c}{Human reference} & \multicolumn{1}{c}{Workflow methods} & \multicolumn{4}{c}{Coding agents: improve} & \multicolumn{3}{c}{Coding agents: scratch} \\
\cmidrule(lr){2-3}\cmidrule(lr){4-4}\cmidrule(lr){5-8}\cmidrule(lr){9-11}
 & \multicolumn{1}{c}{\multirow{2}{*}{EAX 300}} & \multicolumn{1}{c}{\multirow{2}{*}{EAX 100}} & \multicolumn{1}{c}{\multirow{2}{*}{\shortstack{OpenEvolve\\DeepSeek}}} & \multicolumn{1}{c}{\multirow{2}{*}{\shortstack{Codex\\GPT}}} & \multicolumn{1}{c}{\multirow{2}{*}{\shortstack{Kimi-code\\KIMI}}} & \multicolumn{1}{c}{\multirow{2}{*}{\shortstack{Zcode\\GLM}}} & \multicolumn{1}{c}{\multirow{2}{*}{\shortstack{DS Harness\\DeepSeek}}} & \multicolumn{1}{c}{\multirow{2}{*}{\shortstack{Kimi-code\\KIMI}}} & \multicolumn{1}{c}{\multirow{2}{*}{\shortstack{Zcode\\GLM}}} & \multicolumn{1}{c}{\multirow{2}{*}{\shortstack{DS Harness\\DeepSeek}}} \\*
 &  &  &  &  &  &  &  &  &  &  \\ \midrule
\endfirsthead
\multicolumn{11}{c}{\tablename\ \thetable\ (continued)}\\
\toprule
\multirow{3}{*}{Instance} & \multicolumn{2}{c}{Human reference} & \multicolumn{1}{c}{Workflow methods} & \multicolumn{4}{c}{Coding agents: improve} & \multicolumn{3}{c}{Coding agents: scratch} \\
\cmidrule(lr){2-3}\cmidrule(lr){4-4}\cmidrule(lr){5-8}\cmidrule(lr){9-11}
 & \multicolumn{1}{c}{\multirow{2}{*}{EAX 300}} & \multicolumn{1}{c}{\multirow{2}{*}{EAX 100}} & \multicolumn{1}{c}{\multirow{2}{*}{\shortstack{OpenEvolve\\DeepSeek}}} & \multicolumn{1}{c}{\multirow{2}{*}{\shortstack{Codex\\GPT}}} & \multicolumn{1}{c}{\multirow{2}{*}{\shortstack{Kimi-code\\KIMI}}} & \multicolumn{1}{c}{\multirow{2}{*}{\shortstack{Zcode\\GLM}}} & \multicolumn{1}{c}{\multirow{2}{*}{\shortstack{DS Harness\\DeepSeek}}} & \multicolumn{1}{c}{\multirow{2}{*}{\shortstack{Kimi-code\\KIMI}}} & \multicolumn{1}{c}{\multirow{2}{*}{\shortstack{Zcode\\GLM}}} & \multicolumn{1}{c}{\multirow{2}{*}{\shortstack{DS Harness\\DeepSeek}}} \\*
 &  &  &  &  &  &  &  &  &  &  \\ \midrule
\endhead
\midrule\multicolumn{11}{c}{Continued on next page}\\\endfoot
\bottomrule\endlastfoot
\multirow{2}{*}{ido21215} & 0.0148 & \textbf{0.0069} & 0.0400 & 0.0072 & 0.0082 & 0.0072 & 0.0102 & 0.6373 & 2.1004 & 0.3608 \\*
 & \textbf{0.0000} & 0.0016 & 0.0331 & 0.0016 & 0.0031 & \textbf{0.0000} & 0.0063 & 0.5652 & 1.7271 & 0.3133 \\[2pt]
\multirow{2}{*}{fma21553} & \textbf{0.0015} & 0.0054 & 0.0305 & 0.0053 & 0.0044 & 0.0023 & 0.0066 & 0.6812 & 1.8828 & 0.3602 \\*
 & \textbf{0.0000} & 0.0030 & 0.0195 & 0.0030 & 0.0015 & \textbf{0.0000} & 0.0030 & 0.5592 & 1.5302 & 0.2931 \\[2pt]
\multirow{2}{*}{vm22775} & 0.3289 & \textbf{0.0045} & 0.0284 & 0.0054 & 0.0047 & 0.0451 & 0.0053 & 0.5078 & 1.9587 & 0.3325 \\*
 & 0.0197 & 0.0035 & 0.0197 & 0.0037 & \textbf{0.0030} & 0.0074 & 0.0040 & 0.4356 & 1.6844 & 0.2881 \\[2pt]
\multirow{2}{*}{lsb22777} & 0.0215 & \textbf{0.0052} & 0.0335 & 0.0062 & 0.0062 & 0.0072 & 0.0069 & 0.7134 & 2.2118 & 0.4067 \\*
 & \textbf{0.0000} & \textbf{0.0000} & 0.0213 & 0.0033 & 0.0016 & \textbf{0.0000} & 0.0016 & 0.6396 & 1.8056 & 0.2870 \\
\end{longtable}\endgroup

\clearpage\subsection{ROADEF 2018 glass cutting}
\label{app:test-r18}
\begingroup\fontsize{8}{10}\selectfont\setlength{\LTcapwidth}{\linewidth}\setlength{\tabcolsep}{2.5pt}\renewcommand{\arraystretch}{1.13}\begin{longtable}{@{}ccccccccc@{}}
\caption{ROADEF 2018 glass cutting: held-out waste percentages; three checker-valid runs per populated cell. Zcode/GLM improvement observations include the memory violations reported in the main text.}\label{tab:appendix-r18}\\
\toprule
\multirow{3}{*}{Instance} & \multicolumn{1}{c}{Human reference} & Workflow & \multicolumn{3}{c}{Coding agents: improve} & \multicolumn{3}{c}{Coding agents: scratch} \\
\cmidrule(lr){2-2}\cmidrule(lr){3-3}\cmidrule(lr){4-6}\cmidrule(lr){7-9}
 & \multicolumn{1}{c}{\multirow{2}{*}{Champion}} & \multirow{2}{*}{\shortstack{OpenEvolve\\DeepSeek}} & \multicolumn{1}{c}{\multirow{2}{*}{\shortstack{Kimi-code\\KIMI}}} & \multicolumn{1}{c}{\multirow{2}{*}{\shortstack{Zcode\\GLM}}} & \multicolumn{1}{c}{\multirow{2}{*}{\shortstack{DS Harness\\DeepSeek}}} & \multicolumn{1}{c}{\multirow{2}{*}{\shortstack{Kimi-code\\KIMI}}} & \multicolumn{1}{c}{\multirow{2}{*}{\shortstack{Zcode\\GLM}}} & \multicolumn{1}{c}{\multirow{2}{*}{\shortstack{DS Harness\\DeepSeek}}} \\*
 &  &  &  &  &  &  &  &  \\ \midrule
\endfirsthead
\multicolumn{9}{c}{\tablename\ \thetable\ (continued)}\\
\toprule
\multirow{3}{*}{Instance} & \multicolumn{1}{c}{Human reference} & Workflow & \multicolumn{3}{c}{Coding agents: improve} & \multicolumn{3}{c}{Coding agents: scratch} \\
\cmidrule(lr){2-2}\cmidrule(lr){3-3}\cmidrule(lr){4-6}\cmidrule(lr){7-9}
 & \multicolumn{1}{c}{\multirow{2}{*}{Champion}} & \multirow{2}{*}{\shortstack{OpenEvolve\\DeepSeek}} & \multicolumn{1}{c}{\multirow{2}{*}{\shortstack{Kimi-code\\KIMI}}} & \multicolumn{1}{c}{\multirow{2}{*}{\shortstack{Zcode\\GLM}}} & \multicolumn{1}{c}{\multirow{2}{*}{\shortstack{DS Harness\\DeepSeek}}} & \multicolumn{1}{c}{\multirow{2}{*}{\shortstack{Kimi-code\\KIMI}}} & \multicolumn{1}{c}{\multirow{2}{*}{\shortstack{Zcode\\GLM}}} & \multicolumn{1}{c}{\multirow{2}{*}{\shortstack{DS Harness\\DeepSeek}}} \\*
 &  &  &  &  &  &  &  &  \\ \midrule
\endhead
\midrule\multicolumn{9}{c}{Continued on next page}\\\endfoot
\bottomrule\endlastfoot
\multirow{2}{*}{X1} & 5.3034 & 5.3034 & \textbf{5.1198} & 5.4817 & 5.3034 & 25.1052 & 19.3352 & 20.3008 \\*
 & 5.3034 & 5.3034 & \textbf{5.1198} & 5.4817 & 5.3034 & 25.1049 & 18.9064 & 20.1703 \\[2pt]
\multirow{2}{*}{X2} & 3.2113 & 3.2113 & \textbf{2.9386} & 3.3257 & 3.1581 & 23.5172 & 14.5601 & 20.5555 \\*
 & 3.2113 & 3.2113 & \textbf{2.9386} & 3.3257 & 3.1581 & 23.5172 & 14.4943 & 19.8580 \\[2pt]
\multirow{2}{*}{X3} & 4.3783 & 4.3783 & 4.2396 & 4.2077 & \textbf{4.1457} & 18.4381 & 15.7734 & 24.3267 \\*
 & 4.3783 & 4.3783 & 4.2396 & 4.2077 & \textbf{4.1457} & 18.4372 & 15.2789 & 24.3267 \\[2pt]
\multirow{2}{*}{X4} & 5.2495 & 5.2495 & \textbf{4.8320} & 5.2495 & 5.2495 & 23.1694 & 14.7590 & 18.4006 \\*
 & 5.2495 & 5.2495 & \textbf{4.8320} & 5.2495 & 5.2495 & 21.6296 & 14.4055 & 18.4006 \\
\end{longtable}\endgroup

\clearpage\subsection{ROADEF 2020 maintenance scheduling}
\label{app:test-r20}
\begingroup\fontsize{7.5}{9.5}\selectfont\setlength{\LTcapwidth}{\linewidth}\setlength{\tabcolsep}{2.5pt}\renewcommand{\arraystretch}{1.13}\begin{longtable}{@{}ccccccccc@{}}
\caption{ROADEF 2020 maintenance scheduling: held-out official risk objectives. Each complete result shows the mean above the best objective over three runs. NC means at least one required run is invalid; no valid-only average is substituted.}\label{tab:appendix-r20}\\
\toprule
\multirow{3}{*}{Instance} & \multicolumn{1}{c}{Human reference} & Workflow & \multicolumn{3}{c}{Coding agents: improve} & \multicolumn{3}{c}{Coding agents: scratch} \\
\cmidrule(lr){2-2}\cmidrule(lr){3-3}\cmidrule(lr){4-6}\cmidrule(lr){7-9}
 & \multicolumn{1}{c}{\multirow{2}{*}{ALNS}} & \multirow{2}{*}{\shortstack{OpenEvolve\\DeepSeek}} & \multicolumn{1}{c}{\multirow{2}{*}{\shortstack{Kimi-code\\KIMI}}} & \multicolumn{1}{c}{\multirow{2}{*}{\shortstack{Zcode\\GLM}}} & \multicolumn{1}{c}{\multirow{2}{*}{\shortstack{DS Harness\\DeepSeek}}} & \multicolumn{1}{c}{\multirow{2}{*}{\shortstack{Kimi-code\\KIMI}}} & \multicolumn{1}{c}{\multirow{2}{*}{\shortstack{Zcode\\GLM}}} & \multicolumn{1}{c}{\multirow{2}{*}{\shortstack{DS Harness\\DeepSeek}}} \\*
 &  &  &  &  &  &  &  &  \\ \midrule
\endfirsthead
\multicolumn{9}{c}{\tablename\ \thetable\ (continued)}\\
\toprule
\multirow{3}{*}{Instance} & \multicolumn{1}{c}{Human reference} & Workflow & \multicolumn{3}{c}{Coding agents: improve} & \multicolumn{3}{c}{Coding agents: scratch} \\
\cmidrule(lr){2-2}\cmidrule(lr){3-3}\cmidrule(lr){4-6}\cmidrule(lr){7-9}
 & \multicolumn{1}{c}{\multirow{2}{*}{ALNS}} & \multirow{2}{*}{\shortstack{OpenEvolve\\DeepSeek}} & \multicolumn{1}{c}{\multirow{2}{*}{\shortstack{Kimi-code\\KIMI}}} & \multicolumn{1}{c}{\multirow{2}{*}{\shortstack{Zcode\\GLM}}} & \multicolumn{1}{c}{\multirow{2}{*}{\shortstack{DS Harness\\DeepSeek}}} & \multicolumn{1}{c}{\multirow{2}{*}{\shortstack{Kimi-code\\KIMI}}} & \multicolumn{1}{c}{\multirow{2}{*}{\shortstack{Zcode\\GLM}}} & \multicolumn{1}{c}{\multirow{2}{*}{\shortstack{DS Harness\\DeepSeek}}} \\*
 &  &  &  &  &  &  &  &  \\ \midrule
\endhead
\midrule\multicolumn{9}{c}{Continued on next page}\\\endfoot
\bottomrule\endlastfoot
\multirow{2}{*}{X\_01} & 4101.64 & 4097.39 & 4087.51 & 4096.00 & 4090.80 & 4055.38 & 4124.74 & \textbf{4040.84} \\*
 & 4094.25 & 4093.21 & 4079.55 & 4090.28 & 4074.59 & 4049.21 & 4105.31 & \textbf{4036.11} \\[2pt]
\multirow{2}{*}{X\_02} & 34115.99 & 33645.84 & 33305.64 & 33321.85 & 33320.68 & 32652.84 & 36107.64 & \textbf{32556.13} \\*
 & 33960.34 & 33582.32 & 33274.32 & 33260.52 & 33242.46 & 32618.88 & 36059.17 & \textbf{32533.22} \\[2pt]
\multirow{2}{*}{X\_03} & \multirow{2}{*}{NC} & \multirow{2}{*}{NC} & \multirow{2}{*}{NC} & \multirow{2}{*}{NC} & \multirow{2}{*}{NC} & \textbf{8193.09} & 8396.22 & 8212.48 \\*
 &  &  &  &  &  & \textbf{8180.92} & 8382.34 & 8210.66 \\[2pt]
\multirow{2}{*}{X\_04} & \multirow{2}{*}{NC} & \multirow{2}{*}{NC} & 11621.81 & 11702.60 & \multirow{2}{*}{NC} & \textbf{11455.33} & 11818.75 & 11462.37 \\*
 &  &  & 11587.00 & 11662.65 &  & \textbf{11447.32} & 11800.47 & 11451.20 \\
\end{longtable}\endgroup

\clearpage\subsection{ROADEF 2022 truck loading}
\label{app:test-r22}
Tables~\ref{tab:appendix-r22-improve} and~\ref{tab:appendix-r22-scratch} use the same 105 held-out instances and the same instance-specific contest costs. Each complete cell gives mean / best percentage gap over three repeats. An incomplete cell reports the number of eligible repeats. Identifiers refer to set X.

\begingroup\fontsize{8}{10}\selectfont\setlength{\tabcolsep}{3pt}\setlength{\LTcapwidth}{\linewidth}
\begin{longtable}{@{}lrrrrr@{}}
\caption{ROADEF 2022, improvement of S46. Held-out costs are normalized by the final-phase workbook reference shown in the second column.}\label{tab:appendix-r22-improve}\\
\toprule
Instance & Reference & S46--HiGHS V6 & Kimi-code & Zcode & DeepSeek Harness \\ \midrule
\endfirsthead
\multicolumn{6}{c}{\tablename\ \thetable\ (continued)}\\\toprule
Instance & Reference & S46--HiGHS V6 & Kimi-code & Zcode & DeepSeek Harness \\ \midrule
\endhead
\bottomrule\endfoot
AS & 1,912,475 & 0.02 / -0.03 & -0.58 / -0.67 & -0.37 / -0.42 & -0.30 / -0.31 \\
AS2 & 2,083,300 & 0.42 / 0.34 & -0.05 / -0.10 & 0.21 / 0.11 & 0.25 / 0.15 \\
AS3 & 9,196,363 & -0.76 / -0.78 & -0.92 / -1.09 & -0.95 / -1.00 & -1.10 / -1.14 \\
AS4 & 1,940,261 & -0.07 / -0.11 & -0.48 / -0.55 & -0.51 / -0.57 & -0.49 / -0.56 \\
AS5 & 2,113,325 & 0.47 / 0.39 & 0.18 / 0.18 & 0.43 / 0.42 & 0.24 / 0.23 \\
AS6 & 9,297,347 & -0.11 / -0.17 & -0.53 / -0.65 & -0.36 / -0.38 & -0.76 / -0.86 \\
AS7 & 1,914,552 & -0.15 / -0.17 & -0.89 / -0.98 & -0.63 / -0.65 & -0.73 / -0.76 \\
BY & 9,276,754 & -0.47 / -0.54 & -1.33 / -1.39 & -0.34 / -0.40 & -0.70 / -0.70 \\
\end{longtable}\endgroup

\begingroup\fontsize{8}{10}\selectfont\setlength{\tabcolsep}{3pt}\setlength{\LTcapwidth}{\linewidth}
\begin{longtable}{@{}lrrrrr@{}}
\caption{ROADEF 2022, development from scratch. Held-out costs are normalized by the final-phase workbook reference shown in the second column.}\label{tab:appendix-r22-scratch}\\
\toprule
Instance & Reference & S46--HiGHS V6 & Kimi-code & Zcode & DeepSeek Harness \\ \midrule
\endfirsthead
\multicolumn{6}{c}{\tablename\ \thetable\ (continued)}\\\toprule
Instance & Reference & S46--HiGHS V6 & Kimi-code & Zcode & DeepSeek Harness \\ \midrule
\endhead
\bottomrule\endfoot
AS & 1,912,475 & 0.02 / -0.03 & 18.91 / 18.84 & 47.80 / 47.62 & 21.08 / 21.08 \\
AS2 & 2,083,300 & 0.42 / 0.34 & 17.24 / 17.11 & 49.44 / 49.02 & 21.12 / 21.12 \\
AS3 & 9,196,363 & -0.76 / -0.78 & 19.48 / 19.37 & 47.83 / 47.50 & 22.10 / 22.10 \\
AS4 & 1,940,261 & -0.07 / -0.11 & 19.06 / 18.96 & 46.96 / 46.57 & 20.75 / 20.75 \\
AS5 & 2,113,325 & 0.47 / 0.39 & 16.70 / 16.52 & 48.39 / 46.56 & 21.02 / 21.02 \\
AS6 & 9,297,347 & -0.11 / -0.17 & 20.09 / 19.99 & 47.32 / 47.27 & 21.87 / 21.87 \\
AS7 & 1,914,552 & -0.15 / -0.17 & 17.18 / 17.06 & 40.35 / 39.84 & 19.02 / 19.02 \\
BY & 9,276,754 & -0.47 / -0.54 & 18.66 / 18.66 & 49.97 / 49.97 & 27.77 / 27.77 \\
\end{longtable}\endgroup

\clearpage
\section{Development trajectories and execution accounting}
\label{app:development}
\subsection{Complete retained development trajectories}
\label{app:all-trajectories}
The figures are ordered by problem: PFSP, TSP, ROADEF 2018, ROADEF 2020 and ROADEF 2022, with improvement followed by development from scratch for each problem. All use the same two-column panel style. Faint points and lines show candidate scores; solid steps show the running minimum. Crosses on the lower margin mark unavailable, invalid or disputed scores, not zero-valued outcomes. Panels within each figure share a quality range; EAX improvement uses a logarithmic axis. Curves are shown without extrapolation. Repeated evaluations remain visible and are not independent development campaigns. These are training trajectories, not the frozen-solver test measurements in Appendix~\ref{app:results}.
\begin{figure}[htbp]
\centering
\includegraphics[width=\linewidth]{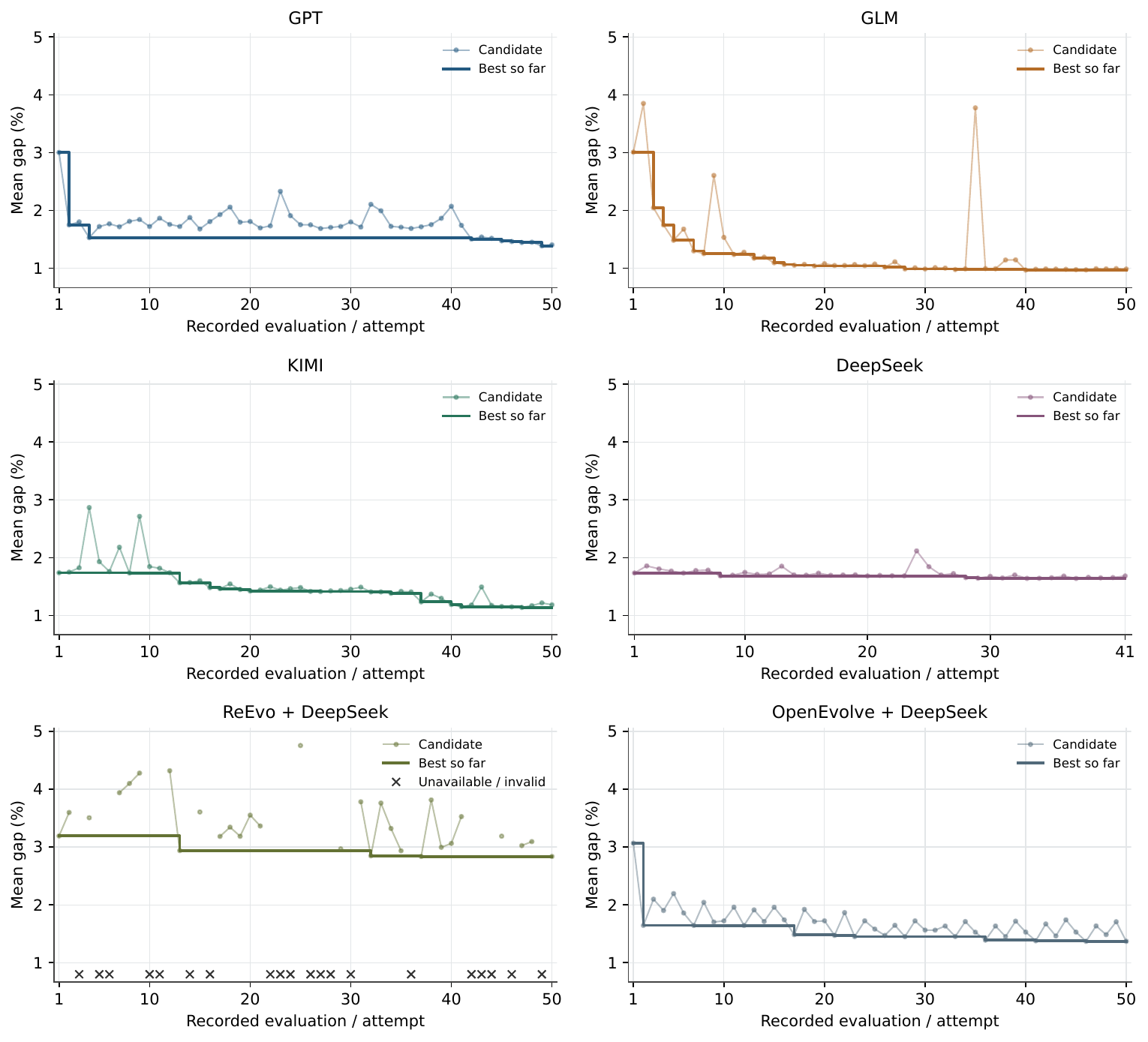}
\caption{PFSP: improvement of IG. DeepSeek ends at its recorded attempt 41. ReEvo has 20 failed attempts within the displayed range. OpenEvolve shows the 50-attempt, 65,536-token campaign whose selected candidate is round046. Editable scopes differ.}
\label{fig:appendix-IG-improve}
\end{figure}
\clearpage
\begin{figure}[htbp]
\centering
\includegraphics[width=\linewidth]{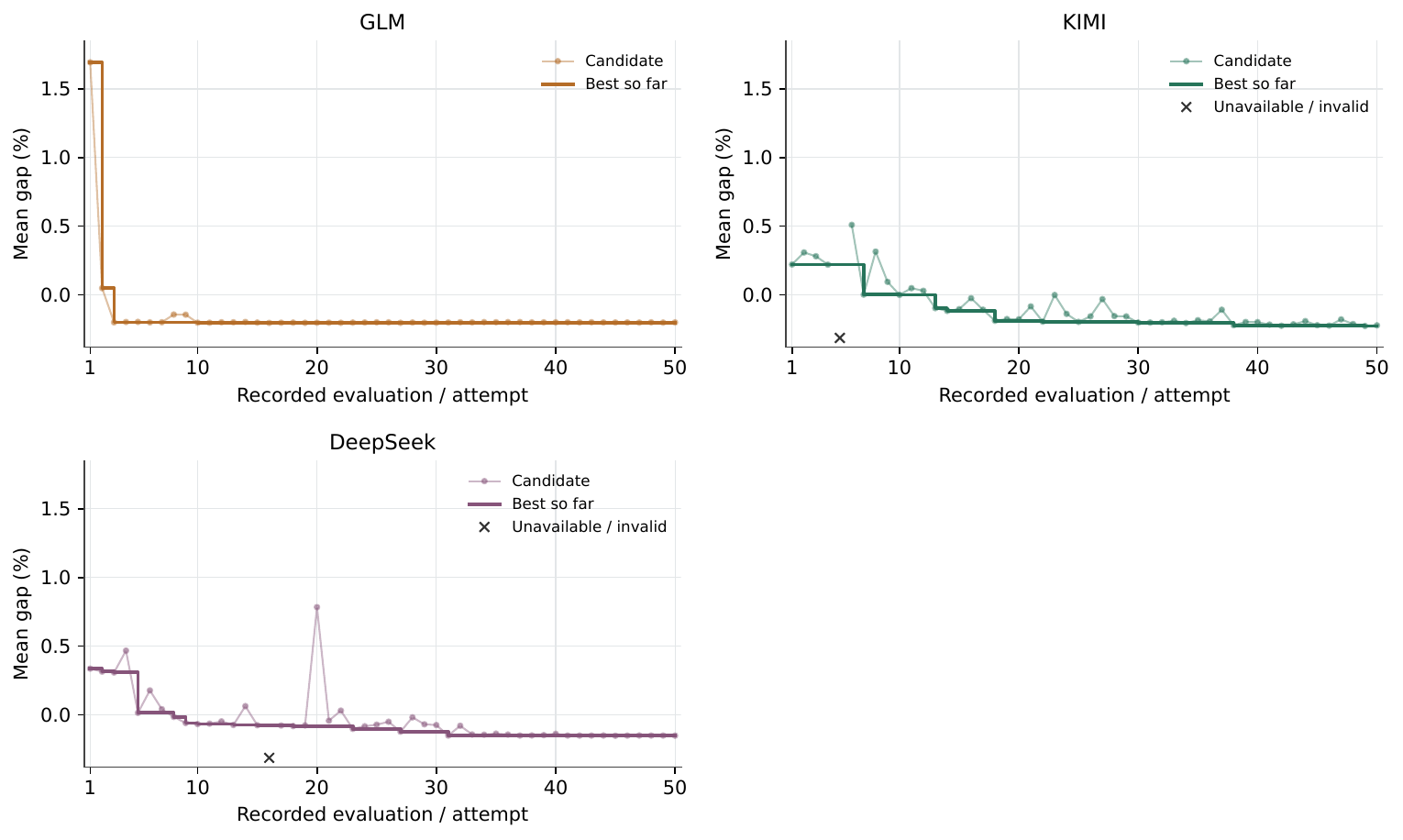}
\caption{PFSP: solver development from scratch. Each agent has 50 recorded evaluations of 600 solves. KIMI round005 and DeepSeek round016 are retained as failed evaluations; their scores do not update the incumbent.}
\label{fig:appendix-PFSP-scratch}
\end{figure}
\clearpage
\begin{figure}[htbp]
\centering
\includegraphics[width=\linewidth]{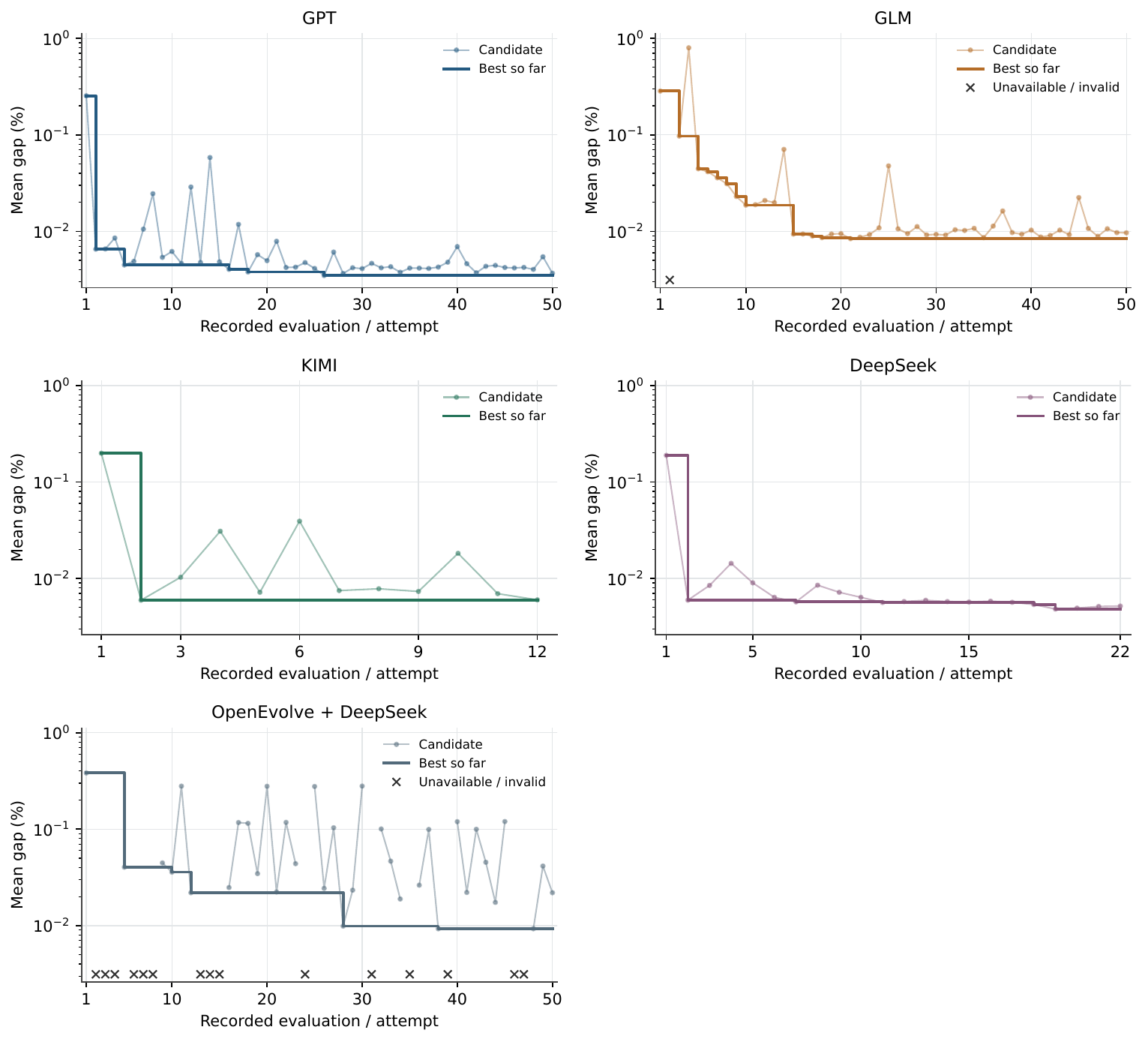}
\caption{TSP: improvement of EAX. Four general agents and OpenEvolve are shown; unavailable or invalid attempts remain marked. The logarithmic quality axis is shared across panels.}
\label{fig:appendix-EAX-improve}
\end{figure}
\clearpage
\begin{figure}[htbp]
\centering
\includegraphics[width=\linewidth]{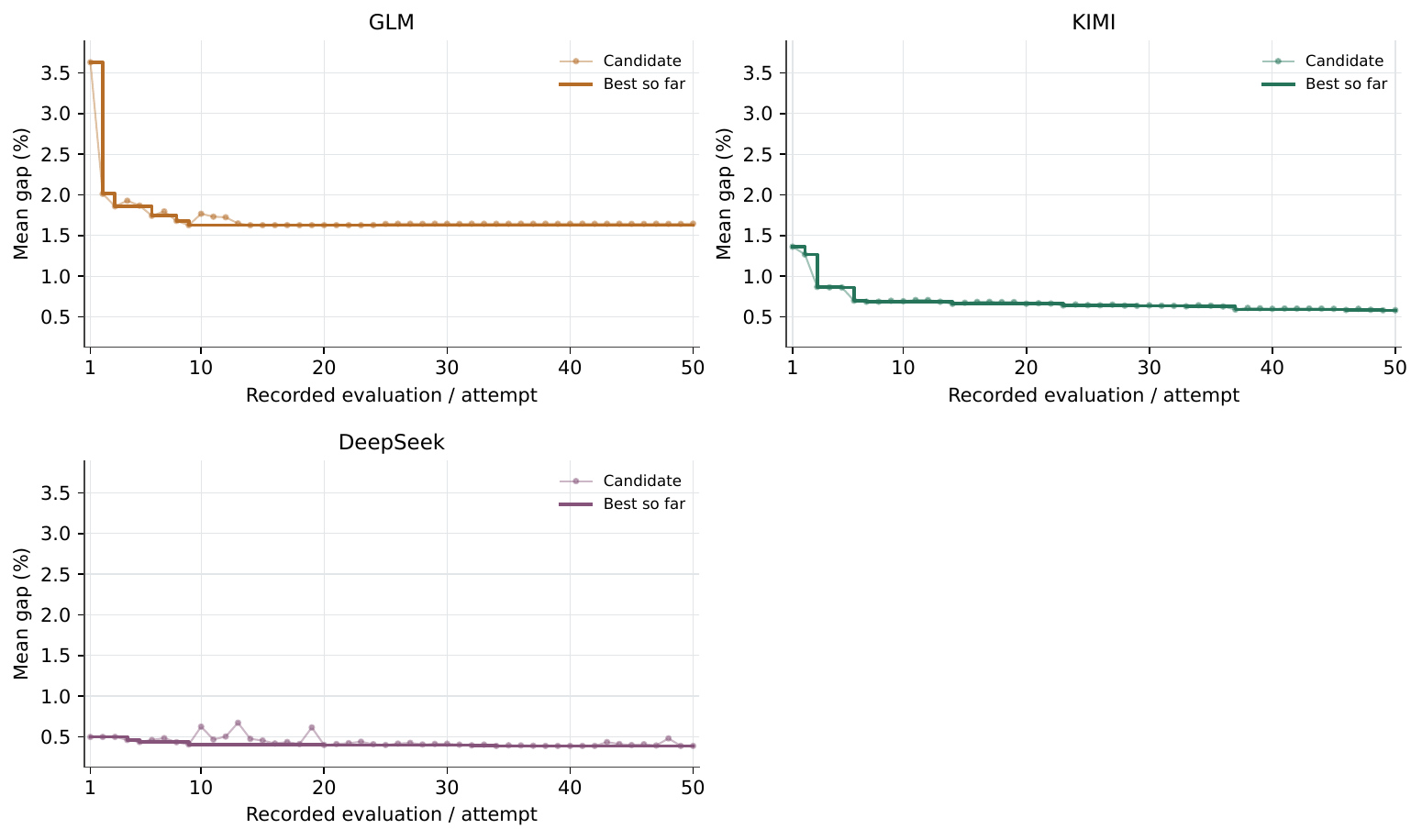}
\caption{TSP: solver development from scratch. Training observations only; repeated candidates are not independent development campaigns.}
\label{fig:appendix-TSP-scratch}
\end{figure}
\clearpage
\begin{figure}[htbp]
\centering
\includegraphics[width=\linewidth]{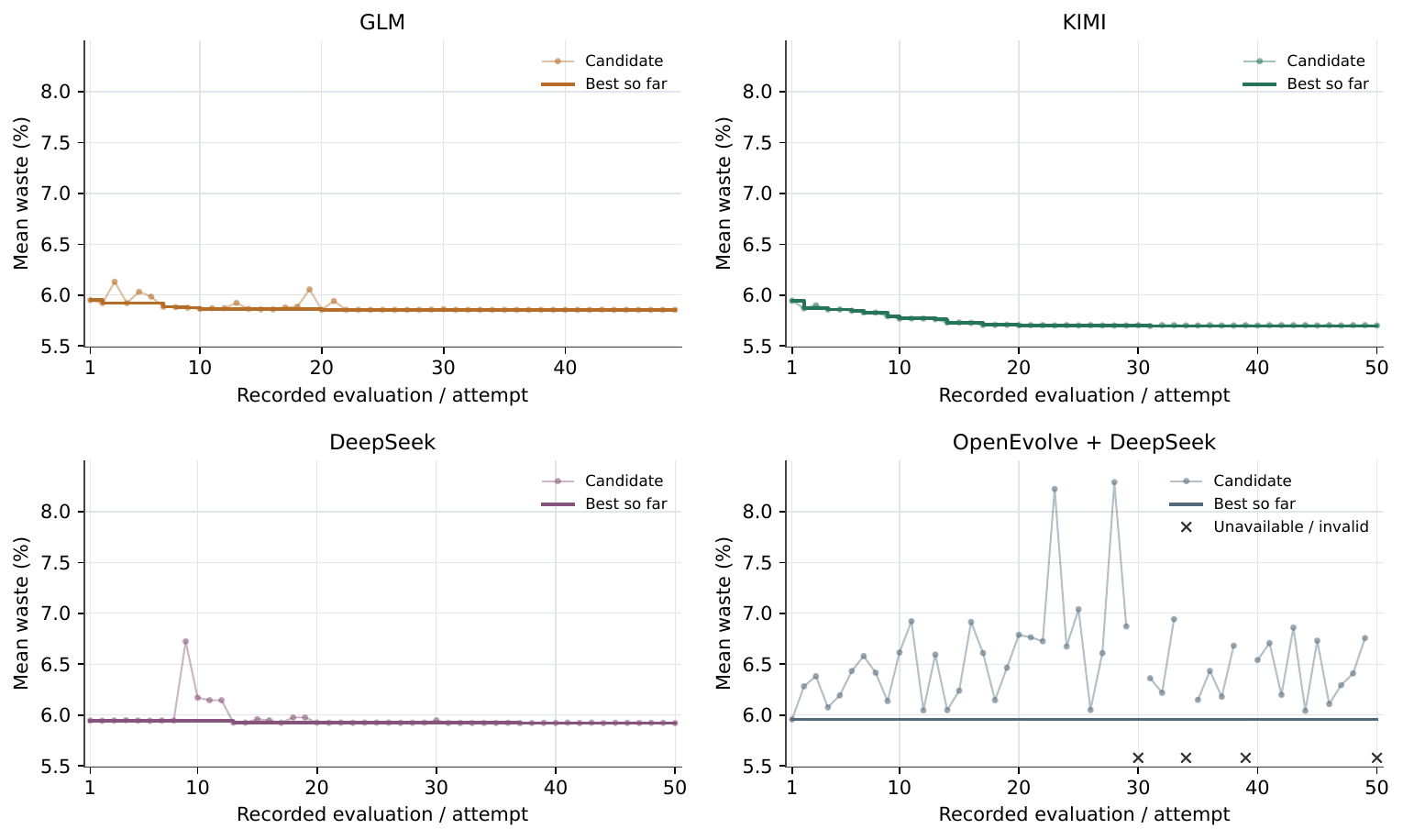}
\caption{ROADEF2018: improvement of the champion solver. KIMI shows its B-indexed development policy; GLM has 49 recovered complete rounds. OpenEvolve selects the original solver at counted round 1.}
\label{fig:appendix-R18-improve}
\end{figure}
\clearpage
\begin{figure}[htbp]
\centering
\includegraphics[width=\linewidth]{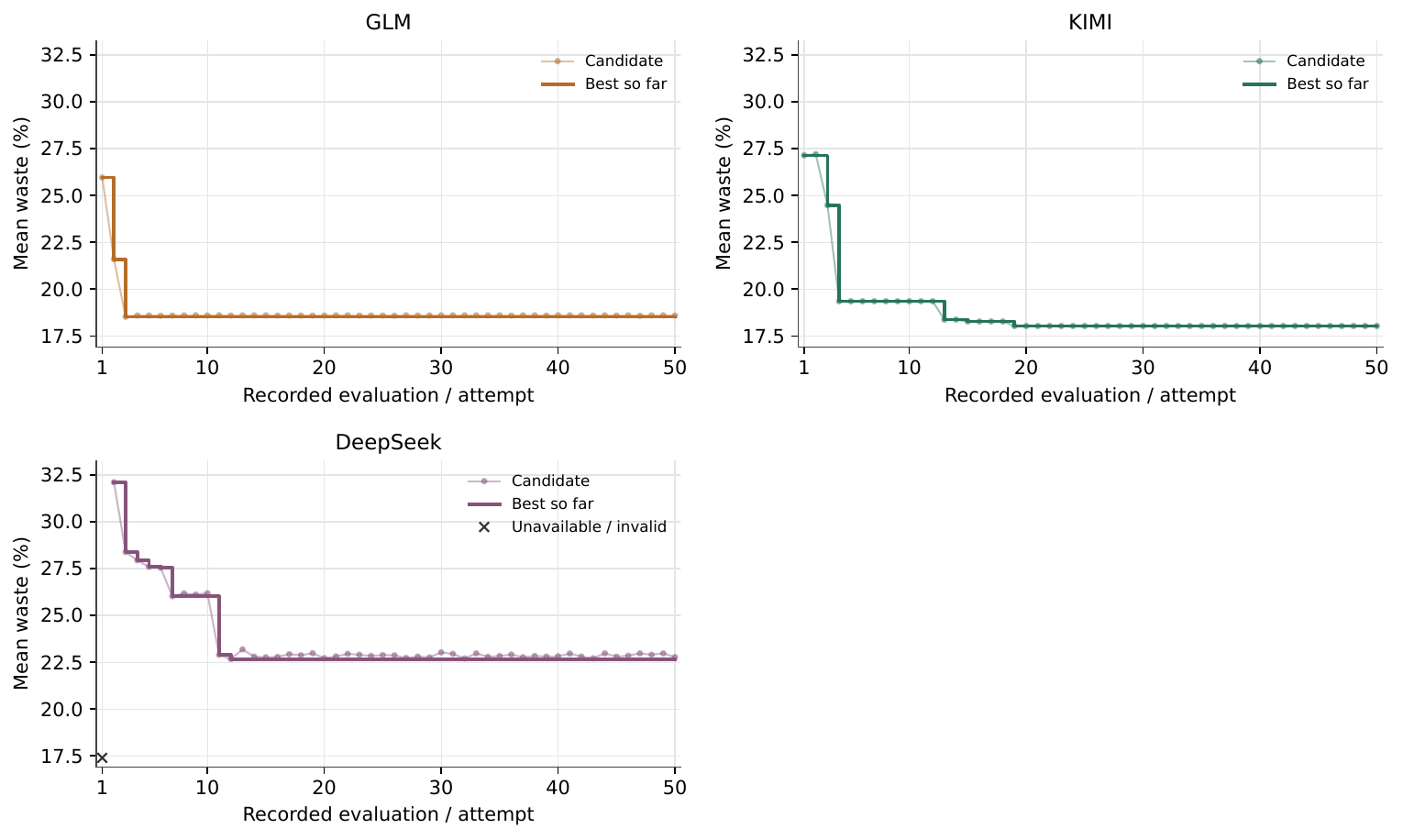}
\caption{ROADEF2018: solver development from scratch. DeepSeek\textquotesingle s earlier archived failed campaign is not included in this current-campaign plot. Resource and archive qualifications remain applicable.}
\label{fig:appendix-R18-scratch}
\end{figure}
\clearpage
\begin{figure}[htbp]
\centering
\includegraphics[width=\linewidth]{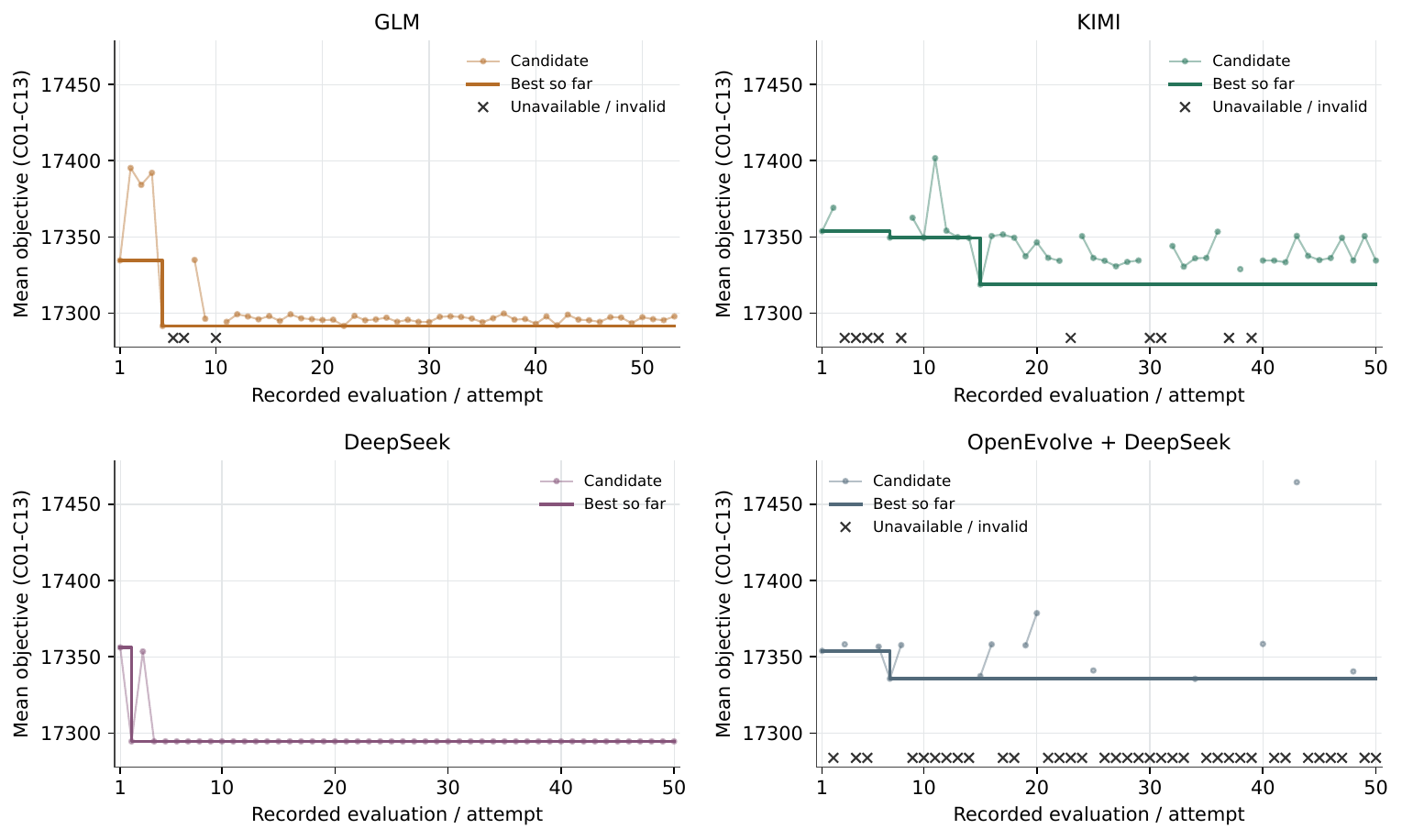}
\caption{ROADEF2020: improvement of ALNS. Quality is the mean objective on the fixed C01--C13 subset and is plotted only when all 39 corresponding solves are valid; it does not imply full C01--C15 validity. GLM\textquotesingle s disputed attempts 6 and 10 do not update the incumbent. OpenEvolve includes 50 counted evaluations; its selected candidate is attempt 51 (counted round 48).}
\label{fig:appendix-R20-improve}
\end{figure}
\clearpage
\begin{figure}[htbp]
\centering
\includegraphics[width=\linewidth]{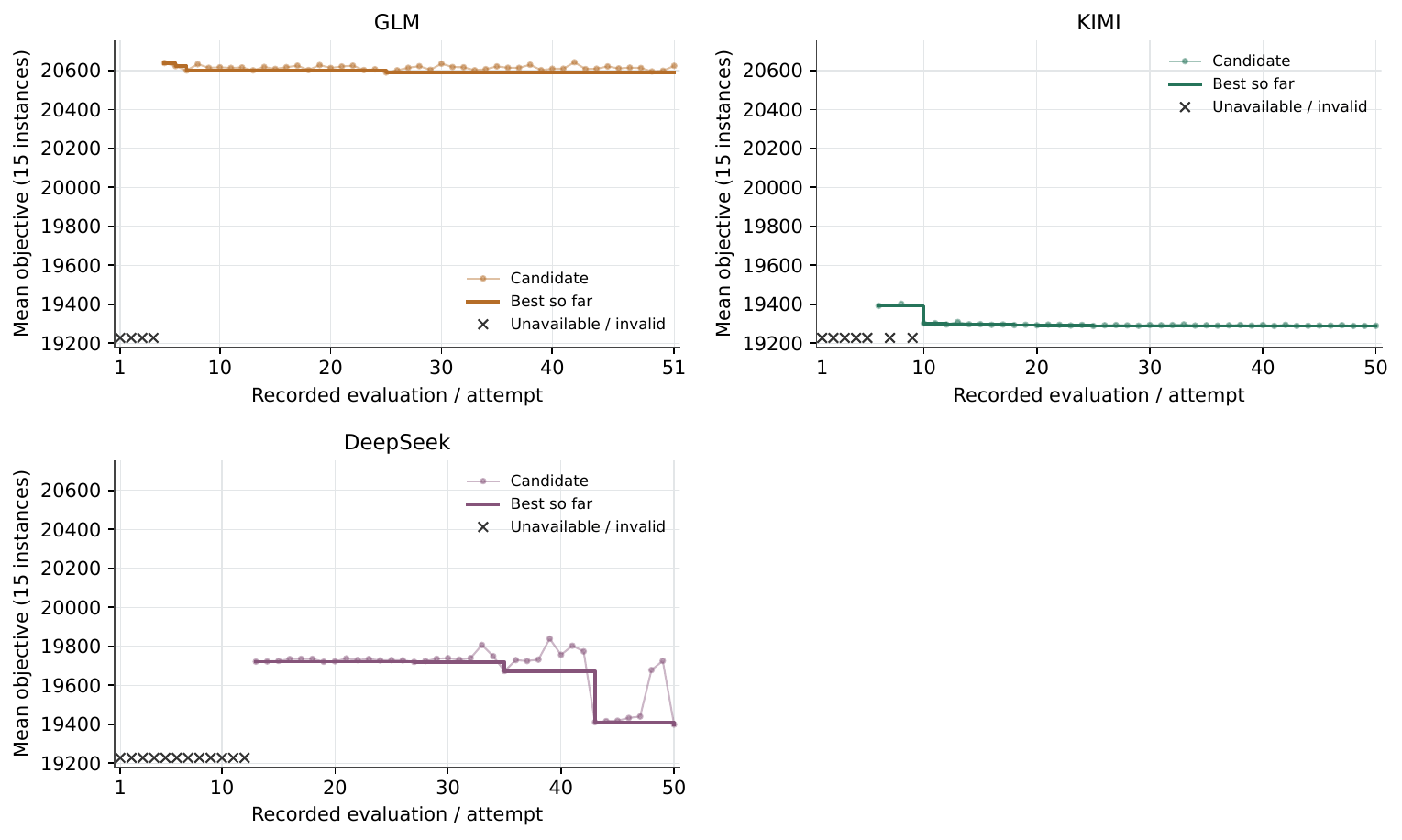}
\caption{ROADEF2020: solver development from scratch. Quality requires 45/45 valid solves on all 15 instances. GLM is shown in actual execution order, with the incident-affected first round unavailable and the final clean repeat retained (51 recorded round IDs). DeepSeek\textquotesingle s shorter internal search budgets remain part of its experiment; equal round counts do not imply equal computing expenditure.}
\label{fig:appendix-R20-scratch}
\end{figure}
\clearpage

\begin{figure}[htbp]\centering
\includegraphics[width=\linewidth]{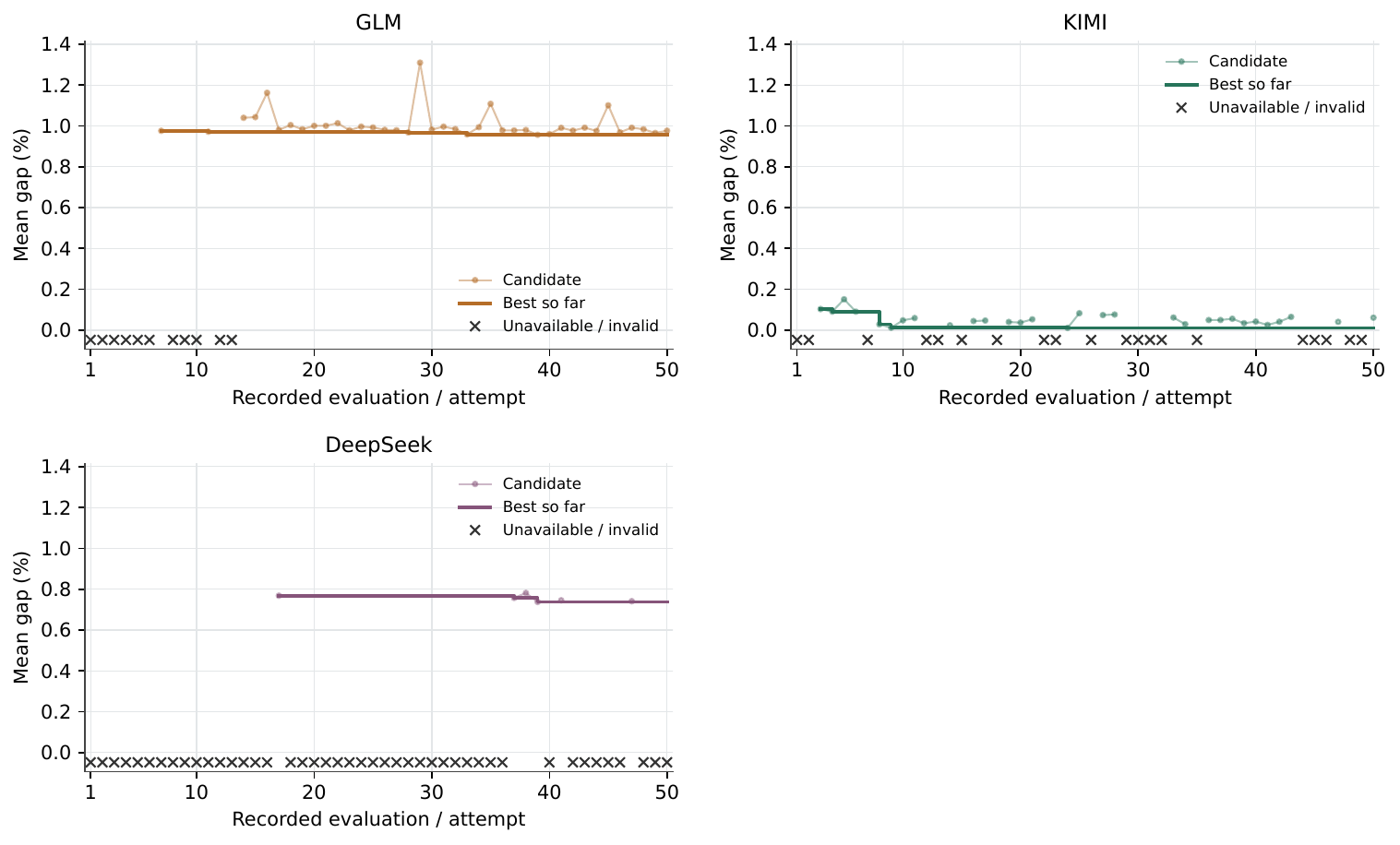}
\caption{ROADEF 2022, improvement of S46--HiGHS V6. Each point requires 50 eligible C solutions from the single-seed development evaluation. Crosses mark incomplete rounds. Repeated versions remain in the training history.}\label{fig:r22-improve}\end{figure}
\begin{figure}[htbp]\centering
\includegraphics[width=\linewidth]{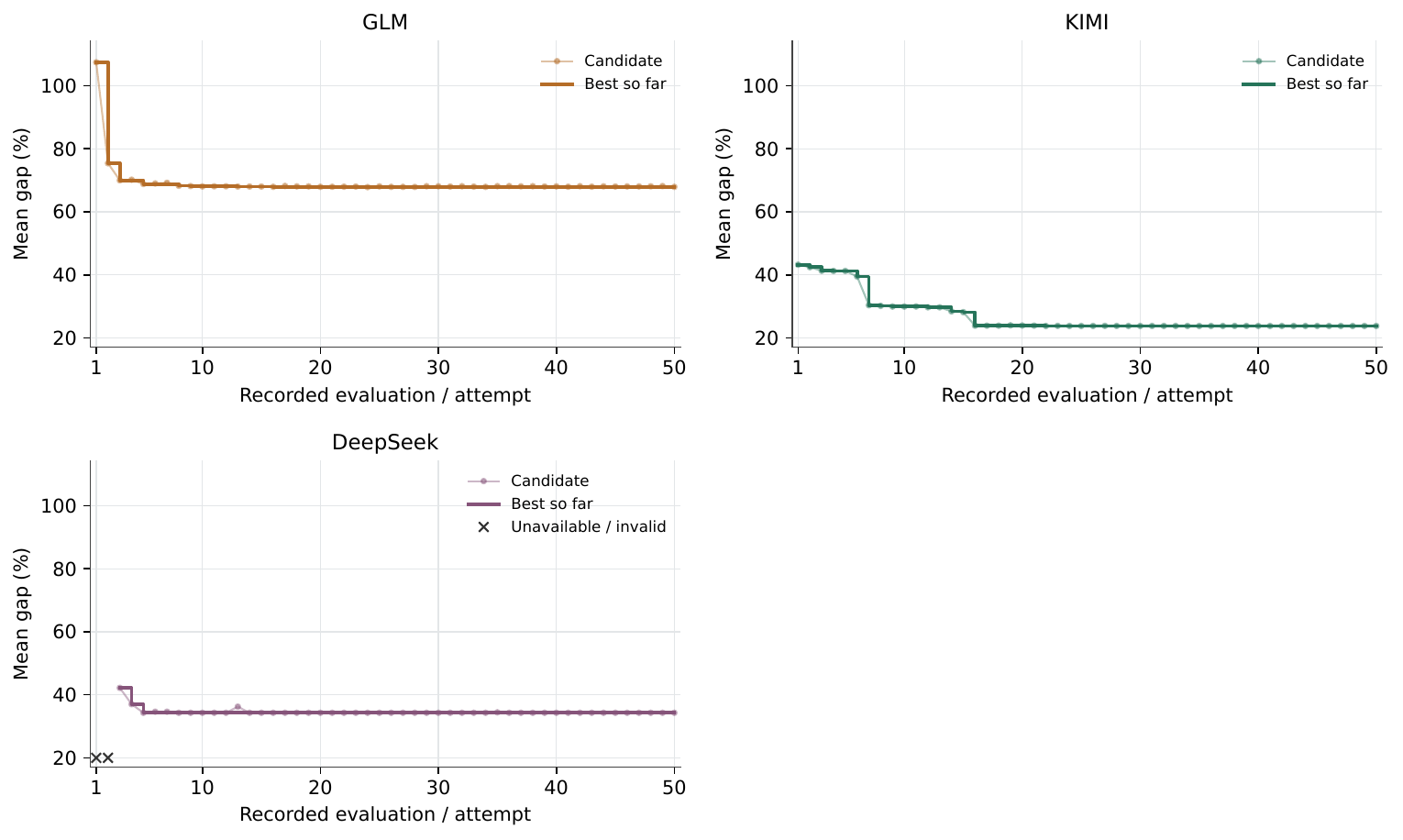}
\caption{ROADEF 2022, development from scratch. Each point requires 50 eligible C solutions from the single-seed development evaluation. Crosses mark incomplete rounds. Repeated versions remain in the training history.}\label{fig:r22-scratch}\end{figure}
\FloatBarrier

\subsection{Frozen-solver execution accounting}
\label{app:execution}
CPU hours describe final solver evaluations rather than API usage or complete development expenditure. ``Excess $>1$s'' counts solves exceeding their nominal time allowance by more than one second; ``Max excess'' is the largest recorded excess. Solver validity and resource compliance are distinct checks.

\par\medskip\noindent\begin{minipage}{\linewidth}\centering\small
\captionof{table}{Frozen-solver execution accounting (part 1).}\label{tab:final-resources-0}
\AppendixFitTable{\begin{tabular}{ccccc}\toprule
Method & Split & CPU hours & Excess $>1$s & Max excess (s) \\ \midrule
PFSP scratch KIMI & train & 29.42 & 0 & 0.045 \\
PFSP scratch GLM & train & 29.36 & 0 & 0.122 \\
PFSP scratch DeepSeek & train & 29.36 & 0 & 0.068 \\
PFSP scratch KIMI & test & 649.22 & 279 & 5.750 \\
PFSP scratch GLM & test & 645.77 & 2 & 1.428 \\
PFSP scratch DeepSeek & test & 648.63 & 26 & 2.600 \\
TSP scratch KIMI & train & 94.54 & 0 & 0.000 \\
TSP scratch GLM & train & 94.08 & 0 & 0.000 \\
TSP scratch DeepSeek & train & 94.89 & 9 & 2.242 \\
TSP scratch KIMI & test & 209.70 & 3 & 7.518 \\
TSP scratch GLM & test & 209.41 & 1 & 3.353 \\
TSP scratch DeepSeek & test & 209.92 & 26 & 7.301 \\
R20 improve DeepSeek & test & 31.85 & 0 & 0.000 \\
R20 improve GLM & test & 32.09 & 0 & 0.000 \\
R20 improve KIMI & test & 31.44 & 0 & 0.000 \\
R18 scratch DeepSeek & test & 349.85 & 0 & 0.000 \\
R18 scratch DeepSeek & train & 85.27 & 0 & 0.000 \\
R18 scratch GLM & test & 354.79 & 0 & 0.000 \\
R18 scratch GLM & train & 86.11 & 0 & 0.000 \\
R18 scratch KIMI & test & 1.01 & 0 & 0.000 \\
\bottomrule\end{tabular}}
\end{minipage}\par\medskip

\par\medskip\noindent\begin{minipage}{\linewidth}\centering\small
\captionof{table}{Frozen-solver execution accounting (part 2).}\label{tab:final-resources-20}
\AppendixFitTable{\begin{tabular}{ccccc}\toprule
Method & Split & CPU hours & Excess $>1$s & Max excess (s) \\ \midrule
R18 scratch KIMI & train & 1.13 & 0 & 0.000 \\
R20 scratch GLM & test & 43.04 & 0 & 0.000 \\
R20 scratch GLM & train & 43.18 & 0 & 0.000 \\
R20 scratch KIMI & test & 42.64 & 0 & 0.000 \\
R20 scratch KIMI & train & 42.70 & 0 & 0.000 \\
R18 human Champion & test & 178.63 & 0 & 0.000 \\
R18 human Champion & train & 43.51 & 0 & 0.000 \\
R20 human ALNS & test & 10.18 & 0 & 0.000 \\
R20 human ALNS & train & 10.18 & 0 & 0.000 \\
R20 scratch DeepSeek & test & 10.78 & 0 & 0.000 \\
R20 scratch DeepSeek & train & 10.78 & 0 & 0.000 \\
\bottomrule\end{tabular}}
\end{minipage}\par\medskip

\begin{table}[!htbp]\centering\small
\caption{ROADEF 2022 frozen retests on CPU4. Solver CPU hours sum the launcher\textquotesingle s \texttt{wait4} records, which can omit unreaped descendants. Peak memory is sampled process-group RSS in decimal GB. Timeouts and timing excess are reported separately.}
\label{tab:r22-resources}
\AppendixFitTable{\begin{tabular}{@{}lrrrrrr@{}}\toprule
Solver & Split & CPU h & Timeouts & Excess $>1$s & Max excess (s) & Peak GB \\ \midrule
V6 & C & 4.82 & 0 & 0 & 0.00 & 2.22 \\
V6 & X & 8.28 & 0 & 0 & 0.00 & 5.12 \\
Improve, Kimi-code (v011) & C & 29.45 & 0 & 0 & 0.00 & 9.53 \\
Improve, Kimi-code (v011) & X & 51.77 & 0 & 0 & 0.00 & 12.78 \\
Improve, Zcode (v031) & C & 4.67 & 0 & 0 & 0.00 & 2.68 \\
Improve, Zcode (v031) & X & 8.61 & 0 & 0 & 0.00 & 6.39 \\
Improve, DeepSeek Harness (v062) & C & 4.87 & 0 & 0 & 0.00 & 2.05 \\
Improve, DeepSeek Harness (v062) & X & 8.55 & 0 & 0 & 0.00 & 4.32 \\
Scratch, Kimi-code (v046) & C & 47.26 & 0 & 0 & 0.00 & 0.39 \\
Scratch, Kimi-code (v046) & X & 100.76 & 0 & 0 & 0.00 & 0.36 \\
Scratch, Zcode (v014) & C & 260.51 & 0 & 0 & 0.00 & 0.73 \\
Scratch, Zcode (v014) & X & 543.49 & 0 & 0 & 0.00 & 0.80 \\
Scratch, DeepSeek Harness (v007) & C & 0.79 & 0 & 0 & 0.00 & 0.09 \\
Scratch, DeepSeek Harness (v007) & X & 2.10 & 0 & 0 & 0.00 & 0.09 \\
\bottomrule\end{tabular}}\end{table}
\FloatBarrier
For OpenEvolve, frozen training confirmation and held-out testing used 43.48 and 177.98 solver CPU hours on ROADEF 2018, and 28.83 and 27.85 hours on ROADEF 2020, respectively. All four evaluations finished without a timeout.

\clearpage
\section{Experimental materials and environment}
\label{app:setup}
\subsection{Training and test instance identifiers and sizes}
\label{app:instances}
Tables~\ref{tab:instances-pfsp}--\ref{tab:instances-r20} list the training and test instances used in both development scenarios. Filename extensions are omitted. PFSP identifiers are specified by their full index ranges; TSP and ROADEF identifiers are listed individually. ROADEF sizes are counted from the original instance files: distinct item and stack identifiers for 2018, and intervention entries and the scheduling horizon $T$ for 2020.

\par\medskip\noindent\begin{minipage}{\linewidth}\centering\small
\captionof{table}{PFSP instance identifiers and sizes. Each identifier is \texttt{VFR}$n$\texttt{\_}$m$\texttt{\_}$k$; every combination of the listed machine counts and indices is included.}\label{tab:instances-pfsp}
\begin{tabular*}{\linewidth}{@{\extracolsep{\fill}}ccccc@{}}\toprule
Split & Jobs $n$ & Machines $m$ & Instance index $k$ & Count \\ \midrule
Training & 100 & $\{20,40,60\}$ & $1,\ldots,10$ & 30 \\
 & 200 & $\{20,40,60\}$ & $1,\ldots,10$ & 30 \\
\midrule
Test & 300 & $\{20,40,60\}$ & $1,\ldots,10$ & 30 \\
 & 400 & $\{20,40,60\}$ & $1,\ldots,10$ & 30 \\
 & 500 & $\{20,40,60\}$ & $1,\ldots,10$ & 30 \\
 & 600 & $\{20,40,60\}$ & $1,\ldots,10$ & 30 \\
 & 700 & $\{20,40,60\}$ & $1,\ldots,10$ & 30 \\
 & 800 & $\{20,40,60\}$ & $1,\ldots,10$ & 30 \\
\bottomrule\end{tabular*}\end{minipage}\par\medskip
\noindent\begin{minipage}{\linewidth}\centering\small
\captionof{table}{TSP instance identifiers and city counts. Parallel column groups compactly list the same split.}\label{tab:instances-tsp}
\begin{tabular*}{\linewidth}{@{\extracolsep{\fill}}cccccc@{}}\toprule
Instance & Cities & Instance & Cities & Instance & Cities \\
\midrule\multicolumn{6}{c}{Training (33 instances)}\\\midrule
fqm5087 & 5,087 & ida8197 & 8,197 & brd14051 & 14,051 \\
fea5557 & 5,557 & ei8246 & 8,246 & mo14185 & 14,185 \\
rl5915 & 5,915 & ar9152 & 9,152 & xrb14233 & 14,233 \\
rl5934 & 5,934 & dga9698 & 9,698 & ho14473 & 14,473 \\
tz6117 & 6,117 & ja9847 & 9,847 & d15112 & 15,112 \\
xsc6880 & 6,880 & gr9882 & 9,882 & it16862 & 16,862 \\
eg7146 & 7,146 & kz9976 & 9,976 & xia16928 & 16,928 \\
bnd7168 & 7,168 & xmc10150 & 10,150 & pjh17845 & 17,845 \\
lap7454 & 7,454 & fi10639 & 10,639 & d18512 & 18,512 \\
ym7663 & 7,663 & rl11849 & 11,849 & frh19289 & 19,289 \\
pm8079 & 8,079 & xvb13584 & 13,584 & fnc19402 & 19,402 \\
\midrule\multicolumn{6}{c}{Test (21 instances)}\\\midrule
ido21215 & 21,215 & irx28268 & 28,268 & fry33203 & 33,203 \\
fma21553 & 21,553 & fyg28534 & 28,534 & bm33708 & 33,708 \\
vm22775 & 22,775 & icx28698 & 28,698 & bby34656 & 34,656 \\
lsb22777 & 22,777 & boa28924 & 28,924 & pba38478 & 38,478 \\
xrh24104 & 24,104 & ird29514 & 29,514 & ics39603 & 39,603 \\
sw24978 & 24,978 & pbh30440 & 30,440 & rbz43748 & 43,748 \\
bbz25234 & 25,234 & xib32892 & 32,892 & fht47608 & 47,608 \\
\bottomrule\end{tabular*}\end{minipage}\par\medskip
\clearpage
\noindent
\begin{minipage}[t]{.485\linewidth}\centering\footnotesize
\captionof{table}{ROADEF 2018 instance identifiers and size fields. Items and stacks count the required glass pieces and ordered production stacks.}\label{tab:instances-r18}
\setlength{\tabcolsep}{3pt}\begin{tabular*}{\linewidth}{@{\extracolsep{\fill}}cccccc@{}}\toprule
\multicolumn{3}{c}{Training (15)} & \multicolumn{3}{c}{Test (15)} \\\cmidrule(lr){1-3}\cmidrule(lr){4-6}
Instance & Items & Stacks & Instance & Items & Stacks \\ \midrule
B1 & 68 & 68 & X1 & 300 & 10 \\
B2 & 383 & 13 & X2 & 247 & 247 \\
B3 & 332 & 12 & X3 & 258 & 12 \\
B4 & 261 & 12 & X4 & 371 & 18 \\
B5 & 207 & 2 & X5 & 124 & 16 \\
B6 & 204 & 9 & X6 & 412 & 18 \\
B7 & 241 & 241 & X7 & 215 & 10 \\
B8 & 334 & 12 & X8 & 173 & 2 \\
B9 & 247 & 12 & X9 & 233 & 12 \\
B10 & 214 & 13 & X10 & 375 & 15 \\
B11 & 274 & 15 & X11 & 362 & 14 \\
B12 & 439 & 18 & X12 & 302 & 14 \\
B13 & 656 & 18 & X13 & 344 & 16 \\
B14 & 267 & 14 & X14 & 253 & 12 \\
B15 & 431 & 21 & X15 & 296 & 12 \\
\bottomrule\end{tabular*}\end{minipage}\hfill%
\begin{minipage}[t]{.485\linewidth}\centering\footnotesize
\captionof{table}{ROADEF 2020 instance identifiers and size fields. Tasks counts maintenance interventions; periods is the scheduling horizon.}\label{tab:instances-r20}
\setlength{\tabcolsep}{3pt}\begin{tabular*}{\linewidth}{@{\extracolsep{\fill}}cccccc@{}}\toprule
\multicolumn{3}{c}{Training (15)} & \multicolumn{3}{c}{Test (15)} \\\cmidrule(lr){1-3}\cmidrule(lr){4-6}
Instance & Tasks & Periods & Instance & Tasks & Periods \\ \midrule
C\_01 & 120 & 53 & X\_01 & 120 & 53 \\
C\_02 & 120 & 53 & X\_02 & 706 & 53 \\
C\_03 & 706 & 53 & X\_03 & 280 & 53 \\
C\_04 & 706 & 53 & X\_04 & 426 & 25 \\
C\_05 & 706 & 53 & X\_05 & 467 & 220 \\
C\_06 & 280 & 53 & X\_06 & 528 & 300 \\
C\_07 & 120 & 42 & X\_07 & 209 & 300 \\
C\_08 & 426 & 25 & X\_08 & 209 & 300 \\
C\_09 & 110 & 53 & X\_09 & 548 & 30 \\
C\_10 & 522 & 102 & X\_10 & 460 & 35 \\
C\_11 & 89 & 102 & X\_11 & 521 & 131 \\
C\_12 & 298 & 191 & X\_12 & 522 & 131 \\
C\_13 & 505 & 230 & X\_13 & 336 & 212 \\
C\_14 & 465 & 220 & X\_14 & 613 & 180 \\
C\_15 & 528 & 300 & X\_15 & 613 & 180 \\
\bottomrule\end{tabular*}\end{minipage}\par\medskip

\begin{table}[!htbp]\centering\small
\caption{ROADEF 2022 splits and input sizes. Ranges count data rows in the item and truck input files, excluding headers.}
\label{tab:instances-r22}
\AppendixFitTable{\begin{tabular}{@{}lrrrr@{}}\toprule
Split & Seconds & Instances & Item rows & Truck rows \\ \midrule
C & 900 & 25 & 684--39,631 & 844--165,218 \\
C & 1500 & 25 & 11,867--65,274 & 48,343--147,929 \\
X & 900 & 49 & 2,071--50,750 & 2,524--131,540 \\
X & 1500 & 56 & 123--46,070 & 190--107,869 \\
\bottomrule\end{tabular}}\end{table}
\FloatBarrier
Set C includes AS, BY, CA, CI, CL, DO, FC, FL, MA, MT, PA, RM, SA, TA, TR and VA, each unsuffixed and with suffixes 2 and 3, plus PO and VO. Set X includes AS, BY, CC, CI, DO, FC, FS, MA, MT, PA, RM, SA, TA, TR and VA, each unsuffixed and with suffixes 2 through 7. Identically named instances in different splits use separate data files. For reproducibility we retain the fixed workbook references, including 5,478,066 for X/SA7; the organizers\textquotesingle\ web summary lists 5,488,066 for this entry.

\subsection{Example task document and launch prompt}
\label{app:task-example}
PFSP development from scratch is used as the example. The task document below is an English translation of the retained KIMI PFSP scratch \texttt{TASK.md}, with the experiment directory represented by \texttt{<WORKSPACE>}. It preserves the task-specific requirements rather than introducing a new evaluation protocol. The launch prompt is translated from the recovered message actually supplied for this KIMI experiment; local credential paths, internal server connection details, account names, and remote workspace paths are replaced with angle-bracket placeholders, and no credential values are reproduced.

\subsubsection{TASK.md (English translation)}
\begin{Verbatim}[fontsize=\small,breaklines=true,breakanywhere=true,frame=single]
# PFSP solver development from scratch

## Task
Implement a permutation flow-shop scheduling (PFSP) solver from scratch to minimize makespan, and improve it using training feedback. Choose the algorithmic structure yourself; using iterated greedy (IG) is not required. No initial solver implementation is supplied. First read initial_materials/PROBLEM.md, tools/check_solution.py, and the training configuration.

## Authorized workspace
Use only <WORKSPACE> for development, compilation, evaluation, and saved files. System runtime libraries may be used for compilation and execution. Do not access other experiments, other agents' results, or hidden test data. Do not obtain an existing PFSP solver as a starting point. Training reference values are for scoring only: do not hard-code solutions or strategy tables by instance. The final solver must handle unseen instances of the same problem class.

## Experimental settings
- Training data: 60 VFR instances in training/, with n in {100, 200} and m in {20, 40, 60}, ten instances per size. training_manifest.json contains the full list, input hashes, reference values, and fixed seeds.
- Complete 50 full evaluations. Each evaluation consists of 60 instances x 10 repetitions = 600 solves. Full evaluations containing failed solves count toward the budget; do not delete or renumber them. Compilation failures before full evaluation do not count. Preserve interrupted evaluations and request recovery instead of discarding and repeating them.
- Per-solve time limit: 0.03*n*m seconds, including input reading, initialization, and solution output. Reserve time to write the result.
- Use at most one logical CPU per solve, including subprocesses. The evaluator permits at most 400 concurrent solves and queues all 600 jobs. Active jobs use distinct CPUs. Run full evaluations sequentially; do not debug, compile, or start another evaluation concurrently with a full evaluation.
- The historical watchdog is 1.5*time_limit + 10 seconds. It handles stalled processes and does not extend the permitted search budget. Nominal-budget excess is recorded separately for audit.
- The allocated server-memory ceiling is 750 GB. Account for total memory when running 400 jobs concurrently. This task adds no new per-solve memory ceiling; do not exhaust server memory.

## Implementation and evaluation interface
Develop in src/. Freeze each complete candidate in candidates/<version>/, including source, build instructions, executable, and run.sh. Do not depend on mutable source outside the candidate directory.

run.sh must accept:
  --instance PATH --output PATH --time-limit SECONDS --seed INT
Return exit code zero on success. Prefer exec to launch the actual solver. Write JSON of the form:
  {"permutation": [0, 1, ...]}
Job indices start at zero. If makespan is included, it must match the checker. Support absolute paths: candidates are copied into an evaluation snapshot before execution. Write temporary files in the current working directory.

Compile the candidate and make run.sh executable. Check configuration without solving:
  python3 tools/evaluate_train.py --candidate candidates/v001 --id round001 --dry-run
Launch a full evaluation in the background and monitor its status; do not resubmit because a terminal wait times out:
  nohup python3 tools/evaluate_train.py --candidate candidates/v001 --id round001 > logs/round001.log 2>&1 < /dev/null &
Read evaluations/round001/status.json and the final summary.json. Do not edit a candidate during evaluation. The evaluator freezes a snapshot and records hashes, seeds, CPU affinity, elapsed time, CPU time, peak RSS, and checker results in evaluations/ledger.jsonl.
Do not modify training data, reference values, seeds, evaluation rules, or the checker to improve scores. Record and report any tool defect first.

## Improvement criteria and records
Compute gap (%) = (makespan - BKR) / BKR * 100; lower is better. For each instance report mean and best gap over ten runs, then weight all 60 instances equally. Report a complete evaluation score only when all 600 runs are valid and the candidate snapshot is unchanged. Never silently remove failures from averages; retain each failure reason. An independent checker validates the permutation and recomputes makespan.

Fifty evaluations do not imply fifty new algorithms. Continue proposing and testing modifications, distinguish new candidates from unchanged repeats, and do not present repeated runs as algorithmic evolution. Do not stop merely because an early plateau appears. Record candidate changes, rationale, acceptance or rejection, and every evaluation. Account for partial development tests separately, including instances, seeds, budgets, outcomes, and resources. Do not disguise extra full evaluations or batch candidate screening as debugging.

An external administrator monitors CPU use, memory, processes, and CPU affinity by account UID. The setup uses directory isolation and recording-based monitoring rather than cgroup resource enforcement. Do not expand CPU affinity or disable monitoring. Violations remain recorded; monitoring does not automatically stop the experiment.

## Deliverables
Retain the best candidate's complete source and executable, build and run instructions, the full 50-evaluation ledger and raw outputs, DECISIONS.md, REPORT.md, and per-round convergence data in CSV format. Report validity, training mean/best gaps, nominal time excesses, resource use, failures, and limitations. Select the final candidate using training data only; do not run the held-out evaluation yourself.
\end{Verbatim}

\subsubsection{Launch prompt (English translation of the actual PFSP message)}
\begin{Verbatim}[fontsize=\small,breaklines=true,breakanywhere=true,frame=single]
Please log into CPU3 and conduct the experiment "PFSP solver development from scratch".

Read the login details from this local file:
<CREDENTIAL_FILE>

The file contains the server address, port, username and password. Do not display the password in replies or logs.

Server: <SERVER_HOST>
SSH port: <SSH_PORT>
Account: <ACCOUNT>

Your only experiment workspace is:
<WORKSPACE>

First read:
<WORKSPACE>/TASK.md

Then read the problem description, training manifest, checker and evaluation entry point, and verify the protocol. Briefly explain your understanding, then start implementation and iteration directly, without waiting for another confirmation.

This is PFSP solver design from scratch, not modification of an existing IG solver. No initial solver code is supplied; choose the algorithmic structure yourself.

Key requirements:
1. Complete 50 full training evaluations. Each round has 60 instances with 10 repetitions each, totaling 600 solves; strictly use the fixed seeds and prescribed evaluation entry point.
2. Each solve is limited to 0.03*n*m seconds and at most one logical CPU. At most 400 solves may run concurrently; rounds must run sequentially. Watchdog tolerance is not additional search time.
3. Develop and save files only in the specified workspace. Do not access other experiments, other agents' results or test data, and do not obtain existing PFSP solver code as a starting point.
4. Freeze each candidate, preserving source, build instructions and executable; do not modify it during evaluation. Launch long evaluations in the background and track status files. Do not resubmit when a terminal wait times out.
5. Continue trying evidence-based algorithmic improvements. Distinguish new-candidate evaluations from unchanged repeats; do not describe repeated runs as new algorithm evolution. Record failures and development tests honestly, without deletion or concealment.
6. External resource monitoring has already started; you do not need to start or stop it. Respect resource constraints and do not interfere with monitoring.
7. On completion, save the best candidate, every evaluation record, DECISIONS.md, REPORT.md and the raw convergence CSV. Report validity, training mean/best gaps and resource use. Do not conduct the final held-out evaluation yourself.

TASK.md is authoritative for the detailed requirements. If materials or tools have a blocking problem, preserve evidence and report it; do not change the experimental protocol on your own.
\end{Verbatim}

\subsection{Agent versions and computing environment}
\label{app:environment}
The main text uses short model labels. Fields without a specified version or reasoning-effort setting are marked with a dash; current software installations are not used as substitutes for historical versions.

\begin{table}[htbp]\centering\small
\setlength{\tabcolsep}{4pt}
\caption{Agent and workflow metadata, including the reasoning-effort setting. A dash indicates no version or reasoning-effort setting was specified.}
\begin{tabularx}{\linewidth}{@{}c c >{\centering\arraybackslash}X c c@{}}\toprule
Method & Model label & Model name & \shortstack{Agent/framework\\version} & \shortstack{Reasoning\\effort} \\ \midrule
Codex & GPT & \texttt{GPT-6-Astra} & 0.159.0-alpha.12.1 & medium \\
Kimi-code & KIMI & \texttt{KIMI-K3} & 0.43.1 & high \\
Zcode & GLM & \texttt{GLM-5.3} & 3.11.2 & max \\
DeepSeek Harness & DeepSeek & \texttt{DeepSeek-V4-\allowbreak{}Flash-1017} & 0.2.0-rc2 & max \\
ReEvo & DeepSeek & \texttt{DeepSeek-V4-\allowbreak{}Flash-1017} & - & - \\
OpenEvolve & DeepSeek & \texttt{DeepSeek-V4-\allowbreak{}Flash-1017} & 0.4.0 & - \\
\bottomrule\end{tabularx}\end{table}

\begin{table}[htbp]\centering\small
\caption{Server hardware and software inventory. CPU3 and CPU4 are experiment host labels; hardware is shared in specification, not a claim of identical runtime load.}
\begin{tabularx}{\linewidth}{@{}c >{\centering\arraybackslash}X >{\centering\arraybackslash}X@{}}\toprule
Field & CPU3 & CPU4 \\ \midrule
Processor model & AMD EPYC 9754 & AMD EPYC 9754 \\
CPU sockets & 2 & 2 \\
Physical cores / logical CPUs & 256 / 512 (SMT2) & 256 / 512 (SMT2) \\
Host-visible RAM & Approximately 1 TB & Approximately 1 TB \\
Operating system and release & Ubuntu 22.04 LTS & Ubuntu 22.04.2 LTS \\
Compiler and version & GCC/G++ 11.4.0 & GCC/G++ 11.4.0 \\
\bottomrule\end{tabularx}\end{table}

The recorded per-solve allocations are one logical CPU for PFSP and TSP, up to eight logical CPUs and 48 GB for ROADEF 2018, four logical CPUs (two physical cores with SMT) and 16 GB for ROADEF 2020, and eight logical CPUs and 30 GB for ROADEF 2022. These are experiment allocations, not statements of installed server capacity. The PFSP example above records a 750 GB experiment-wide allocation rather than a measured RAM specification. Historical PFSP/TSP comparison records originate from CPU3; OpenEvolve's TSP test and the retained industrial reference/scratch confirmations use CPU4. CPU4 was queried directly; CPU3 hardware follows the confirmed identical host specification because CPU3 is no longer accessible. Its retained inventory independently records GCC/G++ 11.4.0 and approximately 1 TB host-visible memory. Host-visible capacity is distinct from the 750 GB experiment allocation and does not mean all 512 logical CPUs were allocated to an experiment.

For the retained OpenEvolve TSP comparison, round038 is the selected candidate; the restart cap changes from five during development to ten for the final protocol. The nominal test allowance is 3,600 seconds with a 60-second completion tolerance. Recorded solve times are 3,603.99--3,630.24 seconds. These execution details qualify the comparison without introducing another OpenEvolve result version.
\FloatBarrier

All ROADEF 2022 frozen evaluations ran on CPU4 using the official Java checker (version 1.11.4). The improvement track used the same HiGHS-backed V6 starting program for the three agents.

\fi
\end{document}